\documentclass[ROB,biber]{nowfnt} 
\usepackage[utf8]{inputenc}
\usepackage[T1]{fontenc}
\usepackage{amsmath}
\usepackage{amsfonts}
\usepackage{amssymb}

\DeclareMathOperator*{\argmax}{arg\,max}

\usepackage{color}
\usepackage{xcolor}
\definecolor{bistre}{rgb}{0.24, 0.17, 0.12}
\definecolor{auburn}{rgb}{0.43, 0.21, 0.1}

\usepackage{csquotes}
\usepackage[linesnumbered,ruled,vlined]{algorithm2e}

\SetCommentSty{mycommfont}
\SetKwInput{KwInput}{Input}                
\SetKwInput{KwOutput}{Output}              

\title{Social Interactions for Autonomous Driving: A Review and Perspectives}

\maintitleauthorlist{
Wenshuo Wang \\
McGill University \\
wenshuo.wang@mcgill.ca
\and
Letian Wang \\
University of Toronto \\
lt.wang@mail.utoronto.ca
\and
Chengyuan Zhang \\
McGill University \\
chengyuan.zhang@mail.mcgill.ca
\and
Changliu Liu \\
Carnegie Mellon University \\
cliu6@andrew.cmu.edu 
\and
Lijun Sun \\
McGill University \\
lijun.sun@mcgill.ca
}

\issuesetup
{%
 copyrightowner={W.~Wang, L.~Wang, C.~Zhang, C.~Liu~and~L. Sun},
 volume        = 10,
 issue         = 3-4,
 pubyear       = 2022,
 isbn          = xxx-x-xxxxx-xxx-x,
 eisbn         = xxx-x-xxxxx-xxx-x,
 doi           = 10.1561/2300000078,
 firstpage     = 198, 
 lastpage      = 376
 }

\usepackage{mwe}

\author[1]{Wang, Wenshuo}
\author[2]{Wang, Letian}
\author[3]{Zhang, Chengyuan}
\author[4]{Liu, Changliu}
\author[5]{Sun, Lijun}

\affil[1]{McGill University; wenshuo.wang@mcgill.ca}
\affil[2]{University of Toronto; lt.wang@mail.utoronto.ca}
\affil[3]{McGill University; chengyuan.zhang@mail.mcgill.ca}
\affil[4]{Carnegie Mellon University; cliu6@andrew.cmu.edu}
\affil[5]{McGill University; lijun.sun@mcgill.ca}

\articledatabox{\nowfntstandardcitation}

\begin{document}

\makeabstracttitle

\begin{abstract}
No human drives a car in a vacuum; she/he must negotiate with other road users to achieve their goals in social traffic scenes. A rational human driver can interact with other road users in a socially-compatible way through implicit communications to complete their driving tasks smoothly in interaction-intensive, safety-critical environments. This paper aims to review the existing approaches and theories to help understand and rethink the interactions among human drivers toward social autonomous driving. We take this survey to seek the answers to a series of fundamental questions: 1) What is social interaction in road traffic scenes? 2) How to measure and evaluate social interaction? 3) How to model and reveal the process of social interaction? 4) How do human drivers reach an implicit agreement and negotiate smoothly in social interaction? This paper reviews various approaches to modeling and learning the social interactions between human drivers, ranging from optimization theory, deep learning, and graphical models to social force theory and behavioral \& cognitive science. We also highlight some new directions, critical challenges, and opening questions for future research.
\end{abstract}

\chapter{Introduction}
\label{c-intro} 

\section{Background}
\label{subsec:background}
Humans can be trained to be remarkable drivers with powerful capabilities in social interaction. In real-world traffic, rational human drivers can make socially-compatible decisions in complex and crowded scenarios by efficiently negotiating with their surroundings using  non-linguistic communications such as gesturing (e.g., waving hands to the other car to give way), deictics (e.g., using turn signals to indicate intentions), and motion cues (e.g., accelerating/decelerating/turning) \citep{kauffmann2018makes}. Understanding the principles and rules of the dynamic interaction among human drivers in complex traffic scenes allows 1) generating diverse social driving behaviors that leverage beliefs and expectations about others' actions or reactions; 2) predicting the future states of a scene with moving objects, which is essential to building probably safe intelligent vehicles with the capabilities of behavior prediction \citep{wang2021socially, anderson2020low} and potential collision detection \citep{roy2020detection}; and 3) creating realistic driving simulators \citep{luo2019gamma}. However, this task is not trivial since various social factors exist along the driving interaction process, including social motivation\footnote{Social motivations are the factors that drive people to take action to interact with other people. Unlike motivation, which emphasizes the reasons or desires to do some actions, social motivation often requires interaction with other human agents.}, social perception\footnote{The social perception here refers to the processes by which a person uses the behavior of others to understand or reason about those individuals, particularly regarding their motives, attitudes, or values. Unlike object perception, social perception often involves sophisticated \textit{inferences} which go far beyond the data observed.}, and social control\footnote{Social control refers to sets of rules and standards that bound individuals to specific pressures, thus maintaining conformity to established norms \citep{socialcontrol}.}, from the perspective of traffic psychologists \citep{zajonc1966social, wilde1980immediate}. Generally, human driving behavior is compounded by human drivers' \textbf{social interactions} and their \textbf{physical interactions} with the scene. 

\begin{itemize}
	\item \textbf{Social Interactions.} When driving on the road, humans often interact with other surrounding drivers \textit{socially} via implicit and/or explicit communications. For example, a courteous driver (denoted as driver $A$) on the main road would actively give way to another vehicle merging at highway on-ramps (denoted as driver $B$) to avoid potential conflicts, and interactively, driver $B$ can understand driver $A$'s intentions.
	\item \textbf{Physical Interactions.} Human driving behaviors depend not only on other human agents around them but also on the physical traffic scenes. This includes the static physical obstacles (e.g., parked vehicles, road boundaries) and dynamic physical cues (e.g., traffic lights and signs), which may influence human drivers' decisions and movements during interactions.
\end{itemize}
Social interactions are more intricate than physical interactions due to the continuous closed-loop feedback among human agents, and many uncertainties exist. The social interaction may only require \textbf{simple} decision-making, which directly maps human perceptions to actions without specific reasoning and planning (e.g., stimulus-response, reactive interaction, car-following). The social interaction may also require \textbf{complex} decision making, forcing human drivers to cautiously decide an action among alternatives (e.g., yield or pass) by predicting other agents' behaviors and evaluating the influence of all possible alternatives \citep{johora2018modeling}. On the other hand, human drivers can interact with each other via explicit communications, such as using hand gestures and flashing lights. However, explicit communication options are not always available or the most efficient in practice. In many cases, human drivers prefer to use implicit rather than explicit communications to complete their driving tasks in interactive traffic scenarios \citep{lee2021road}. Therefore, this tutorial will mainly discuss the \textbf{complex, implicit social interactions} among human drivers in measurement approaches, modeling methods, and potential challenges.

\section{Definitions of Social Interactions in Road Traffic}
\label{subsec:definition}
\subsection{Interactions in Road Traffic} 
\label{subsubsec:interaction definition}
\textbf{\textsf{What is interaction?}} Interaction is a common term that can have many definitions in different disciplines. In the context of transportation, \citet{markkula2020defining} proposed a unified definition of interaction among all types of road users. In this survey, we follow this unified definition to describe \textbf{inter-vehicle interactions} as
\begin{displayquote}
	`\textit{A situation where the behavior of at least two road users can be interpreted as being influenced by the possibility that they are both intending to occupy the same region of space at the same time in the near future.}'
\end{displayquote}
This definition provides clear criteria for recognizing whether a traffic scenario is interactive. This definition implies that interaction should consist of at least three fundamental elements: (i) there are two or more agents involved, (ii) these agents are influencing each other, and (iii) there are potential spatiotemporal conflicts among agents.  For example, two human drivers on different road arms at a \textit{signalized} urban intersection usually do \textsc{not} influence each other. The two drivers should not be recognized as interactive since the traffic light regulartes their behaviors: one passes first with green light, and the other keeps static with red light.

\subsection{Social Interactions in Road Traffic} 
\textbf{\textsf{What is social interaction?}}  Social interaction has various definitions across psychology, behavioral science, and robotics. In general, social interaction is a behavior that tries to affect or account for each other's \textit{subjective experiences}\footnote{A subjective experience is produced by the individual human mind and refers to the emotional and cognitive impact of a human experience \citep{ledoux2018subjective}.} or \textit{intentions} \citep{rummel1976understanding}. In road traffic conditions, the definition of interactions among vehicles in Section~\ref{subsubsec:interaction definition} proposed by \citet{markkula2020defining} provides information about \textit{who} will be involved and \textit{when} interaction will occur. However, this definition cannot interpret the underlying dynamic process of interactions, such as \textit{how} one agent should consider the effects of other agents' actions and reactions. Toward this point, traffic psychologists \citep{wilde1976social, wilde1980immediate} conceptually hold that the social interaction process in natural traffic possesses certain characteristics such as the tendencies of social habits and values, social expectations, and social interaction patterns.
In this paper we provide a quantifiable definition of \textbf{\textsf{social interaction in road traffic}} as:  
\begin{displayquote}
	\textit{` ... a dynamic sequence of {\color{olive}acts that mutually consider the actions and reactions of individuals} through an {\color{purple}information exchange process} between two or more agents to {\color{magenta}maximize benefits and minimize costs}.'}
\end{displayquote}
In this way, social interaction possesses the three essential attributes corresponding to: \textbf{\color{olive} Dynamics} (closed-loop feedback among multiple agents), \textbf{\color{purple} measurement} (information exchange), and \textbf{\color{magenta} decision} (utility maximization).
\begin{itemize}
	\item \textbf{\color{olive} Dynamics.} Every road user considers its neighbors' actions and future reactions to social traffic, forming a continuous multi-agent closed-loop feedback system. In this system, every road user contributes to the aggregated dynamics of the traffic system and is affected by that aggregated dynamics. 
	
	\item \textbf{\color{purple} Measurement.} Road users may have different social driving characteristics (e.g., intentions, driving styles, driving preferences), leading to various actions and reactions. For efficient and safe social interaction, every road user needs to deliver their social cues and identify others' social cues, forming an information exchange process.
	
	\item \textbf{\color{magenta} Decision.} Based on the dynamics and measurement, human drivers involved in the interaction are rationally seeking to maximize their utility.
\end{itemize}
Such a unified definition of social interaction provides a computational framework that connects the fields of psychology and robotics. 

\section{From Inter-Human to Human-AV Interactions}

\paragraph{Inter-Human Social Interactions} Humans are natural social communicators; human drivers negotiate with other agents safely and efficiently, forming an interaction-intensive and multi-agent system. In general, human driving behaviors are dominated by two types of norms: legal and social. Traffic rules form the legal norms, and humans' social factors form the social norms. In real traffic, human drivers do not always act with formal behaviors (i.e., legal norms) by strictly and stereotypedly following traffic laws (e.g., keeping under the speed limit on highways).  On the contrary, human drivers will usually drive according to implicit social norms and rules that facilitate efficient and safe behavior on the road \citep{muller2016social}. Existing research also reveals that acting according to the informal behaviors (i.e., social norms) can make behavior recognizable and predictable for other human agents, thus decreasing the interaction uncertainty and facilitating every agent’s decision-making \citep{wilde1980immediate,havarneanu2012norms}. As a result, understanding and inferring other humans' driving behaviors by pure legal norms might be ineffective because:

\begin{figure}[t]
	\centering
	\includegraphics[width=0.7\linewidth]{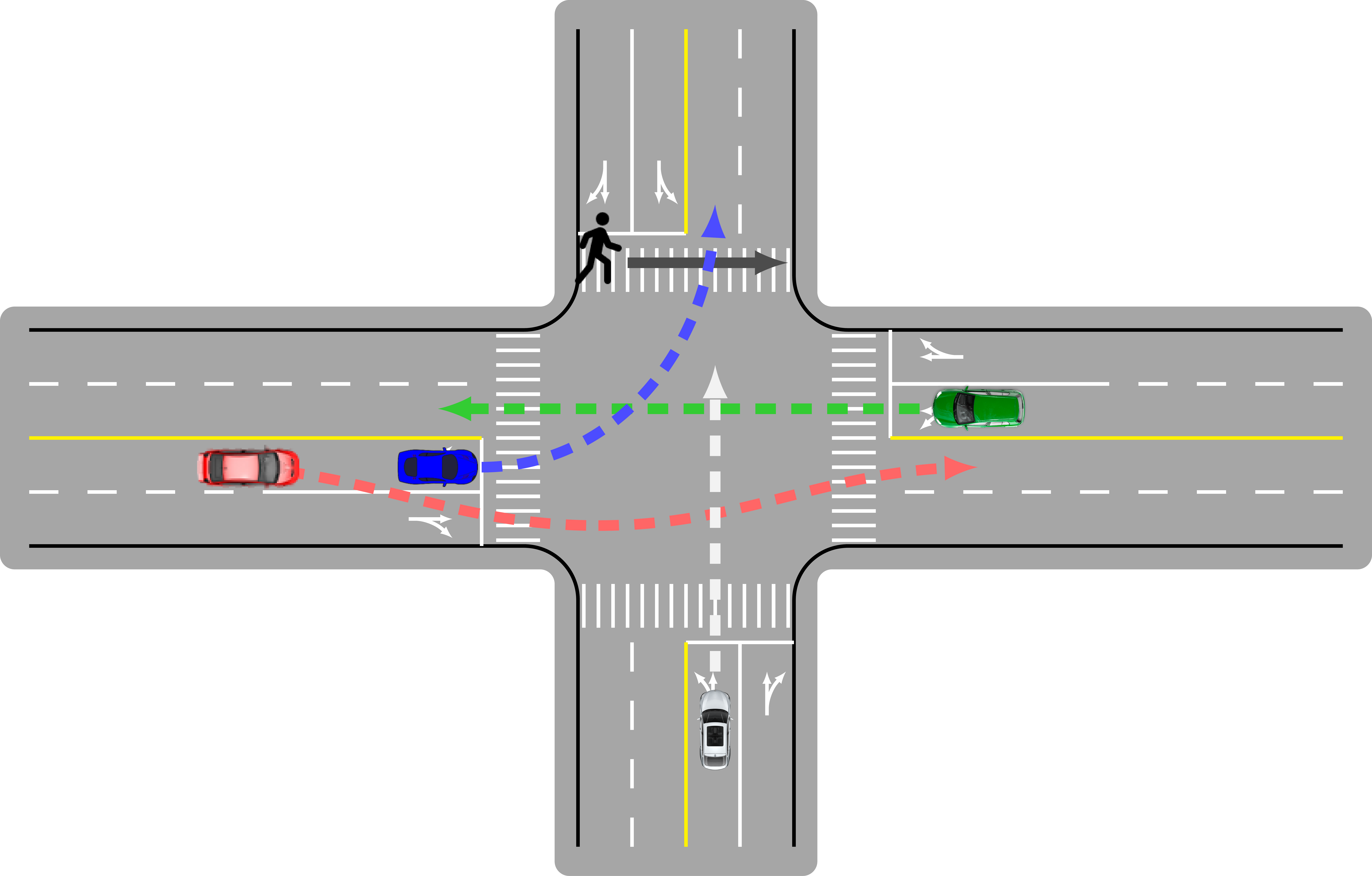}
	\caption{An example of interaction among human agents in uncontrolled traffic scenarios. The leading vehicle (blue) is yielding to the upcoming vehicle (green) and the pedestrian (black) crossing the road.}
	\label{fig:intersection}
\end{figure}

\begin{itemize}
	\item \textbf{traffic rules do not always specify driving behavior.} For example, when a driver intends to change lanes in congested traffic, the traffic laws only forbid collisions but do not specifically describe how the driver should cooperate or compete with others to create gaps. Social norms usually dominate such interaction behaviors.
	\item \textbf{human drivers do not strictly obey traffic rules.} Fig. \ref{fig:intersection} illustrates an intersection scenario that frequently occurs in real life. An experienced driver (red) intends to pass the intersection, but its leading vehicle is waiting to turn left. The driver could overtake the leading vehicle by crossing the solid white line and passing through from the right side to save travel time. Though slightly violating the traffic rules, such behaviors improve traffic flow efficiency.
\end{itemize}
Hence, equipping Autonomous Vehicles (AVs) with an understanding of the collective dynamics of human-human interactions may allow them to make informed and socially compatible decisions in human environments. 

\begin{figure}[t]
	\centering
	\includegraphics[width=0.8\linewidth]{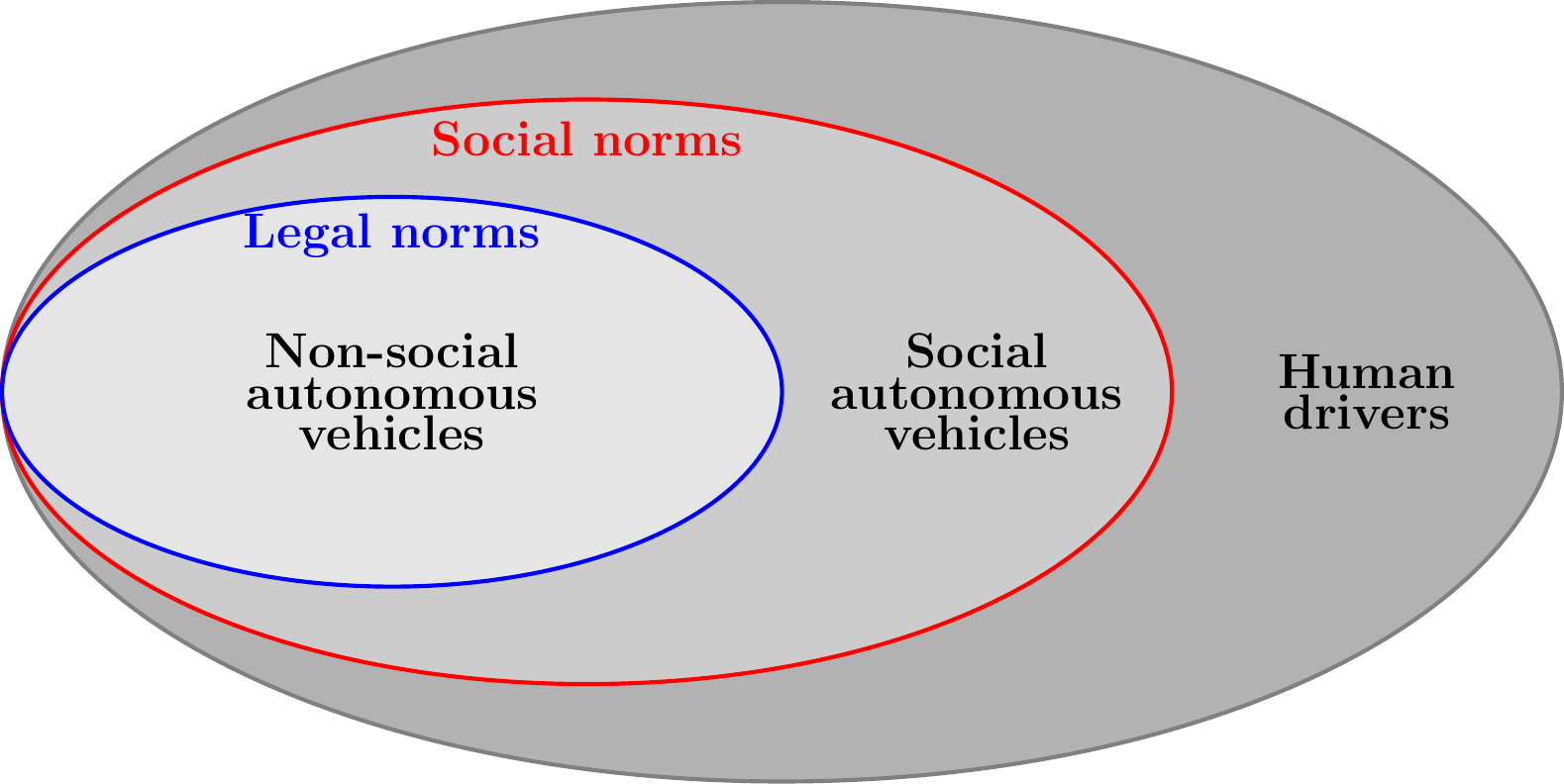}
	\caption{Illustration of the differences between human drivers, social autonomous vehicles, and non-social autonomous vehicles from the perspective of social and legal norms.}
	\label{fig:social and legal norms}
\end{figure}

\paragraph{Social Behaviors for Autonomous Vehicles}
As moving intelligence-embodied agents, autonomous vehicles also need to interact with human agents and will become part of a complex socio-technical system \citep{muller2016social}.  In such a safety-critical system, AVs should blend seamlessly into roads populated with human drivers and be socially compatible with reaching human-level interaction performance. However, a big gap exists between norms followed by human drivers and autonomous vehicles, as illustrated by Figure~\ref{fig:social and legal norms}. Autonomous vehicles strictly following legal norms might be unable to deal with highly-interactive scenarios and confuse other human drivers following the social norms. For example, an AV strictly and stereotypedly follows the 3-second law before a stop sign (could be viewed as the \textit{legal norms}) would deliver confusing social cues to other human agents: `Why the vehicle does not move ahead?' To communicate effectively and efficiently, AVs will need to mimic, or ideally improve upon, human-like driving, which requires them to:
\begin{itemize}
	\item \textbf{understand and adapt others' social and motion cues\footnote{Social cues refer to the clues to social characteristics, such as intention, driving styles, driving preferences, etc.}.} This treats AVs as information receivers, which keeps \textit{themselves} functionally safe and efficient. For example, failing to recognize the aggressiveness level of other drivers would make the AV unsafe or too conservative.
	\item \textbf{deliver recognizable, informative social and motion cues.} This treats AVs as information senders, making AVs' behaviors perceivable and understandable to \textit{other human drivers}, allowing them to make safe and efficient maneuvers. For example, an AV hesitating between yield and pass would confuse other road users, resulting in accidents or traffic jams.
\end{itemize}
It should be emphasized that we are \textit{not} claiming that AVs should violate traffic regulation in order to behave like a human driver or be socially compatible. We believe that learning and understanding the social norms followed by human drivers could benefit efficient and safe interactions.

\begin{figure}[t]
	\centering
	\includegraphics[width=0.9\linewidth]{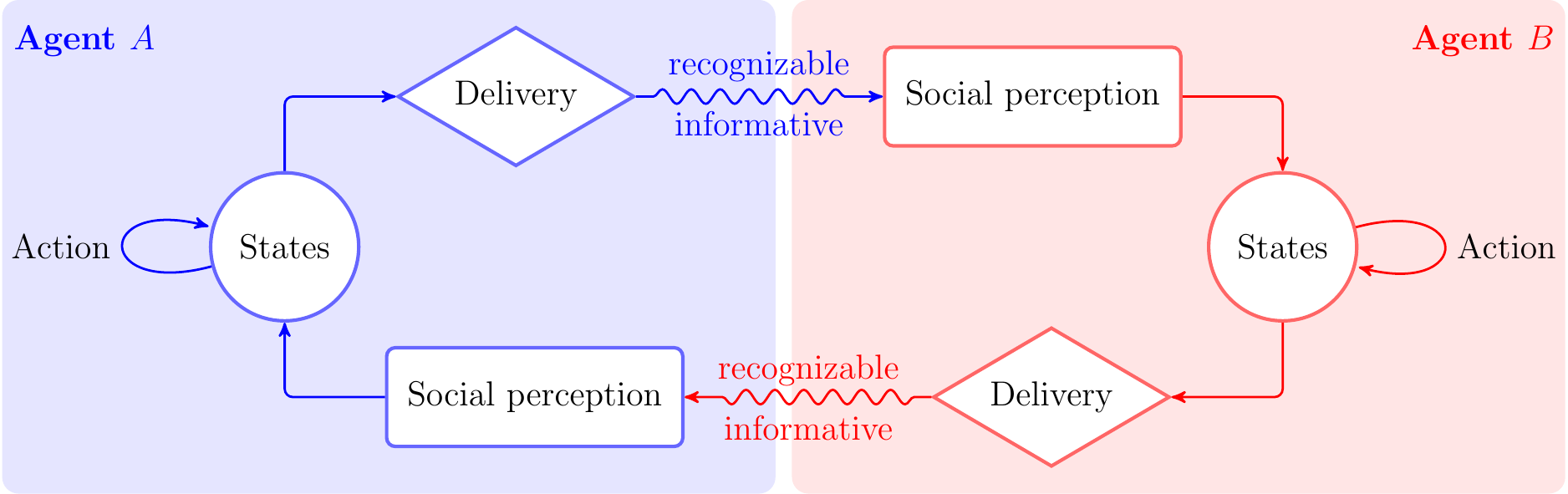}
	\caption{Illustration of the closed-loop formalism of interaction between two agents (Agents $A$ and $B$), which also generalizes to multi-agent systems.}
	\label{fig:receiver_deliver}
\end{figure}

Figure~\ref{fig:receiver_deliver} illustrates the dynamic communication procedure between two agents (human drivers and/or AVs), each of which plays two roles in the information exchange process: information \textit{sender} and \textit{receiver}. For instance, Agent $A$ would act as an information sender to `tell' Agent $B$ about its intents. Meanwhile, Agent $B$ should perceive and understand the information delivered by Agent $A$ (i.e., perception) and then take some actions to respond or adapt to Agent $A$ by delivering recognizable and helpful information. 

Endowing AVs with the human social capability to enhance interaction performance in complex traffic scenarios has shown significant progress. For example, human social preferences (e.g., altruistic, prosocial, egoistic, and competitive) and the levels of cooperation while interacting with an AV are quantitatively evaluated using computational cognitive models \citep{muller2016social,toghi2021social,toghi2021altruistic}.

\section{Paper's Scope and Framework}
This paper aims to comprehensively review the interactions among on-road vehicles (human-driven vehicles and/or AVs) toward socially-compatible self-driving cars. The interactions with other types of road users (e.g., pedestrians and cyclists) are out of the scope of this paper. We suggest readers refer to other literature reviews for autonomous vehicle-pedestrian interactions \citep{rasouli2019autonomous}, driver-cyclist interactions \citep{rubie2020influences,bella2017interaction}, and pedestrian-pedestrian interactions \citep{rudenko2020human}. 

A few literature reviews exist on the interactions between human drivers except literature \citet{mozaffari2020deep, di2021survey, gilles2022uncertainty}, which is limited to AI-guided learning approaches or specific behavior prediction tasks. However, existing works present many approaches to modeling interactions far beyond the content reviewed in \citet{mozaffari2020deep, gilles2022uncertainty}. Although literature review of \citet{di2021survey} tabled and summarized some existing works, they did not provide the behind ideas and principles of approaches to modeling interaction among drivers. To bridge the gap, we will review a wide variety of state-of-the-art works with keywords `social-aware decision-making', `interaction-aware', `cooperative decision/policy', `multi-vehicle interactions' in the ground vehicles and transportation, robotics, and their references and citations. The related approaches range from optimization theory, deep learning, and graph-based models to social fields and behavioral and cognitive sciences.

Section~\ref{sec:measurement_evaluation} discusses the essential definitions and basic ideas of social interactions in road traffic. Section~\ref{sec:modeling and learning} discusses the approaches to modeling and learning social interactions among human drivers. Section~\ref{sec:discussion}
provides some new directions, critical challenges, and opening questions for future, followed by conclusions in Section~\ref{sec:conclusion}.

\chapter{Interactions in Road Traffic - When? Who? How?}
\label{sec:measurement_evaluation}
In the last section, we build up the precise definition of social interactions among human drivers. Before reviewing existing works, we will first address two fundamental questions: `\textbf{when does interaction occur, and who is involved?}' and `\textbf{how to quantify (social) interaction?}'

\section{When does Interaction Occur? Who is Involved?}
\label{sec:when}
Before quantifying social interactions, we first need to figure out `\textit{when does interaction occur?}' or `\textit{whether interactions are happening between human drivers?}' in specific scenarios. A related question is `\textit{who is involved with interactions?}' 

In real traffic, road users do not always have rich interactions. For example, an individual pedestrian moving on the sidewalk usually does not impact other pedestrians except for a task with richer social interaction patterns, such as sports \citep{makansi2021you}. Similarly, we argue that rich social interactions between drivers may \textbf{not} always occur. Human driver primarily drives alone and reacts to the physical environment but do not \textit{directly} interact with other road users in most driving tasks, such as the lane-keeping behavior on highways and the protected left-turn behavior at signalized urban intersections. There are three commonly-used approaches to determine when interaction occurs and who is involved, which will be discussed as follows.

\subsection{Potential Conflicts Checking}
\label{sec:potential conflicts}

One straightforward way to determine if one human driver would interact with the other (directly or indirectly) is to check whether their near-future paths conflict. The interaction occurs if the paths conflict and do not occur otherwise. This checking method is consistent with the definition commonly-used in transportation literature \citep{markkula2020defining}, known as conflict points (see pages 14-15 in \citet{preston2011minnesota}) ---  `the locations in or on the approaches to an area where vehicles paths merge, diverge, or cross.' \citet{wang2021hierarchical, hu2020scenario} simplified the interaction scene by assuming that only vehicles with potential conflicts can interact with each other, which is aligned with human intuition during daily driving. The potential conflicts can be assessed from multiple moving objects' predicted future motions and intentions. 

Human drivers can utilize relevant information on road geometries and traffic regulations to check the potential conflicts with others. When drivers enter an intersection with clear traffic regulations, they can recognize conflict points by checking the crossing of their virtual reference lines to others. Besides, human drivers also use the deictics and their social inferences of others' intentions and motions to identify potential conflict points. For example, when a driver (denoted as \textit{A}) notices its adjacent vehicle (denoted as \textit{B}) with a flashing light (i.e., deictics) \textit{or} probingly approaching to cut into the driver A's front gap (i.e., social inference), driver \textit{A} could recognize driver \textit{B}'s lane-changing intents, and conflicts occur.

\subsection{Region of Interest Setting}
\label{sec:RoI}

In practice, another way to determine when and where the interaction will occur is to set a specific Region of Interest (RoI) in the environment. Interactions exist between any pair of agents occupying the RoI at the same time and disappear once any one of the agents moves outside of the RoI. Setting RoI is usually application-oriented, which can be designed according to the two approaches:
\begin{itemize}
	\item \textbf{Scenario-centric} --- Fixing the RoI on the map and treating all human drivers in this region as interactive agents. This method is usually used to predict and analyze multiagent driving behaviors in a specific traffic region, such as urban intersections and roundabouts. In these cases, researchers fixed an RoI covering such scenarios on the map and assumed that all drivers occupying the RoI would interact with each other  \citep{zhan2019interaction, bock2020ind, krajewski2020round, liu2017distributed}. 
	
	\item \textbf{Agent-centric} --- Attaching the RoI with one interested agent (i.e., ego agent). This method is usually used to investigate the ego agent's interaction behavior with its surrounding agents, such as lane change behaviors on highways. There are multiple choices on the shape of the RoI. For example, when investigating lane-changing interaction behavior on highways, a rectangular region is usually  attached to the ego vehicle and set as the RoI \citep{banijamali2021prediction, zhang2021spatiotemporal, deo2018would, li2019grip}.
\end{itemize}
Note that some works also mixed these two methods. For example, after setting the RoI in a highway segment, \cite{hu2018probabilistic} further calculated the distance between agents to determine the presence of interactions.

The above-mentioned RoI requires hand-crafted rules, and the associated evaluation performance might be sensitive to the configurations of RoI \citep{li2019grip}. In general, a larger RoI would get more agents involved and probably overestimate interactions, while a small one would get fewer agents involved and underestimate interactions. In order to overcome the drawbacks, actively selecting the interactive agents according to the driving tasks is an alternative.

\subsection{Task-oriented Agent Selection}
\label{sec:taskoriented selection}

Humans will selectively determine \textit{which} agents should be paid more attention to and \textit{when} attention should be paid according to specific driving tasks \citep{niv2019learning, wang2021social}. Inspired by this fact, researchers empirically select interactive agents for specific tasks according to the problems and their domain knowledge of the corresponding interaction process. For example, for the left lane-change task, researchers assume that the ego vehicle only interacts with the leading vehicle on the current lanes and the leading and following vehicles on the left target lanes. This assumption matches human driving experience and can simplify the interaction problem by only paying attention to task-involved agents. However, it requires hand-crafted rules with specific domain knowledge and may fail to capture the individuals' differences in how to pay attentions.

\subsection{Summary}
The three methods above have been widely used but might overestimate or underestimate the interaction among human drivers. The potential conflict-based approach would underestimate the interaction since it is defined according to the potential conflict, such as between the agents' right of way. However, some social interactions in driving are derived not only from potential conflicts but from cohesion (see Section~\ref{sec:social_preference}). For example, an aggressive driver would imitate his/her leading vehicle's behavior to take an opportunistic action to proceed forward when facing a yellow light at intersections. It should be emphasized that the potential conflict-based approach will not treat this as an interaction. The RoI approach might overestimate the interaction if assuming that all the agents occupying RoI have interactions with each other. In natural traffic, not all agents in the RoI are activated in the net of interactions; inversely, human drivers might only have a direct interaction with some agents \citep{niv2019learning}. Besides, the size of the RoI is also hard to generally configure since it is relevant to tasks and environments. Task-oriented agent selection is an ideal approach to imitating how human drivers interact with each other. Nevertheless, building such types of models is challenging since the knowledge of the interaction process is rarely known a priori. Moreover, the selective attention of humans in decision-making is dynamic and stochastic \citep{radulescu2019holistic}, which requires temporally-adaptable models.

\begin{figure}[t]
	\centering
	\includegraphics[width=0.6\linewidth]{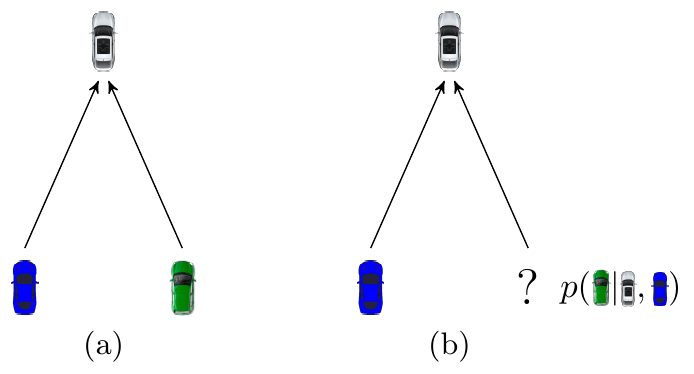}
	\caption{Direct and indirect influence among human drivers (in Figure \ref{fig:intersection}) under probabilistic graph models.}
	\label{fig:direct_indirect}
\end{figure}

\paragraph{Direct and Indirect Interactions} The above approaches in Sections \ref{sec:potential conflicts} - \ref{sec:taskoriented selection} are widely-used since they simplified the interaction among agents as the binary event (i.e., absence and presence), thus enabling the building of models using the off-the-shelf methods. However, they may not reveal that human drivers could \textbf{directly} or \textbf{indirectly} influence them. In a multiagent system, human driving behavior could be influenced \textit{indirectly} by others over space and time. Figure~\ref{fig:intersection} shows an example in which the green car and the blue car directly influence the white car's behavior, which can be formulated as a probabilistic graph model (see \citet{jordan2003introduction} and Figure~\ref{fig:direct_indirect}(a)). Thus, the green car's behavior is dependent on the blue car's decision, given the observation of the white car's behavior from a conditional probabilistic perspective, as shown in Figure~\ref{fig:direct_indirect}(b). In other words, the blue car's behavior would \textit{indirectly} influence the green car's behavior. 

\section{How to Quantify Social Interactions?}
\label{sec:measurement and evaluation}
\begin{figure}[t]
	\centering
	\includegraphics[width=\linewidth]{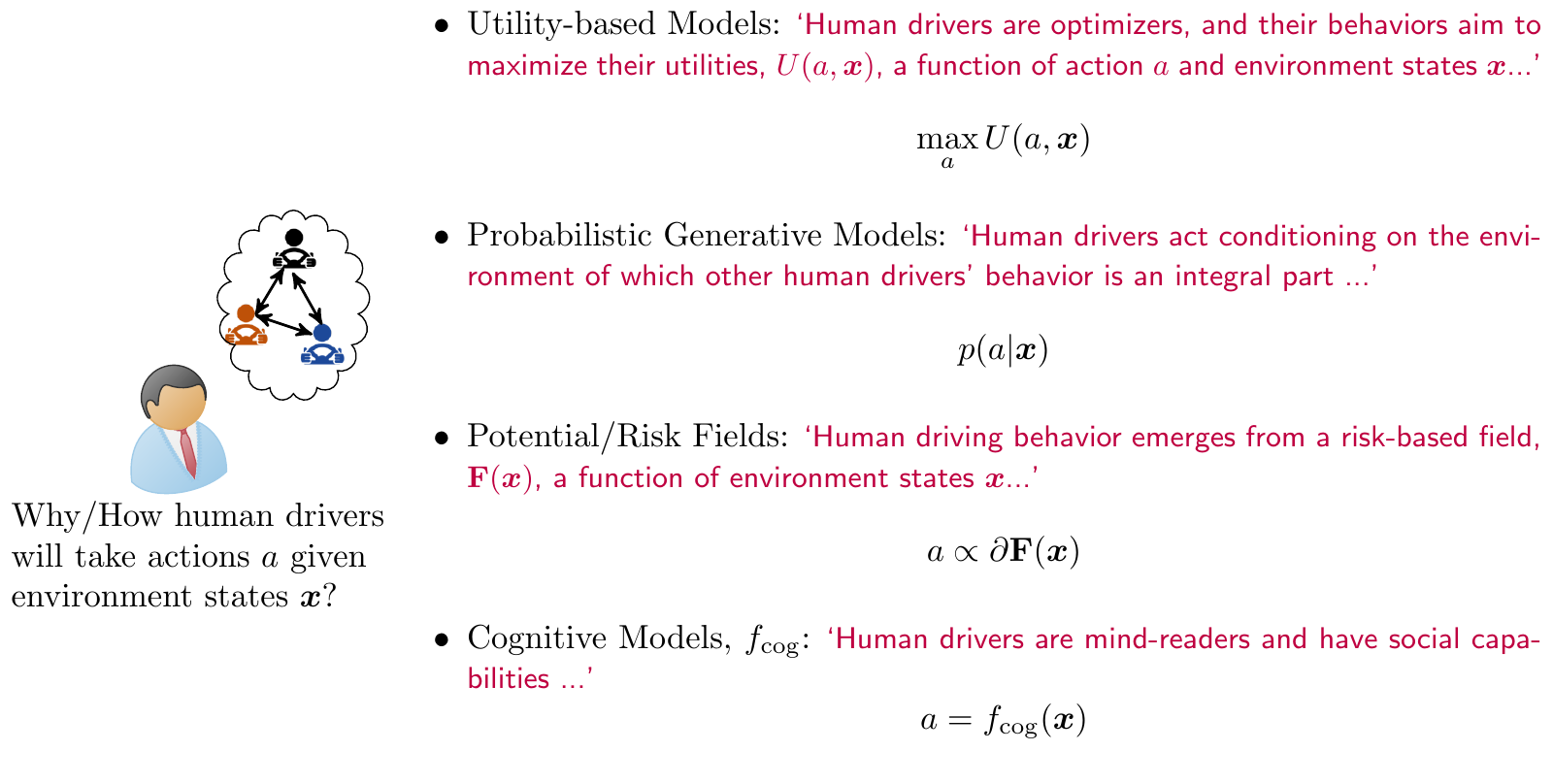}
	\caption{Different quantification approaches to explicitly modeling interactions among agents based on four hypotheses.}
	\label{fig:methodclassification}
\end{figure}

The definition of interactions in Section~\ref{subsec:definition} implies that checking the \textit{influences} of human drivers on each other can identify the absence and presence of human interactions. With interactions among human drivers, one critical task is to figure out `\textit{How to quantify these interactions while leveraging social factors?}' This section will discuss some commonly-used quantification approaches, which can be generally classified into two types: explicit model-based approaches and implicit data-driven approaches. Then, we will also discuss the quantitative approaches to evaluate social driving influences on human drivers.

\subsection{Model-Based Approaches}
The most commonly-used approach to explicitly quantify interactions is constructing an interaction model in which some parameters can be estimated from sensory data to quantify the intensity of social interactions among agents. Researchers have developed many interaction models using sensory information based on different hypotheses, as shown in Figure~\ref{fig:methodclassification}. For sensory data as model inputs, the physical distance-related measures are explicitly interpretable for designers, thereby gaining more attraction from researchers. In applications, it is intuitive to assume that the interaction intensity is related to the agents' relative distance and its variants (e.g., relative speed and acceleration) in traffic scenes. For example, human drivers with a closer distance would be intuitively treated as having heavier interactive influences on each other. 

\paragraph{Rational Utility-based Models} Human driving behaviors or actions are the (near) optimal outcomes that maximize certain utilities to the environment. At this point, researchers would formulate the interactions among human drivers as optimization problems by integrating the physical distance-related information into the objective/cost function, which can be solved using the off-the-shelf dynamic and linear programming algorithms. Usually, the cost function is manually crafted according to the prior domain knowledge of traffic regulation and driving tasks. For example, researchers treated human drivers' lane-change maneuvers as an optimization problem of minimizing the lateral path tracking errors (lateral control)  with vehicle dynamics constraints while keeping the desired speed (longitudinal control). Usually, the utility-based model could reach an expected performance in similar scenarios with careful parameter tuning but have low generalizability in the unseen scenarios. The typically-used models are optimal swarms, game-theoretic models, imitation learning, and the Markov decision process. For more details, we refer readers to Section~\ref{subsec:optimal}.

\paragraph{Probabilistic Generative Models}
As claimed by \citet{wilde1976social}, the description of a social interaction situation is a condition where an individual driver's behavior is determined by his organismic features and the environment of which the other drivers' behavior is an integral part and vice versa.
From the perspective of conditional probability, the interactive influences between human drivers can be interpreted as 
`\textit{How would one driver probably take a specific action after perceiving other surrounding human drivers' states?}' This question can be formulated by a probabilistic conditional distribution or conditional behavior prediction \citep{tolstaya2021identifying, chai2019multipath, tang2022interventional}:

\begin{equation}\label{eq:pgm}
	p(a_{t+1}^{i}|\boldsymbol{x}_{t}^{1:N}, a_{t}^{1:N}),
\end{equation}
where $a^{i}_{t+1}$ represents the action of human driver $i$ at time step $t+1$, and $\boldsymbol{x}_{t}^{1:N}$ represent the states of all surrounding vehicles $1, \dots, i-1, i+1, \dots, N$. Further, the interactive influence should also consider the \textit{future} actions of surrounding vehicles, thus Equation (\ref{eq:pgm}) can be modified as

\begin{equation}
	p(a_{t+1}^{i}|\boldsymbol{x}_{t}^{1:N}, a_{t}^{1:N}, a_{t+1}^{\neg i}),
\end{equation} 
where $a_{t+1}^{\neg i}:=a_{t+1}^{\{1:N\}\setminus \{i\}}$.
This concept is the basis of Bayesian networks, a commonly used model, which will be detailed in Section~\ref{subsubsec:bayesian networks}. Besides, the agent interactivity can be quantified by a surprising interaction \citep{tolstaya2021identifying} -- one in which one agent (denoted as agent $B$) experiences a change in their behavior due to the other agent's (denoted as agent $A$) observed trajectory, computed by 
\begin{equation}
	\Delta_{\mathrm{LL}}:= \log p(\boldsymbol{x}^{B}|\boldsymbol{x}^{A}) - \log p(\boldsymbol{x}^{B}).
\end{equation} 
A large change in the log-likelihood indicates that the agent $B$ is influenced significantly by agent $A$'s action. If the agent $B$'s trajectory $\boldsymbol{x}^{B}$ is \textit{more} likely given the agent $A$'s trajectory $\boldsymbol{x}^{A}$, then $\Delta_{\mathrm{LL}}>0$; if it is \textit{less} likely, then $\Delta_{\mathrm{LL}}<0$; if it has no change, then $\Delta_{\mathrm{LL}}=0$. This idea makes it tractable to use most off-the-shelf similarity measurement approaches in information theory such as KL-divergence. 

On the other hand, the interaction can also be further viewed as a (latent) probabilistic generative process \citep{banijamali2021prediction} or conditional probabilistic model \citep{kafer2010recognition}. For example, \citet{anderson2020low} proposed a probabilistic graphical model to capture the interaction between the future states of the lead vehicle and the history states of dynamic systems (i.e., the lead vehicle and the rear vehicle). However, it fails to consider the influence of the merging vehicle on the interactions between the lead and rear vehicles. \citet{gonzalez2017interaction} proposed an interaction-aware probabilistic driver model to capture the driver's interaction preferences and human drivers will perform a maneuver at the current time step given their prediction for the surrounding driver's behavior. The interaction preferences are then formulated by a combination of weighted features (i.e., navigational and risk features) under the framework of inverse optimization.

\paragraph{Potential/Risk Fields} The third typical model is the potential/risk field \citep{wang2015driving, tan2021risk} based on the hypothesis that human driving behavior emerges from a risk-based field \citep{kolekar2020human}. Using potential functions \citep{bahram2016combined} to model the interactions between agents has been investigated widely in human-robot \citep{khansari2017learning, najafi2020using} and multi-vehicle interactions \citep{wolf2008artificial, kim2017local, kim2018trajectory, yi2020using, kruger2020interaction}. The physical distance-related metrics allow efficiently formulating interactions with certain learnable and interpretable functions, called potential functions, that can embed domain knowledge of traffic regulations and driving scene context. On the other hand, the derivatives of potential functions to the coordinates systems (e.g., $x$ and $y$ directions) result in the scaled virtual forces that `push' or `pull' the vehicle to minimize the local planning cost of the vehicle while interacting with its surrounding human drivers. Researchers also design energy functions to capture the inter-vehicle interactions based on the relative distance (usually the minimum value or the closest point of two-vehicle trajectories) between surrounding vehicles and the ego vehicle \citep{deo2018would}. However, the relative distance-based measurement does not always correctly capture the interactions between human drivers. Human drivers with a closer distance might have very weak or no interactions when physical constraints exist between agents such as highway barriers or dividers for opposite lanes. More potential fields-based models and approaches are discussed in Section~\ref{subsec:social_force_based_model}.

\paragraph{Cognitive Models} Researchers used the relative distance to characterize driver styles and reveal the interaction process between multiple agents, for example, using the theory of mind \citep{tian2021learning} and the information accumulation measures \citep{zgonnikov2020should}. Other types of interaction models from the behavioral science and psychology perspective have also been developed to mimic human driving behaviors. For more details, refer to see Section~\ref{subsec:cognition}.

\subsection{Data-driven Approaches}
Unlike the models above that directly utilize explicit sensory information to characterize the interaction among human drivers, another approach uses the encoded implicit information to quantify interactions. Such implicit information is usually in the form of low dimensional scalars or vectors, also called embeddings\footnote{An embedding is a \textit{function} mapping every graph node and relation type to a hidden representation that is supposed to contain the necessary information about the nodes and relations.} in the context of graphical models. In what follows, we will mainly introduce three approaches to encode interactions. 

\paragraph{Deep Neural Networks} Neural networks represent functions that map multiple sensory information to low dimensional vectors through a sequence of basic layers (e.g., convolutional and recurrent, see Section~\ref{subsubsec:deepmodules}), for example, in the structure of autoencoder and Generative Adversarial Networks (GANs). Besides, the attention mechanism can also be integrated into networks to model interactions among human drivers (see Section~\ref{subsubsec:interaction_encoding}). For more details about deep learning-based approaches to modeling interactions, see Section~\ref{subsec:deeplearning}.

\paragraph{Graphic Neural Networks with Social Pooling} Graphic neural networks (GNNs) share some commons with regular deep learning -- a neural network with multiple layers \citep{hutson2017ai} by embedding the structural information as model inputs. Information pooling is a flexible tool to abstract the relationship among agents over spatial and temporal spaces into low-dimension quantifiable embeddings (e.g., a normalized continuous vector) in light of the advantages of deep neural networks and lots of open sources for programming. The embeddings can be time-dependent to capture the temporal information of the nodes and edges in evolving graphs \citep{kazemi2020representation}. Thus, they can characterize the interaction intensity among human drivers through aggregation operations such as averaging aggregation \citep{makansi2021you}, weighted aggregation, and evolving messages over graphs (or graph message passing) \citep{li2020evolvegraph, girase2021loki, mohamed2020social, wang2022transferable}. Moreover, the pooling operation can embed information into low-dimension latent states over time and spatial dimensions independently \citep{deo2018convolutional, li2020evolvegraph, kosaraju2019social, wang2021hierarchical} or simultaneously \citep{yuan2021agentformer, cao2021spectral} using different structures of neural networks. The former usually first applies a temporal model (e.g., LSTM) to independently summarize the features over time for each human agent and then formulates the interaction of the summarized features with a social model such as convolutional social pooling \citep{deo2018convolutional}. The encoded embeddings capable of modeling vehicle interactions can be obtained by training GANs \citep{roy2019vehicle, fei2020conditional} and autoencoders \citep{ju2020interaction}. Although some of the above works show promising results on standard benchmarks, it is still unclear what information these methods should use to predict the future states and how to interpret these embeddings with physical meanings. Another way to quantify the interaction relationship is to use the learnable weights of certain graph edges, also termed weighted graph edges \citep{cao2021spectral}, with the support of sequential observations. A salient feature of these `encoders' is that they make little to no reference to the effectiveness and interpretations of the learned encoded information. For more detailed discussions, see Section~\ref{subsubsec:interaction_encoding} -- Section~\ref{subsubsec:GNN}.

\paragraph{Topological Models} Another idea of encoding interactions among human drivers is to map them into a compact representation of dual algebraic and geometric nature using the formalism of topological braids \citep{tan2019topological,mavrogiannis2020implicit,mavrogiannis2021analyzing}. Such a compact topological representation can help understand the complex interaction behaviors over any environment with any number of human agents. For more details, the readers are referred to Section~\ref{subsubsec:topology}.

\subsection{Modeling Influences during Social Driving} 
\label{sec:social_preference}
Humans can drive remarkably well by leveraging the explicitly sensory information in traffic scenes and the implicit social inferences in the behavior of other human drivers to make safe and socially-acceptable maneuvers. It is human beings' nature to endow the capability of information intake and behavior anticipations with the factors in social preferences, social imitation, and social inferences, which is central to socially-compatible driving behaviors. A quantitative evaluation of these social factors requires computational cognitive science and techniques.

\paragraph{Social Value Orientation (SVO) for Driving Preferences} A human driver would have various social preferences while interacting with others. The social preferences such as the altruistic inclinations of the other human-driven vehicle can be quantitatively evaluated from computational psychology such as SVO \citep{liebrand1984effect,liebrand1988ring}. The SVO model gauges how one driver weights its rewards against the rewards of other agents, which is learnable from observed trajectories under the structure of Inverse Reinforcement Learning (IRL). The online-learned driving preference with the SVO model is then integrated into game scenarios where two or multiple vehicles cooperatively interact. The SVO concept has been widely investigated and applied to socially-compatible autonomous driving \citep{schwarting2019social,muller2016social,toghi2021social,toghi2021altruistic,zhao2021yield}.

\paragraph{Social Cohesion for Social Driving Imitation} `Human drivers follow each other like sheep,' stated by \citet{hutchinson1969evaluation}, and the effects of coaction in traffic behavior are very strong \citep{wilde1980immediate}. Thus, human-driven vehicles' behaviors are \textit{socially cohesive} -- the driver would take similar action to its surrounding drivers. For instance, if the leading vehicle slows down and takes a slight `collision avoidance' behavior, the driver of the ego vehicle would usually take a similar action by socially assuming a virtual obstacle (e.g., a cone, an animal body, a pavement pit) might exist. Inspired by the social cohesion of human drivers, \citet{landolfi2018social} developed a cohesion-augmented reward function to enable an autonomous car to socially follow other vehicles automatically by determining \textit{what aspects}, \textit{who}, and \textit{when} to follow to guarantee safety.

\paragraph{Social Perception for Situation-Awareness Enhancement} Humans can actively gather and derive additional information about the environment to create a relatively complete traffic scene, thus providing sufficient information and improving environmental awareness to take a safe and efficient maneuver. For instance, perceiving the decelerating-and-stop behavior of the adjacent vehicles, a human driver can infer a potential pedestrian crossing the road no matter if the view of the driver is occluded. This ability of humans to socially treat other human drivers as sensors have been formulated and integrated into autonomous vehicles to enhance drivers' situation awareness. For instance,  \citet{sun2019behavior} and \citet{afolabi2018people} formulate the cognitive understanding of the occlusion of the environment via a conditional distribution over a belief space.

\paragraph{Social Interactive Styles for Driving Styles} Human drivers make plans and take action by assessing and balancing the different reward terms over the future. Based on their internal model, the driving tasks, and motivations, humans may pay extra attention to different reward terms, by which different interactive styles with their surroundings are exhibited, such as aggressive, conservative, courtesy, selfishness, and irrationality \citep{wang2021social}. Thus, the interactive styles can be formulated as the weighted outcomes of different features in generating trajectories. For example, researchers quantitatively weighed these interactive styles induced by social factors as reward features. Then, they used inverse reinforcement learning (IRL) to learn the weights of such features \citep{sun2018probabilistic, sun2018courteous}  or the ranking objective functions \citep{lee2017desire} from trajectories.

\chapter{Approaches to Modeling and Learning}

\label{sec:modeling and learning}
\begin{figure}[t]
	\centering
	\includegraphics[width = 0.9\linewidth]{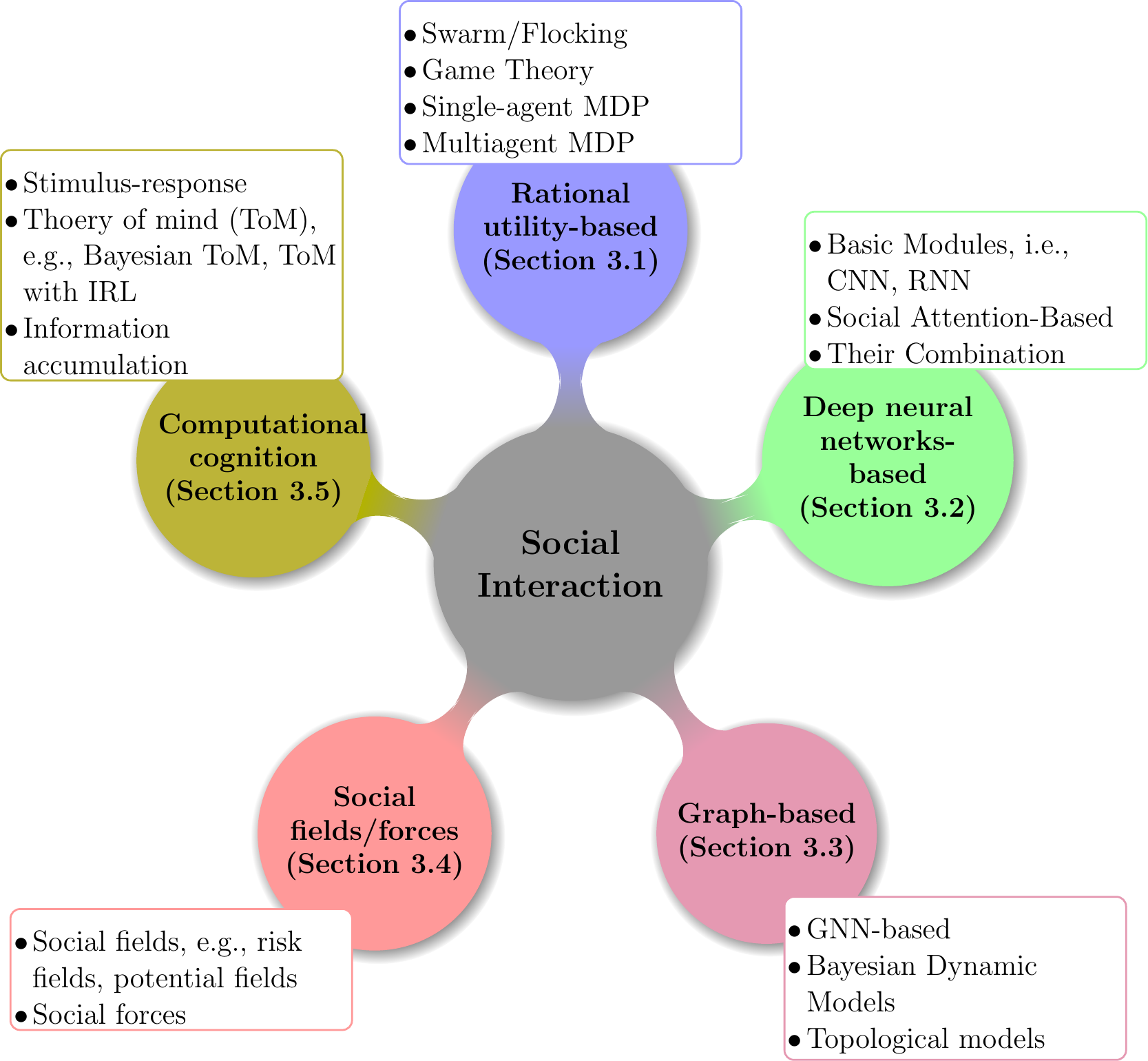}
	\caption{Summary of approaches to modeling and learning interactions among human drivers.}
	\label{fig:structure}
\end{figure}

This section mainly introduces the underlying ideas of different approaches to model and learn the interaction using the methods and quantification approaches discussed in Section~\ref{sec:measurement_evaluation}. 
Generally, among different types of quantification models reviewed in Section~\ref{sec:measurement_evaluation}, the following five types (see Figure~\ref{fig:structure}) are used more often and hence will be discussed in more details, including rational utility-based models (Section~\ref{subsec:optimal}), deep neural networks-based models (Section~\ref{subsec:deeplearning}), graph-based models (Section~\ref{subsec:graph_based_model}), social fields \& social forces (Section~\ref{subsec:social_force_based_model}), and computational cognitive models (Section~\ref{subsec:cognition}). 

\section{Rational Utility-Based Models}
\label{subsec:optimal}
The most commonly-encountered interactive scenarios in our daily traffic are car-following, merge-in/out, and lane change in urban environments and highways. Researchers formulate behaviors in these scenarios by treating the human driver as an optimal controller with an accessible objective function to achieve the predefined goal-oriented tasks. For instance, when merging at highway on-ramps, the rear vehicle's longitudinal behavior (e.g., acceleration/deceleration) to the lead vehicle on the highway can be formulated as an optimal controller \citep{wei2013autonomous} by treating the entire merging process as an explicit dynamic system \citep{anderson2020low}. However, the interaction among drivers or other human agents in natural traffic scenes has physical (e.g., kinematics and geometry) and social (e.g., intention, attention, and responsibility) constraints. \citet{luo2019gamma} developed a general agent interaction model (GAMMA) to predict human agents' behavior by treating the interactive motion as a constrained geometry optimization problem with velocity obstacles \citep{fiorini1998motion}, which is then implemented in a high-fidelity simulator (SUMMIT) to model massive mixed urban traffic \citep{luo2020simulating}. Besides,
\citet{lee2017desire} claimed that the interaction between human agents in the short future should obtain an optimal accumulated reward. Then, the authors formulated the motion prediction problem in an optimization framework of maximizing the potential future reward for a set of prediction hypotheses similar to inverse optimal control. Generally, optimization-based approaches require a specific target (e.g., desired gap and headway between vehicles) and an objective function to be optimized.

In what follows, we will discuss some models built on the idea of optimizing a specific cost function or objective function. Note that this paper did not enumerate all optimization-based approaches but selected some popular ones, including swarm/flocking-based models, game-theoretic models, imitation learning, and the Markov decision process.

\begin{figure}[t]
	\centering
	\includegraphics[width = \linewidth]{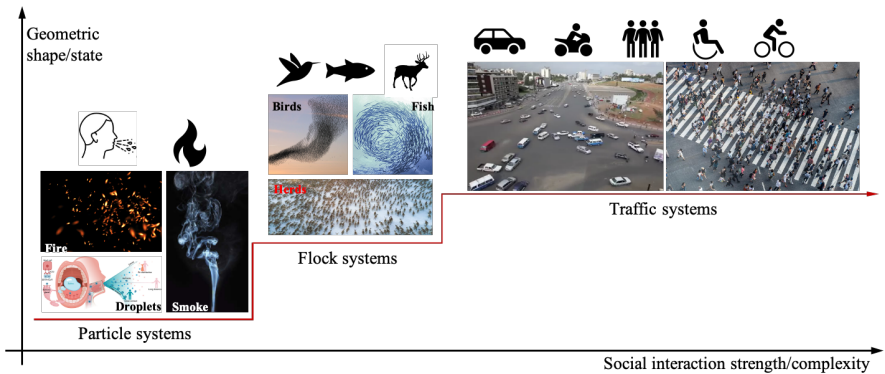}
	\caption{The natural phenomenon of swarms from physical particle systems (e.g., fire, smoke, and droplets), flocks (e.g., birds, fish, and herds) to traffic systems (e.g., vehicles and pedestrians).}
	\label{fig:swarm}
\end{figure}

\subsection{Swarm/Flocking-based Models}
\label{subsubsec:flocking}
In the real world, human driving motions in crowded traffic environments exhibit fluid-like patterns \citep{mahjourian2022occupancy, wang2020learning, zhang2019general} analogous to physical particle systems and flocks\footnote{In this paper, the term `flocks' generally refers to a group of animal objects that exhibit a general class of polarized, noncolliding, aggregate motion, as in \citet{reynolds1987flocks}.}, as shown in Figure~\ref{fig:swarm}. It is reasonable to assume that the evolved interactive behavior of animals and humans is `near optimal' to adapt to their environments and maximize their utilities. Extensive research has revealed the underlying mechanisms of motions in particle and flock systems in physics and ethology \citep{czirok2000collective, bechinger2016active, elgeti2015physics}, which has been extended to dynamic multiagent systems \citep{olfati2006flocking} and crowded traffic scenarios \citep{helbing2001traffic}. 

\paragraph{From Particle Systems to Flocks} Simple rules create complex patterns. Both particle systems and flocks are collections of large numbers of individuals, each having its behavior according to the influences of its \textit{local} neighbors in the dynamic environment.
Natural flock systems inherit specific attributes of particle systems and follow the Reynolds' Rules \citep{reynolds1987flocks}:
\begin{enumerate}
	\item Collision avoidance --- avoid collision with nearby flockmates;
	\item Velocity matching --- attempt to match velocity with nearby flockmates;
	\item Flock centering --- attempt to stay close to nearby flockmates.
\end{enumerate}
These intuitive rules have shown promising performance in cooperative swarm robot design \citep{liu2004stable}.

Although flocks are analogous to physical particle systems, they have apparent differences. Flocking behavior is generally more complex than the typical physical particle behavior. Physical particle systems view each particle as a virtual point with mass; hence individuals in physical particle systems do not have social influences on others. However, individuals in flock systems interact socially with their local neighbors based on geometrical information (e.g., shape and size), states (e.g., orientation), and physical constraints, as illustrated in Figure~\ref{fig:swarm}. 

The particle swarm optimization has been widely investigated in robots (e.g., drones and small mobile robots) but is still relatively new in cooperative autonomous driving with few works \citep{lijcklama2020control} since it requires rich prior knowledge of driving tasks and model calibration efforts. Some works described traffic crowds by directly replacing agents in particle systems with traffic participants without considering related constraints. \citet{helbing2001traffic} provides a detailed survey of the related particle-based approaches to formulate microscope vehicle traffic behaviors\footnote{It is worth mentioning that many particle-based approaches have been developed to model traffic behavior at different levels: macroscopic, mesoscopic, and microscopic \citep{helbing2001traffic}. In our paper, we mainly focus on microscopic traffic behavior.}. 

\paragraph{From Flocks to Traffic Systems} Multi-vehicle interactive behaviors, especially in complex and cluttered traffic scenarios, are one type of interactive behavior of organisms organized into flocks-like swarms. Investigating the underlying mechanisms of interactive behaviors of human drivers, a kind of flocks, can offer some insights into the possible control of cooperative groups of autonomous cars. Inspired by the Reynolds' Rules of flocks, \citet{fredette2016swarm} developed a swarm-inspired decentralized model with several ordinary differential equations and smooth functions to capture the human driving interaction on highways. The dynamic model of the $i$-th ground vehicle in the two-dimensional space is formulated as

\begin{equation}\label{eq:dynamic_model}
	\begin{split}
		\dot{\boldsymbol{q}}^{i} & = \boldsymbol{v}^{i},\\
		\ddot{\boldsymbol{q}}^{i} & = \frac{u^{i}}{m^{i}},
	\end{split}
\end{equation}
where $\boldsymbol{q}^{i} = [x^{i}, y^{i}]^{\top} \in \mathbb{R}^{2}$ is the position, $\boldsymbol{v}^{i} = [v_{x}^{i}, v_{y}^{i}]^{\top} \in \mathbb{R}^{2} $ is the velocity vector, $m^{i}$ is the vehicle mass, and  $u^{i}$ is the associated (force) control input. In order to generate a particular force ($u^{i}$) that compels the car's motion without collisions, the third rule `Flock centering' of the original Reynolds' Rules in particles and flocks needs to be modified coordinately. For example,  \citet{fredette2016swarm} claimed that the vehicles should follow the center of lanes and adapted the Reynolds' Rules to include three force terms:
\begin{enumerate}
	\item Repulsion force, $\mathbf{F}_{\mathrm{r}}$, analogous to collision avoidance;
	\item Velocity profile following force, $\mathbf{F}_{\mathrm{v}}$, analogous to velocity matching;
	\item Road/lane attraction force, $\mathbf{F}_{\mathrm{a}}$, analogous to flock centering.
\end{enumerate}
The required force applied on the vehicle would be the summation of the three forces above

\begin{equation}
	u^{i} = \sum \mathbf{F}_{\mathrm{r}} + \sum \mathbf{F}_{\mathrm{v}} + \sum \mathbf{F}_{\mathrm{a}}.
\end{equation}
The authors also added an extra force term responsible for compelling vehicles to change lanes, called lane-change force. This force term is proportional to the square of the total repulsive force and the desired velocity force acting on the vehicle. Similarly, \citet{park2019flocking} proposed a flock-based cooperative motion mechanism consisting of three types of virtual forces (i.e., cohesion, separation, and alignment) following the Reynolds' Rules. This cooperative motion mechanism aims to mimic the interaction between human drivers using the `attract-repel' type of forces where each driver seeks to be in a `comfortable' position with their neighbors. The effect of cohesion force is analogous to human drivers' social imitation (or social cohesion) behavior \citep{wilde1980immediate, landolfi2018social} (see Section~\ref{sec:social_preference}). Further, to formulate the longitudinal and lateral controls of swarm vehicles in traffic, \citet{favaro2018towards} modified the `cohesion force' as `velocity matching' and `heading selection' to guide the longitudinal and lateral motions to match the average velocity of its neighboring vehicles. Inspired by flocking behavior, \citet{hao2022flock} formulated the multi-vehicle group dynamics using potential theory from the view of physical and artificial potential fields.

The flocking mechanism realizes collision avoidance and inter-group matching behaviors.  However, driving on the road is different from flocking in free space since not all drivers share the same destination and geometry constraints \citep{iftekhar2012safety}. In traffic, heterogeneous behaviors of individuals against the group exist, such as split maneuvers (e.g., driving off the highway by diverging from the main traffic flow). Inspired by flockings,  \citet{iftekhar2012safety} develops a unified framework and a distributed cooperative autonomous algorithm to achieve commonly-performed driving tasks such as lane-keeping, lane-changing, braking and passing, turning left/right, and avoiding tailgating and collision with other vehicles. By treating each traffic agent as a particle, \citet{raj2021moving} developed a dipole field-based model and proposed a traffic rule to formulate the dynamic of motion in a crowded environment, for example, an ambulance driving through heavy traffic. Analogous to the Reynolds' rules, the authors defined three types of forces: 
\begin{enumerate}
	\item Restitution force --- attempts to bring the agent to its desired states;
	\item Collision-avoidance force --- prevents agent collisions;
	\item Dipole force --- induces/coordinates the motion of the surroundings around the ego vehicle.
\end{enumerate}

\paragraph{Relations between Reynolds' Rules and Social Forces} It would be interesting to notice that most concrete instantiations of the Reynolds' Rules in applications (e.g., swarm robots \citep{liu2004stable} and pedestrians \citep{helbing1995social} and road vehicles \citep{fredette2016swarm, park2019flocking}) are based on a dynamic model propagated by Newton’s laws of motion as in Equation (\ref{eq:dynamic_model}). Thus, this requires developing a comprehensive set of force components that reflect the Reynolds' Rules. These force components (i.e., cohesion, separation, and alignment) exactly match each part of the social force theory(see equation (\ref{eq:social_force})) that was initially proposed to model pedestrian interactions \citep{helbing1995social}. Many social force-like formulations can be regarded as instantiations of the Reynolds' Rules. Therefore, the Reynolds' Rules reveal interactions' underlying mechanisms and are more generalized than the social force-based theory. More details about the social force and related works are discussed in Section~\ref{subsec:social_force_based_model}.

Although the flocking-inspired rules successfully reveal the mechanisms of swarm interactions between animals (e.g., birds, fish, and flocks), they can only work well in the free space or the space with static obstacles, for example, birds in the sky and fish in the sea. In interactive traffic scenarios, these rules could break down due to two types of constraints. (i) The constraints of the traffic environment. The space where flocks motion has weak, even no, physical limitations, such as in the open sky and underwater. Conversely, the motion space of road vehicles has many constraints from road profiles and traffic infrastructures. For example, vehicles should drive in lanes rather than within on-road boundaries. (ii) The constraints of the agent itself. Pedestrians and flocks have a high degree of motion, such as pedestrians can turn around without changing positions. However, the vehicles operated by human drivers have physical movement constraints such as their maximum turning angle and minimum turning radius resulting from the structure of vehicles. In summary, deployment of the flocks-inspired approaches to capturing human drivers' interaction lays some challenges in real road traffic:

\begin{enumerate}
	\item \textbf{Traffic regulations}: Traffic scenarios are structural since traffic signs and lights regulate the traffic. Human drivers should obey many traffic norms and rules to guarantee safety and efficiency, such as keeping vehicles in lanes. 
	\item \textbf{Individual heterogeneous}: Not every human driver would strictly follow each item of Reynold's rules; instead, human drivers might adapt the regulations according to their driving tasks. For instance, human drivers' desired states might be different due to their driving styles. Moreover, human behaviors could be stochastic and time-varying, and vehicle dynamics could be different among agents. Hence, the individual heterogeneity of road users makes it intractable to perfectly calibrate models to  match the observations for all individuals.
\end{enumerate}

\subsection{Game-Theoretic Models}\label{sec:game_theory}
Human driving behavior is fundamentally a game-theoretic problem in which human drivers continuously make decisions by coupling with each other over time. Thus, the interaction between human drivers can form a closed-loop dynamic game considering the optimization-based state feedback strategy.
The interaction between human drivers can be modeled as a dynamic Markov game \citep{littman1994markov} where every agent can adapt to other agents' behavior to cooperatively or competitively complete a task. This situation forms multi-agent reinforcement learning (MARL). The developed game-theoretic approaches for imitating human drivers' interaction are usually applied to design the interaction policy between multi-autonomous vehicles or between autonomous vehicles and human-driven vehicles. A frequently used term in existing works on decision-making, planning, and control of autonomous driving is interaction-aware, i.e., considering the influences of interaction with other agents when designing controllers and making decisions and planning. Some researchers only utilized the game theory to model the discrete decision-making process during the interaction. In contrast, some researchers combined game-theoretic approaches with other learning methods (e.g., reinforcement learning, inverse RL \citep{albaba2021driver,schwarting2019social} and imitation learning \citep{espinoza2022deep}) and control theories (e.g., model predictive control \citep{zhang2019game,espinoza2022deep}, linear-quadratic Gaussian control \citep{schwarting2021stochastic}) to mimic the entire interaction process consisting of decision-making and control. 

\begin{figure}[t]
	\centering
	\includegraphics[width=\linewidth]{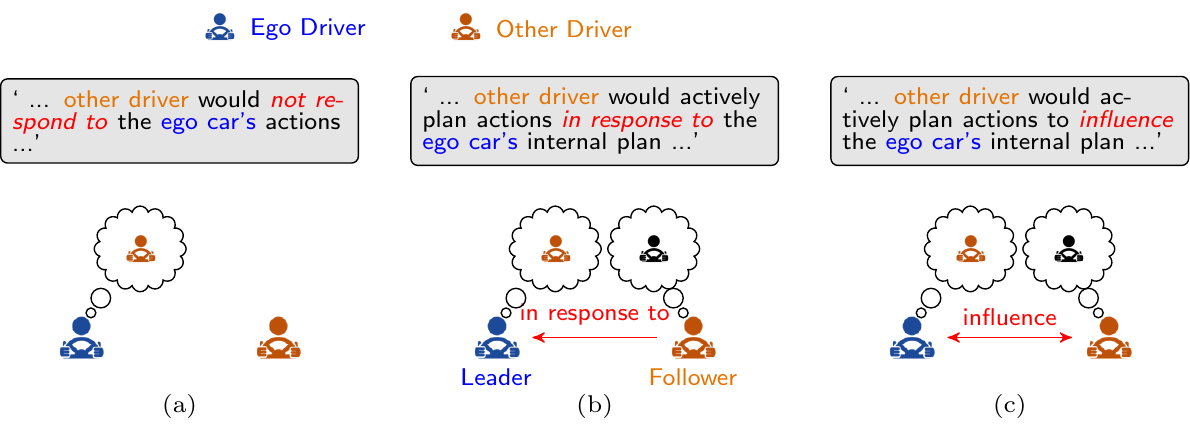}
	\caption{Illustration of three perspectives of relationships between two agents in one single stage of sequential games, which can be extended to multiagent interaction scenarios.  Treating other human drivers as (a) obstacles, (b) rational followers, and (c) mutual-dependence actors.}
	\label{fig:gametheory}
\end{figure}

\paragraph{Game Theory} Most of the earliest game-theoretic models of human driver interactions focus on matrix games \citep{kita1999merging,elvik2014review}. Currently, most of them switch on dynamic games \citep{lemmer2021maneuver} by translating interactive behavior into an iterative optimization problem. In games, there are different perspectives on the role of human drivers, and the role assignments would impact model performance. Hence, the first question needed for the dynamic games is `\textit{How should the ego vehicle consider the effects and roles of other human drivers in one single stage of sequential games?}' Generally, there are three solutions to this question (as shown in Figure~\ref{fig:gametheory}), discussed as follows. 
\begin{enumerate}
	\item \textbf{Treating other drivers as obstacles.} Most early works follow a direct pipeline when modeling how an ego human-driven vehicle would interact with other surrounding human drivers. First, predicting the trajectories of other human drivers and then feeding the predictions into the planning module of the ego vehicle as unalterable moving obstacles. Figure~\ref{fig:gametheory}(a) illustrates the relationship between the ego agent and other human agents. It is worth noting that this assumption is usually valid in interactions between autonomous and human-driven vehicles by treating human drivers as \textit{leaders} in a sequential game because humans may have less information and longer reaction time than autonomous vehicles. Hence, the human behavior won't suddenly change compared to the planning and control frequency of the robot and thereby can be treated as an obstacle. However, for the interactions among homogeneous agents (i.e., humans \textit{or} robots), this approach can cause excessively conservative behavior and even unsafe behavior in some cases, such as deadlock situations \citep{bouton2019cooperation}.  \citet{sun2022m2i} also used a similar `influencer-reactor' relationship to model pair-wise agents' interaction, where the reactor will respond to the behavior of an influencer who behaves independently without considering other agents. On the other hand, such an interaction scheme is a one-way interaction essentially, where only the ego vehicle is influenced by the other vehicles.
	
	\item \textbf{Treating the other driver as a rational follower.} To tackle conservative behavior and deadlock situations, researchers viewed other human drivers as rational utility-driven agents who will actively plan their trajectory \textbf{in response to} (\textbf{not influence}) the ego vehicle's internal plan in one stage of sequential games, as shown in Figure~\ref{fig:gametheory}(b). The ego vehicle can select a courteous action that will elicit the best actions/responses of the other human drivers in response, which is a typical two-agent Stackelberg (also known as leader-follower) game \citep{simaan1973additional,yu2018human}. Therefore, the leader has \textit{indirect} control over what the follower will do. This leader-follower game assumes (i) that the other human driver is rational and takes his/her own optimal plan by considering the ego vehicle's plans unchanged, and (ii) that the ego vehicle can assess the cost function (or utility) of other human drivers, i.e., the optimization-based model \citep{ozkan2021socially,liu2021interaction,liu2021cooperation,yu2018human,espinoza2022deep,schwarting2021stochastic}. In this framework, the ego car treats other human drivers as reactive followers instead of proactive followers. In one stage, the leader-follower game of human drivers entails a bilevel optimization -- an optimization on the higher level (i.e., the leader) which contains a lower level optimization (i.e., the follower). This optimization problem can be solved by (i) reformulating it as a local single-level optimization problem \citep{schwarting2019social}, (ii) approximating an optimal solution of the follower \citep{sadigh2016planning,sadigh2018planning}, \textit{and/or} (iii) setting assumptions on the uniqueness of each optimizer that maximizes utilities \citep{li2022game}.
	
	The leader-follower games provide an interpretable and computationally tractable scheme but has some drawbacks. First, the leader-follower game scheme neglects the dynamic \textit{mutual} influences stemming from the other human drivers' actions on the ego vehicle. The ego car has access to the internal function of the other human drivers, and the other human drivers only compute the best response to the ego car's actions  \citep{sadigh2018planning, sun2018courteous} rather than trying to \textit{influence} it.  Second, the role of leader or follower in applications is critical, but there are no unique ways to determine the leader and follower roles of interactive agents in applications. For example,  \citet{yu2018human,liu2021interaction} applied different rules to determine the agents' roles in the same driving tasks.
	
	\item \textbf{Treating other drivers as mutual-dependence actors.} The interaction between traffic agents depends on each other at each time step, as shown in Figure~\ref{fig:gametheory}(c). Such dynamic mutual dependence can be realized by a hierarchical game-theoretic framework consisting of strategic and tactical planners \citep{fisac2019hierarchical}. The strategic planner is modeled as a closed-loop dynamic game, and the tactical planner is modeled as an open-loop trajectory optimization. Another approach to capture dynamic mutual dependence is the simultaneous game in which all vehicles have the same reasoning strategies but each vehicle chooses their action without knowledge of the actions chosen by other vehicles. The simultaneous game formulation can solve the conflict problem. Usually, the multi-vehicle navigation problem can be formulated as a combinatorial optimization problem \citep{murgovski2015convex}, which could be centralized or distributed. In the centralized formulation, the objective function is a weighted sum of the cost functions for every participant. In the distributed formulation, each individual vehicle only considers its neighborhood locally. Directly solving the optimization problem might lead to a trapped situation -- all vehicles decide to slow down to yield. To solve this problem, \citet{liu2017distributed} proposed a communication-enabled conflict resolution.
\end{enumerate}

The game-theoretic framework provides an explainable explicit solution to model the dynamic interactions between human drivers. However, it is still hard to meet the real-time constraints regarding the computational tractability in its continuous state and action spaces although some progress has been made with simplified system dynamics and information structure \citep{fisac2019hierarchical}. With such limitations, most current game-theoretic interaction modeling approaches have scalability issues, thereby being limited to two-vehicle settings and simulation experiments or handling multiagent scenarios pairwisely \citep{wang2021socially,sun2018courteous,sun2018probabilistic,zhang2019game,mejia2022game}. To overcome the limitations, \citet{liu2022potential} proposed two practical, reliable, robust  frameworks using potential games \citep{monderer1996potential} with two solution algorithms to make real-time decisions for autonomous vehicles.

\begin{figure}[t]
	\centering
	\includegraphics[width = \linewidth]{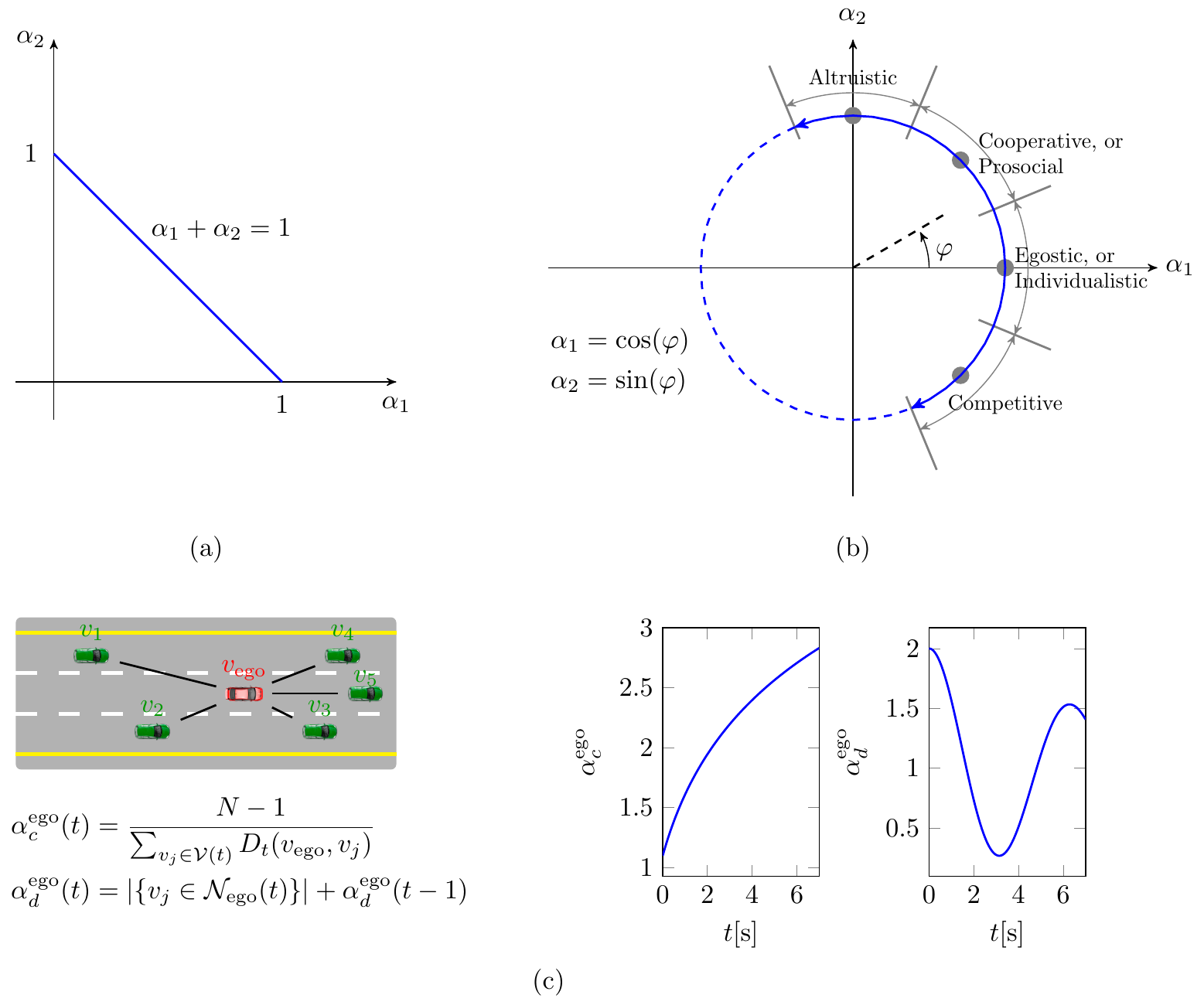}
	\caption{Illustration of two ways to evaluate the social preference of human drivers in the game-theoretic framework. (a) The linear summation equal to one. (b) Ring measure of social values. (c) CMetric measurement, $\boldsymbol{\alpha}^{\top}=[\alpha_{c}^{\mathrm{ego}}, \alpha_{d}^{\mathrm{ego}}]$, of driving styles based on dynamic geometric graphs.}
	\label{fig:SVO}
\end{figure}

\paragraph{Games with Social Factors} Social preference is a way to compensate for the long-term effect of repeated games into the reward for single-stage games. It is tractable to quantitively evaluate human drivers' social preferences in interactive scenarios, as reviewed in Section~\ref{sec:social_preference}.  The social value reflects the agents' experience in interactions, which is embeddable for the agent's utility (also named as a reward in the field of RL or a cost function in the field of control theory). Here we consider a multi-player\footnote{Player is a common terminology used in game theory, and agent is commonly used in machine learning and robotics. We do not discriminate between their usages in this paper. Both players and agents are referred to human drivers and/or autonomous vehicles.} game for an intuitive explanation, and each player represents a driver. The one we focus on denotes the ego agent (e.g., human driver or autonomous vehicle). Each driver will receive the reward by evaluating each action combination. The reward received by one driver is usually a weighted combination of the reward to itself and the reward to other agents,

\begin{equation}\label{eq:reward}
	r_{\mathrm{self}}(a_{\mathrm{self}}, a_{\mathrm{others}}, \alpha) = \alpha_{1} \cdot r_{\mathrm{self}}(\cdot) + \alpha_{2} \cdot r_{\mathrm{others}}(\cdot),
\end{equation}
where $\alpha = \{\alpha_{1}, \alpha_{2}\} $ with $ \alpha \in [0,1]$ indexes the driver's social preference, $r_{\mathrm{self}} (\cdot)$ and $r_{\mathrm{others}} (\cdot)$ represent the reward to ego driver and the reward to others excluding the ego driver, respectively. The balance of social preferences can be reflected by tuning the indexes $\alpha_{1}$ and $\alpha_{2}$, which can use one of the following formulations:

\begin{enumerate}
	\item \textbf{Linear summation equal to one.} This is the most straightforward way to define the relationship between $\alpha_{1}$ and $\alpha_{2}$ (as illustrated by Figure~\ref{fig:SVO}(a)), i.e., 
	
	\begin{equation}
		\alpha_{1} + \alpha_{2} = 1,
	\end{equation}
	by assuming that $\alpha_{1}$ is the altruistic tendency of the ego driver. Specifically, when setting $\alpha_{1}$ close to $0$ (i.e., $\alpha_{1} \rightsquigarrow 1$), maximizing the expectation of equation (\ref{eq:reward}) indicates that the ego driver would prefer to `give the right of way' to other drivers with an altruistic act. Inversely, setting $\alpha_{1}$ close to $1$ would lead to the ego driver's selfish act. This structure has been applied to social decision-making design for autonomous vehicles in interactive scenarios such as merge-in behavior \citep{geary2021active}.
	
	\item \textbf{Ring measure of social values.} McClintock developed a social value approach \citep{mcclintock1972social} by projecting the social value on a two-dimensional space with a ring measure. This approach is one of the most robust models for measuring an individual's interpersonal utilities \citep{liebrand1984effect,liebrand1988ring}. With this definition, \citet{schwarting2019social} evaluated the social value \textit{online} using the SVO angular preference by setting
	
	\begin{equation}
		\alpha_{1} = \cos (\varphi_{\mathrm{self}}), \qquad \alpha_{2} = \sin (\varphi_{\mathrm{self}}),
	\end{equation}
	where $\varphi_{\mathrm{self}}$ represents the ego agent's SVO, as illustrated by Figure \ref{fig:SVO}(b).
	This approach is then widely applied to decision-making and planning for autonomous driving to generate socially compatible behavior as human drivers in interactive traffic scenarios \citep{toghi2021social,zhao2021yield, crosato2021human, ozkan2021socially,buckman2019sharing}. However, these models do not estimate human's social preference in real-time; instead, they choose and learn a fixed parameter for each agent.  
	
	\item \textbf{CMetric.}  CMetric \citep{chandra2020cmetric} is a new approach to quantitively measuring human drivers' driving preference and style (i.e., aggressive and patient) using centrality functions \citep{rodrigues2019network} with a combination of computational graph theory and social traffic psychology, as shown in Figure~\ref{fig:SVO}(c). Unlike game-theoretic approaches leveraging other human drivers' social preferences with the assumption that the ego vehicle can access other human drivers' cost functions, the CMetric does not rely on such assumption. \citet{chandra2021gameplan} integrated the CMetric-based evaluation of driving styles into a Sponsored Search Auction (SSA) game-theoretic model to compute the optimal turn-based ordering in interactive scenarios such as unsignalized intersections, highway on-ramp merges, and roundabouts. The most profitable feature of CMetric is providing a real-time social preference estimation. 
\end{enumerate}

In addition to quantifying social factors through extra functions as mentioned above, some researchers also incorporate other social considerations (e.g., courteousness level, limited perception) of humans by adjusting the action set for each game agent and associated decision-making models \citep{li2022game}. \cite{yu2018human,zhang2019game} estimated aggressiveness online and then integrated it into the utility function of leader-follower games to leverage the driver's preference in lane-change tasks. \cite{zhang2021human} estimated the belief of the surrounding drivers' aggressiveness from the ego vehicle's perspective using a probabilistic model and then integrated this belief into a Bayesian game to capture interactions among drivers.

In real-world traffic, many different driving behavioral factors can influence the cooperativeness of human drivers. So, the other question is `\textit{how to leverage these factors into computational models and ensure their fidelity?}' The answer lies in the behavioral theory that describes how we assume one human driver to act and react to other human drivers on the road. 
\cite{hoogendoorn2009generic} summarized a list of behavioral assumptions supported by empirical evidence from different sources, which provides a basis for advanced interaction model derivation. The authors also proposed a subjective effort minimization on which a generic driving behavior is modeled using differential game theory.

\begin{figure}[t]
	\centering
	\includegraphics[width=\linewidth]{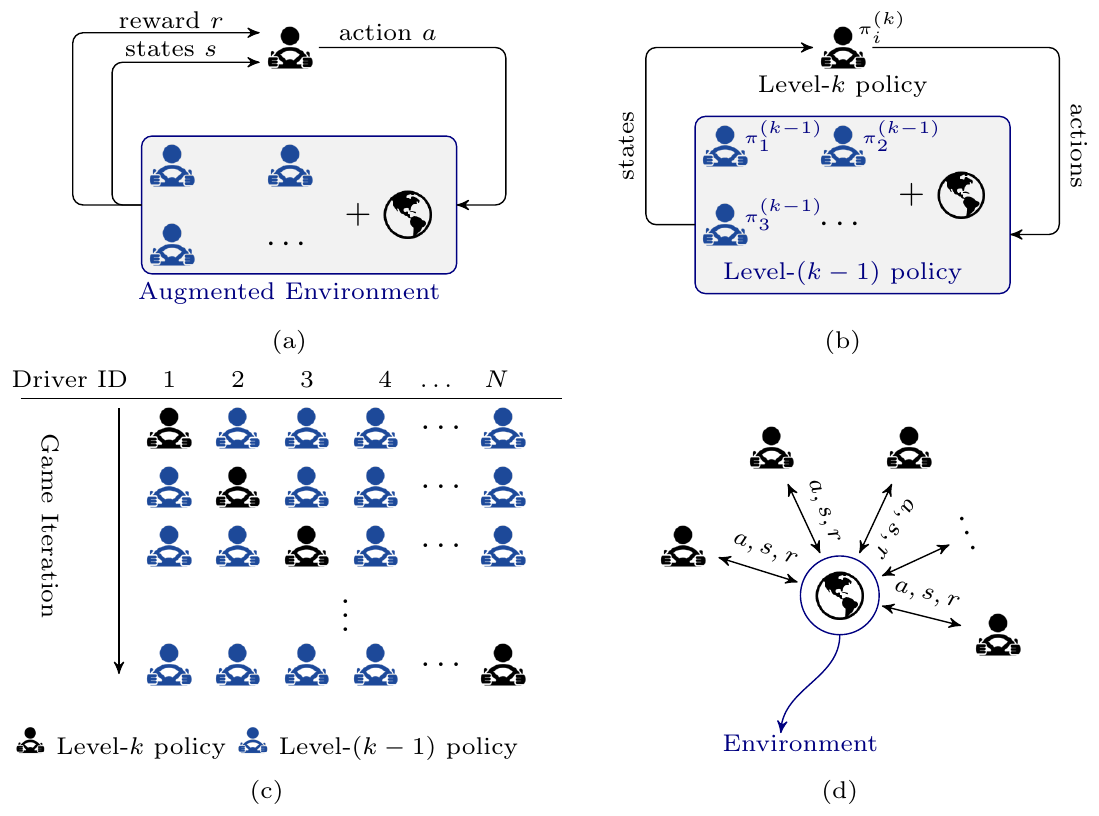}
	\caption{Asynchronous scheme  from (a) an RL perspective (i.e., a single-agent RL task) and (b) a level-$k$ game perspective. (c) Iteratively learning procedure of updating from level-$(k-1)$ to level-$k$ policies over all agents. (d) Synchronous scheme: Multi-agent MDPs or a Markov game.}
	\label{fig:MARL}
\end{figure}

\begin{algorithm}[t]
	\DontPrintSemicolon
	\KwInput{$n$, the number of agent; $K$, interaction level}
	\KwOutput{$\Pi^{K}$, the set of all agents' policies at level-$K$} 
	\textbf{Initialize:} $\Pi^{0} = \{\pi_{1}^{0}, \pi_{2}^{0}, \dots, \pi_{n}^{0}\}$, $k=0$, $i=1$ \tcp*{$\pi_{i}^{0}$, $i$-th agent's level-$0$ policy}
	\While{$k<K$}
	{
		\tcc{Iteratively obtain $i$-th agent's level-($k+1$) policy by holding other agents at the level-$k$ policies, $\Pi^{k}_{\neg i} = \{\dots, \pi_{i-1}^{k}, \pi_{i+1}^{k} \dots\}$}
		\For{$i \leq n$ }
		{
			\tcc{Learn policy}
			$\pi_{i}^{k+1} \leftarrow$ \texttt{RL Function}($\Pi^{k}_{\neg i}$) \;
			\tcc{Update and save policy}
			$\pi_{i}^{k+1}\leftarrow \pi_{i}^{k}$\;
			\tcc{Select next agent}
			$i= i + 1$
		}
		\tcc{Update levels}
		$k= k + 1$
	}
	
	\caption{Iterated Learning Procedure of Level-$k$ Games with RL}
	\label{alg:GameTheoryRL}
\end{algorithm}

\paragraph{Games with Agent Adaptation} Humans are \textit{adaptive} agents and can learn to drive via a  reward-reinforcement mechanism\footnote{The terminology `reinforcement' refers to the learning mechanism since decisions and actions that lead to high rewards (i.e., satisfactory outcomes) are reinforced in the human driver's set of behaviors.} through safely interacting with environments. Inspired by this, the procedure of learning to interact with other drivers can be formulated via reinforcement learning with the game-theoretic scheme. Treating the other agents except for the ego agent as part of the environment produces two types of game schemes of modeling interactions (Figure~\ref{fig:MARL}): asynchronous and synchronous. 

\begin{figure}[t]
	\centering
	\includegraphics[width=\linewidth]{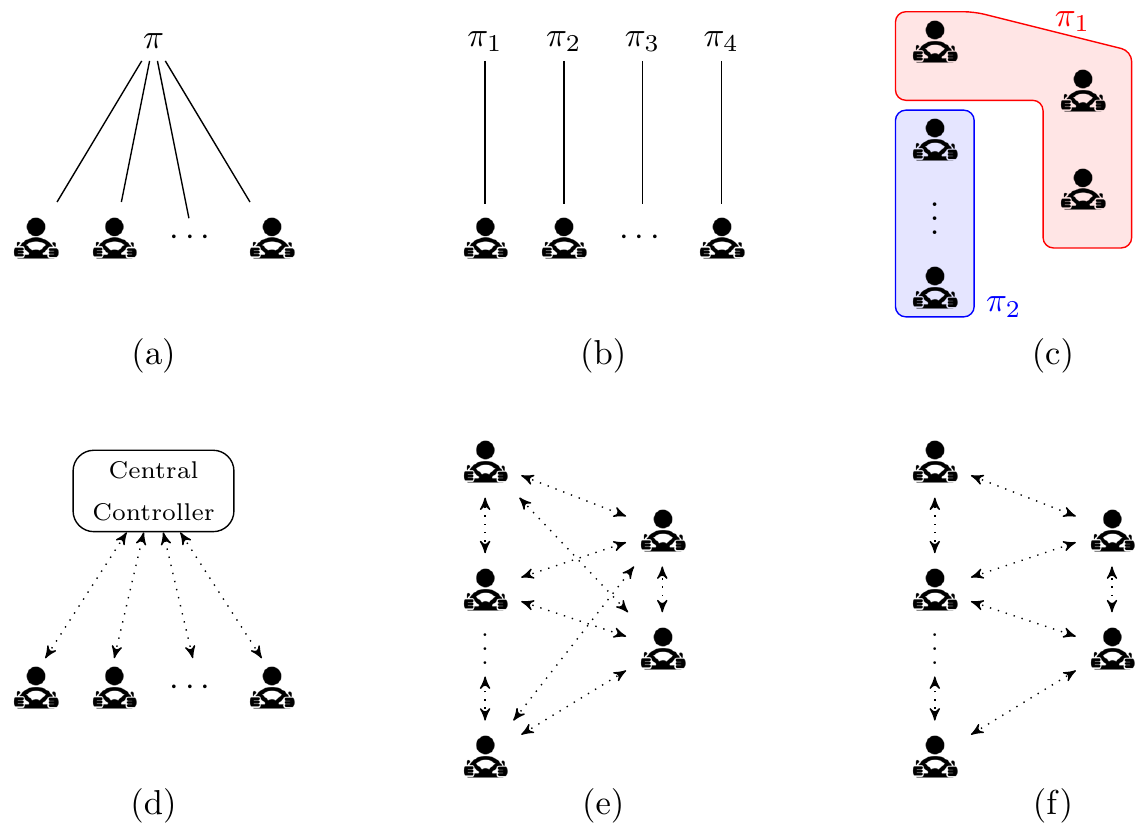}
	\caption{Commonly-used learning paradigms of MARL algorithms under a synchronous scheme. Independent human drivers with (a) shared policy, (b) independent policies, and (c) shared policies within a group. (d) Once the central controller governs all human drivers. (e) Centralized training with decentralized execution: During training, human drivers can exchange information with any other human drivers at any time; during execution,  human drivers operate independently. (f) Decentralized training with networked human drivers: During training, human drivers can exchange information with their neighbours in the network; during execution, human drivers operate independently. Dotted lines indicate the communication.} 
	\label{fig:synchronous}
\end{figure}

\begin{itemize}
	\item \textbf{Asynchronous Scheme.} In the asynchronous scheme, each driver treats all other surrounding drivers as part of the environment, as illustrated in  Figure~\ref{fig:MARL}(a). The interaction dynamics among human drivers under the asynchronous scheme can be realized using specific game-theoretic schemes such as level-$k$ games\footnote{The level-$k$ games involve the `level-$k$ reasoning', also known as a refinement of the game-theoretical approaches with hierarchical reasoning. Therefore, some works also denote level-$k$ games as hierarchical games.}, where human drivers' actions are predicted in an \textit{iterated} manner (Figure~\ref{fig:MARL}(c)) instead of being evaluated simultaneously. Specifically, to obtain the policy of a level-$k$ agent, the policies of all other agents are set to level-($k-1$), which effectively makes them a part of the environment whose dynamics are known. As a result, the policy of the level-$k$ agent is estimated as the best response to other agents' actions from the level-($k-1$) policy, as shown in Figure~\ref{fig:MARL}(b). Therefore, in a game with $N$ players, the $i$-th agent's level-$k$ policy can be obtained by
	
	\begin{equation}\label{eq:level-k game}
		\pi_{i}^{(k)} = \argmax_{\pi_{i}} U_{i} (\pi_{i}|\boldsymbol{\pi}_{\neg i}^{(k-1)})
	\end{equation}
	where $\pi_{i}^{(k)}$ is the $i$-th agent with the level-$(k)$ policy, $U(\cdot)$ represents the utility function, and $\boldsymbol{\pi}_{\neg i}^{(k-1)}$ are the level-$(k-1)$ policies of all agents except the $i$-th agent, as illustrated in \textbf{Algorithm} \ref{alg:GameTheoryRL}. Note that the reasoning levels of policies of all agents could remain the same \citep{li2018game,li2016hierarchical,li2017game,tian2020game}, be different from each other (leads to a dynamic level-$k$ policy) \citep{koprulu2021act}, \textit{or} be discrete/continuous over the policy space \citep{yaldiz2022modeling}. 
	
	An solution to the problem defined by (\ref{eq:level-k game}), is viewing the iterated policy learning task as an RL problem, known as a single-agent RL, by properly defining the states, actions, reward functions, and environment dynamics (see comparisons between Figures~\ref{fig:MARL}(a) and (b)). For instance, the ego human driver (black) would act \textit{after} perceiving the states and the rewards from the `augmented environment' constituted by other human drivers (blue) and traffic conditions. Once a single interactive level of the level-$k$ game is viewed as an RL task, many off-the-shelf RL algorithms can be used, such as deep Q-learning \citep{albaba2021driver,koprulu2021act}.  However, such a single-agent RL scheme usually leads to unstable control policies; even if the trained policies converge, they still lack performance guarantees \citep{matignon2012independent}. Moreover, the single-agent RL formalization might lead to risky behavior or even cause a collision since other drivers' strategic behavior does not change and would not influence the ego agent's actions at each level iteration \citep{ding2018game, albaba2019modeling}.
	
	\item \textbf{Synchronous Scheme.} In multi-driver interaction scenarios, each human driver is trying to solve the sequential decision-making problem \textit{simultaneously} through a trial-and-error-like\footnote{Here we use `trial-and-error-like' rather than `trial-and-error' since most human drivers would not actively or randomly execute `error' behaviors to kill themselves when driving; they could not survive in traffic otherwise.} procedure. The evolution of the environmental state and the reward function each human driver receives is determined by all drivers' joint actions. As a result, the human driver needs to consider and interact with the environment and the other human drivers. To this end, a synchronous scheme can capture the decision-making processes that involves multiple human drivers through a Markov game \citep{littman1994markov}, also known as a stochastic game (see definitions in Appendix~\ref{app:MarkovGames}) \citep{shapley1953stochastic, solan2015stochastic}. Each agent is formulated as an MDP-based agent, which would form Multi-Agent Reinforcement Learning (MARL). More details about MARL from a game-theoretical perspective refer to in the literature by \cite{yang2020overview}.
	
	Many different learning paradigms of MARL algorithms can be designed with specific assumptions for different interaction tasks. Theoretically, the learning paradigms can be classified into six groups \citep{yang2020overview} (as illustrated in Figure~\ref{fig:synchronous}), some of which with transportation applications are comprehensively reviewed and analyzed by \cite{schmidt2022introduction}. Moreover,
	\cite{ding2018game} developed a synchronous scheme under a two-agent Markov game that combined reinforcement learning to model the interaction process of lane-changing behavior. However, their proposed approach is limited to predefined traffic scenes where each interactive driver's social preference is predefined. In real traffic, human drivers can leverage the social cooperativeness of other drivers to avoid deadlocks and proactively convince others to change their behavior. Inspired by this, \cite{hu2019interaction} developed a MARL based on Markov games with the curriculum learning strategy to consider the level of cooperativeness and mimic the social priority of the right of way in merging scenarios. The RL can also predict the time-extended interaction dynamics of agents in dynamic games \citep{albaba2019modeling}.
\end{itemize}

\paragraph{Games with Incomplete Information} The most commonly used game-theoretic models are complete-information games, assuming that (i) all human drivers are rational, and (ii) each agent's information is accessible to others, e.g., one human driver knows the other human driver's objective/utility function and intentions. However, such information might not be available in real traffic, leading to information asymmetry between agents. As a result, there needs a non-empathetic estimation of other human's driving behaviors and acting irrationally toward others. In such interactive situations,  \cite{chen2021shall} reveals that \textit{empathy} is necessary when the true parameters of agents mismatch with their shared belief about all agents' parameters. To generate a safety-guaranteed decision of autonomous vehicles when interacting with irrational human drivers, \cite{tian2021safety} introduced one pair of social parameters $(\beta, \lambda)$ to encode the human drivers' rationality level and the role (i.e., leader or follower) and updated them via a Bayesian rule once observing new information. Besides, when the environment states are partially available, the interactions can be formulated via the partially observable stochastic game (POSG) \citep{toghi2021social} and partially observable Markov decision process (POMDP) \citep{li2017game}, which can be solved using reinforcement learning algorithms such as Q-learning \citep{fisac2019hierarchical}.

\paragraph{Games with Estimated Others} The social preferences of interactive agents can be parameterized and then embedded into the cost functions of each agent in a game considering the optimization-based state feedback strategy during the interaction, as formulated in equation (\ref{eq:reward}). In order to leverage another human agent's behavior into decision-making, researchers model interactions among human drivers based on two assumptions:
\begin{itemize}
	\item All agents are rational and aim to find the utility-maximizing control actions.
	\item The ego agent (e.g., human or robot) can access another human agent's reward/cost functions.
\end{itemize}
However, the above assumptions are not always directly available in natural environments, and it is necessary to estimate more information using the access data. 

\textbf{Utility Maximization.} The first assumption allows translating another human driver's decision-making process into an optimal maximization problem

\begin{equation}\label{eq:inter_cost}
	\mathbf{u}^{\ast}_{\mathrm{others}} (\boldsymbol{x}^{t}, \mathbf{u}_{\mathrm{self}}) = \argmax_{\mathbf{u}_{\mathrm{others}}} r_{\mathrm{others}} (\boldsymbol{x}^{t}, \mathbf{u}_{\mathrm{self}}, \mathbf{u}_{\mathrm{others}}),
\end{equation}
where $\boldsymbol{x}^{t}= [\boldsymbol{x}^{t}_{\mathrm{self}}, \boldsymbol{x}^{t}_{\mathrm{others}}]$ are the states of all agents. Equation (\ref{eq:inter_cost}) indicates that other human drivers are rational and respond to the actions of the ego vehicle (i.e., $\mathbf{u}_{\mathrm{self}}$) by finding the actions maximizing their utilities (or rewards). The optimal actions executed by other human drivers endow the associated maximum reward function of the current states and the action executed by the ego vehicle

\begin{equation} \label{eq:opt_cost}
	r^{\ast}_{\mathrm{others}} (\boldsymbol{x}^{t}, \mathbf{u}_{\mathrm{self}}) = r_{\mathrm{others}} (\boldsymbol{x}^{t}, \mathbf{u}_{\mathrm{self}}, \mathbf{u}_{\mathrm{others}}^{\ast}),
\end{equation}
which is the reward functions of other human drivers in equation (\ref{eq:reward}).

\textbf{Reward/Utility Function Estimation.} Usually, researchers formulate other human drivers' reward functions, i.e., $r_{\mathrm{self}}$, as the linear-structured weighted features of current states

\begin{equation}
	r_{\mathrm{others}} (\boldsymbol{x}^{t}, \mathbf{u}_{\mathrm{self}}, \mathbf{u}_{\mathrm{others}}) = \mathbf{w}_{\mathrm{others}}^{\top}\mathbf{f}_{\mathrm{others}}(\boldsymbol{x}^{t}, \mathbf{u}_{\mathrm{self}}, \mathbf{u}_{\mathrm{others}}).
\end{equation}
The associated weight vector, $\mathbf{w}_{\mathrm{others}}$, can be estimated from the interactive demonstrations via inverse optimal control theory (e.g., IRL) with the principle of maximum entropy \citep{ziebart2008maximum}. The IRL aims to learn an underlying cost function that encodes the driving preferences of the human driver in driving demonstrations. \cite{sun2018courteous,sun2018probabilistic} applied IRL to learn the weights in a linear combination of weighted features of human demonstrations from interactive driving scenarios by considering the influence of courteous behaviors of the ego vehicles on other human drivers under a structure of the Stackelberg game. \cite{ozkan2021socially} encoded the human driving behavior via IRL and integrated it into the autonomous vehicle's objective function to create a socially compatible control in car-following scenarios. 

\textbf{Future Behavior Prediction.} On the other hand, the reward function to be maximized for each agent (i.e., $r_{\mathrm{self}}$ and $r_{\mathrm{others}}$) in equation (\ref{eq:reward}) can be cumulative over a fixed horizon instead of a single step ahead. This operation requires the ego vehicle to predict other human agents' possible actions and states over the fixed horizon while leveraging the interactions. The level-$k$ game theory \citep{costa2009comparing}, assuming that all other players can be modeled as level-($k-1$) reasoners and act accordingly, allows accounting for the vehicle-to-vehicle interdependence while predicting future vehicle actions and states over a fixed horizon  \citep{schwarting2019social,li2017game, tian2020game}.

\subsection{Single-Agent Markov Decision Process}
Another pipeline of modeling \textit{how} an ego agent learns to interact with others is using the scheme of single-agent MDPs\footnote{A single-agent MDP does not mean only one agent in the environment. On the contrary, a single-agent MDP here is a general concept of treating other agents as a part of the environment.}. The single-agent MDP assumes that the environment containing other agents is \textit{stationary} and thus can be formulated by a Markov decision process (MDP) (see definition of MDP in Appendix~\ref{app:MarkovGames}).  The ego agent tries to select the optimal plans to maximize the associated rewards with consideration of the influences of its behaviors through dynamically rolling out interaction trajectories in mind when interacting with the environment. As a result, the ego agent's driving task can be formulated as an optimization problem over the policy, $\pi$, that maximizes the value function over a fixed horizon $T$, starting from the environment state $s$

\begin{equation}
	\label{eq:valuefunction}
	V_{\pi}(s) = \mathbb{E}_{a\sim\pi(s)} \left[ \sum_{t=0}^{T} \gamma^{t} r(s_{t}, a_{t}, s_{t+1})|s_{0} = s \right]
\end{equation}
From the ego's perspective, other agents are always part of the environment, as shown in Figure~\ref{fig:MARL}(a). Note that the single-agent MDPs are used to model interactions between the ego agent (fixed) and other agents except the ego, and many existing RL algorithms are available to use. The interactive processes between human drivers are considered when solving the problem, for example, using Q-learning. However, the stochastic games with an asynchronous scheme would alternately treat only one driver as the ego agent at each stage game, as shown in Figure~\ref{fig:MARL}(c).

In real traffic, human drivers can \textbf{anticipate} the possible outcomes of other human agents in their minds \citep{gallese1998mirror} and then \textbf{integrate} these potentials into their real-time planning to generate socially-compatible decisions and behaviors in interactive scenarios. Formulation of the interaction using MDPs derives two fundamental questions:
\begin{enumerate}
	\item How does the ego agent make predictions of other agents' future behavior?
	\item How does the ego agent utilize these predictions, i.e., integrate these estimated predictions into their planning?
\end{enumerate}

Answers to the first question of behavior prediction can be classified into \textit{reactive} and \textit{interactive} prediction based on how the ego agent considers the influences between itself and other agents.
\begin{itemize}
	\item \textbf{Unidirectional influence.} The ego agent makes future predictions of other human drivers' behavior without considering the effects of the ego agent's current and future maneuvers, treating other agents as response-free agents. These agents' behavior can be \textit{deterministic} or \textit{stochastic}. Deterministic agents only follow the predefined rules and conditions, e.g., a fixed velocity profile or driving with a priori known states/intents. For the stochastic agents, the ego agent cannot specifically know and predict other agents' behavior but know the probabilistic distribution of uncertainties regarding intentions and goals
	
	\begin{equation}
		p(\boldsymbol{x}^{\mathrm{other}}_{t:t+\tau}|\boldsymbol{x}_{t}^{\mathrm{other}}, \boldsymbol{x}_{0:t-1}^{\mathrm{other}}).
	\end{equation}
	Considering the influence unidirectionally allows the ego agent to do \textbf{reactive} planning.
	
	\item \textbf{Bidirectional influence.} The ego agent makes future predictions of other human drivers' behavior by considering the effects of the ego agent's current and future maneuvers on the surroundings. That is, assuming that the other agents would make a rational response to the ego agent's potential future states. A probabilistic interactive behavior prediction of other agents can be formulated by
	
	\begin{equation}
		p(\boldsymbol{x}^{\mathrm{other}}_{t+1}| \underbrace{\boldsymbol{x}_{0:t-1}^{\mathrm{other}},  \boldsymbol{x}_{0:t-1}^{\mathrm{ego}}}_{\mathrm{historical}}, \underbrace{\boldsymbol{x}_{t}^{\mathrm{other}}, \boldsymbol{x}_{t}^{\mathrm{ego}}}_{\mathrm{current}}, 
		\underbrace{\boldsymbol{x}_{t+1}^{\mathrm{ego}}}_{\mathrm{prediction}}).
	\end{equation}
	A widely adopted paradigm of such interactions with uncertainties is POMDP. Considering the influence bidirectionally allows the ego agent to do \textbf{interactive} planning.
\end{itemize}
Note that the underlying ideas of \textbf{unidirectional} and \textbf{bidirectional} influence formulations between the ego agent and other agents are analogous to the influences between agents in game-theoretic schemes illustrated by Figure~\ref{fig:gametheory}(a) and Figure~\ref{fig:gametheory}(b), respectively.

In what follows, we will discuss how existing works utilize the single-agent MDP formalism to formulate the interaction in dynamic and uncertain environments. 

\begin{itemize}
	\item \textbf{Partially observable MDP.} In POMDPs, the uncertainty information such as the intentions and replanning procedures of the other agents, observation uncertainty, and occlusions are usually unobservable and encoded in hidden variables. A commonly-used way is to build a probabilistic distribution of the current states, forming a belief state, which is available. The solutions to the POMDP formalism could be offline or online. \textit{Offline} means solving the POMDP problem concerns the best possible action, not for the current but for every imaginable belief state \citep{schwarting2018planning}. POMDPs are practically formulated in discrete spaces (i.e., discrete states, discrete actions, discrete observations, or their combinations) to make the problem computationally solvable. \cite{sezer2015towards} formulated the interaction problem between an autonomous vehicle and the other human drivers at an unsignalized intersection as a mixed Observability Markov Decision Process with discretized a state space using a simple behavior model considering the unknown intentions of other vehicles. Some works also applied the partially continuous spaces\footnote{Here, the \textit{partially} continuous spaces in POMDP mean that either state space, action space, or observation space is discrete.} to build POMDP models. \cite{bai2014integrated} developed a continuous-state, continuous-observation POMDP to solve the decision-making problem at a simple intersection. \cite{hubmann2018automated} developed a continuous POMDP model by encoding the uncertainties from sensory data and human driver's intention into belief states. The authors obtained an optimized policy by leveraging the most likely future scenarios from other agents'  probabilistic interactive motion models. 
	
	On the other hand, solving the POMDP problem can be \textit{online} but needs a tradeoff between the solution accuracy and computationally efficiency \citep{hubmann2018automated, schwarting2018planning}.
	
	\item \textbf{Q-learning.} Human behavior in nature is shaped by reinforcement rather than free will, and interactions follow this rule \citep{skinner1958reinforcement}. Hence, reinforcement learning with the MDP formalism enables us to formulate the interaction problem because the human agent learns to drive in dynamic and uncertain environments by continuously interacting with the environment. So the followed question would be `\textbf{How do we integrate the interactive influences between agents into RL algorithms?}' 
	
	The influence of other agents on the ego agent can be considered while planning via an associated value iteration process. Existing works applied Double Q-Learning (DQL) algorithms to consider the effects of other agents' actions and states on the ego agent's value evaluation. For instance, \cite{ngai2011multiple} developed a double action Q-learning (DAQL) to evaluate the Q-function value of avoiding collisions while other goals are evaluated with pure QL independently
	
	\begin{equation}
		\begin{split}
			Q_{i}(s_{i,t}, a_{t}^{\mathrm{ego}}, a_{t}^{\mathrm{other}}) \leftarrow & Q_{i}(s_{i,t}, a_{t}^{\mathrm{ego}}, a_{t}^{\mathrm{other}}) + \\
			&  \alpha \left[ r_{i,t+1} + \gamma \max_{a_{t+1}^{\mathrm{ego}}} Q_{i}(s_{i,t+1}, a_{t+1}^{\mathrm{ego}}, a_{t+1}^{\mathrm{other}}) \right],
		\end{split}
	\end{equation}
	where $s_{i,t}$ and $s_{i,t+1}$ are the input states,  $a_{t}^{\mathrm{ego}}$ and $a_{t}^{\mathrm{other}}$ are the ego and other agent's actions, respectively.
	In natural traffic environments, the complex interactive driving task is a typically multiple-goal problem, such as target seeking and lane following while avoiding collisions in lane change tasks. For instance, \cite{ngai2011multiple} developed a multiple-goal RL framework to learn the control policy for the ego vehicle while driving in a dynamic environment where other agents' states change over time. Only considering the \textit{unidirectional influence} (i.e., the influence of the ego vehicle's actions on other agents) could lead to overly conservative plans far from the real cases because it does not explicitly model the \textit{mutual influence} of the actions of interacting agents. The solution to this limitation is formulating the mutual influence via a partially observable Markov decision process (POMDP) in which the driver type (i.e., aggressive or caution) is not observable to others \citep{chen2021midas}. The interaction policy is learned via the off-policy RL, while other drivers utilize a shareable Oracle policy.
\end{itemize}

\paragraph{Connection to Stochastic Games for Modeling Multiagent Interactions} Single-agent MDPs and stochastic games can capture the interactions among multiple agents, but some distinguishing grounds exist between them. In general, stochastic games can be derived in two ways \citep{bowling2000analysis}:
\begin{itemize}
	\item Extending single-agent MDPs  to multiagent MDPs. The main difference between single-agent and multiagent MDPs for modeling interactions among human drivers lies in whether each driver must take strategic actions considering others when determining their policies. This can be realized by extending Figure~\ref{fig:MARL} (a) to (d). 
	\item Extending matrix games to multiple states. Each state in a stochastic game can be viewed as a matrix game with the rewards for each joint action of all human drivers. The states of all agents are transited into another state (i.e., matrix game) according to their joint action after playing the matrix game and receiving the rewards.
\end{itemize}
Therefore, stochastic games contain MDPs and matrix games as subsets of the framework. For the solutions to stochastic games from the perspective of game theory and MARL, we refer to reading literature of \cite{bowling2000analysis,yang2020overview}.

\subsection{Learning-from-Human Demonstration}
The above-discussed swarm optimization and game-theoretic approaches are all designed forwardly. Some solve an optimization problem by setting their hyperparameters heuristically rather than optimizing with data. They analyze the scenarios with cognitive insights and then design an associated cost/objective function \textit{known a priori} to be optimized to mimic the interaction behavior between human drivers. 

The underlying mechanisms of decisions and motions behind the social interactions between human drivers are complex and difficult to encode into simple, manually-programmed rules. Generally, demonstrating interaction behaviors is much easier than specifying a reward function that would generate the same behavior. This fact provides an alternative to modeling and learning human drivers' interactions: directly learning interaction behaviors from human demonstrations via \textbf{imitation learning}. Based on what is learned (behavior trajectories or utilities), there are two types of approaches: (i) \textbf{behavior cloning}, which directly learns a mapping from observations (e.g., images) to actions (e.g., steering angle and gas pedal) or (ii) \textbf{utility recovering}, which indirectly uses data to retrieve a reward function for which planned interactive behavior mimics the demonstrations as close as possible. 

\paragraph{Behavioral Cloning} Behavioral cloning is the simplest form of imitation learning, which focuses on copying the agent's policy using supervised learning. Its main advantages are simplicity and efficiency. Behavioral cloning aims to solve a regression problem in which the optimization is achieved by maximizing an objective function (e.g., the likelihood of the actions taken in training data) \textit{or} minimizing the loss such as the derivations of behaviors between simulation (i.e., model outputs) and real data (i.e., demonstrations). Behavioral cloning has demonstrated efficiency in producing driving policies for simple driving behaviors \citep{pomerleau1988alvinn} such as lane tracking and car-following on highways. The success of behavioral cloning relies on sufficient data that can adequately cover the state and action space of both training and testing data sets \citep{kuefler2017imitating}. However, behavioral cloning does not leverage the cascaded learning errors in the \textit{training} process. The cascading errors will occur in the \textit{testing} process, assuming independent and identically distributed (i.i.d.) data \citep{bagnell2015invitation}. Hence, the models learned by behavioral cloning usually perform poorly in complex interactive scenarios. 

\paragraph{Utility Recovering} Behavioral cloning aims to copy the expert's behaviors or trajectories directly via regression techniques. On the contrary, IRL methods rely on retrieving cost functions from the observed interactive behaviors/trajectories \citep{ng2000algorithms} with the presupposition that reward functions are the most succinct, robust, and transferable \citep{abbeel2004apprenticeship} across different traffic scenarios. This presupposition coincides with the fact that human drivers can interact with other agents efficiently and safely across scenarios they have never seen. 

Recovering the interaction process between agents usually assumes that the state of the environment essentially has Markov property\footnote{The environment state has Markov property if its `state' contains sufficient relevant information about the current and the past interaction dynamics between the agent and the environment.} \citep{sutton2018reinforcement}, which allows formulating a learning task as a Markov decision process (MDP). Thus, the interaction process of human drivers is formulated through a parameterized model whose parameters can be estimated by optimizing a designed objective function. The straightforward way is to use a standard MDP, where the other human drivers are treated as part of the environment (i.e., Figure~\ref{fig:synchronous}(a)). In real-world traffic, human drivers have limited sensing capabilities and cannot precisely perceive the information they need; thus, a partially observable MDP (POMDP) can be used to model the interactions \citep{luo2021interactive}. Under the structure of MDP, human drivers' interactive decisions and controls are viewed as optimal solutions to the current traffic scene by maximizing their rewards (or minimizing their costs) by considering possible outcomes in the short future. Such an assumption allows learning human driver's interaction process via inverse optimal control (IOC) or IRL \citep{ng2000algorithms}. For example, \cite{sun2018probabilistic} developed a hierarchical IRL to explicitly mimic the hierarchical trajectory-generation process: making discrete decisions and executing continuous operations (i.e., steering wheel and hitting gas/braking pedals). They hierarchically model the discrete decision as a game in which one driver needs to determine the homotopy class of his future trajectory (e.g., yield or pass) and the continuous control as an IRL problem. 

Note that the IRL is used as a function to retrieve rewards in imitation learning and game-theoretic models but plays different roles in these two settings. In imitation learning, the IRL aims to learn reward functions for \textit{ego agents} to mimic the ego agent's driving behavior. In game-theoretic models, the IRL is used to learn \textit{others}' reward functions treated as the ego agent's inputs.

\subsection{Summary}
Rational human driving behaviors are `near-optimal' or `optimal' outcomes among all possible solutions in response to dynamic environments, though sometimes they could also converge to a local minimum or Nash equilibrium. This observation enables us to formulate human interactions into a computationally tractable optimization model maximizing a specific objective. The other popular behavior optimization approach for mimicking human driving interactions while guaranteeing collision-free is the velocity obstacles that have been widely used in the field of multiple mobile robots \citep{fiorini1998motion}. This approach has succeeded in simulating the interactions in heterogeneous traffic environments and traffic trajectory prediction \citep{luo2019gamma}.

Optimization-based approaches are analytically explainable and mathematically provable, embracing different constraints to avoid collisions \citep{li2021safe}. However, solving such complex optimization problems would be challenging for online applications with satisfying computation performance.

\section{Deep Neural Networks-based Models}
\label{subsec:deeplearning}
Deep learning has become increasingly powerful recently, with notable achievements in tasks such as computer vision and natural language processing. Due to its powerful expressiveness with low human effort and abundant data, it is also rising as a method for interaction modeling and relationship reasoning to solve various autonomous driving problems including planning and prediction, with diverse supervision manners (e.g., supervised and unsupervised) and different learning procedures (e.g., imitation learning and reinforcement learning).

This section will first introduce the basic modules of deep neural networks. Then we will discuss different ways of utilizing these modules for interaction modeling. Note that here we mainly focus on methods that use the vehicle states (e.g., position, speed, acceleration, and heading angle) as the inputs by considering the two followed aspects. First, deep learning methods that directly process raw sensory data (e.g., convolutions over camera RGB images and 3D LiDAR point cloud) usually follow pure end-to-end learning procedures \citep{cui2019multimodal,codevilla2018end,rhinehart2019precog}, leading to weak and implicit interaction reasoning. Also, the central focus of these works usually does not lie in interaction modeling.
Second, most existing high-quality autonomous driving motion datasets\footnote{The commonly used trajectory datasets for interaction modeling include \textsc{interaction} dataset \citep{interactiondataset}, HighD dataset \citep{highDdataset}, inD dataset \citep{inDdataset}, and others (see \url{https://levelxdata.com/trajectory-datasets}).} can provide sufficient and precise trajectory information of agent states.

\begin{figure}[t]
	\centering
	\includegraphics[width=\linewidth]{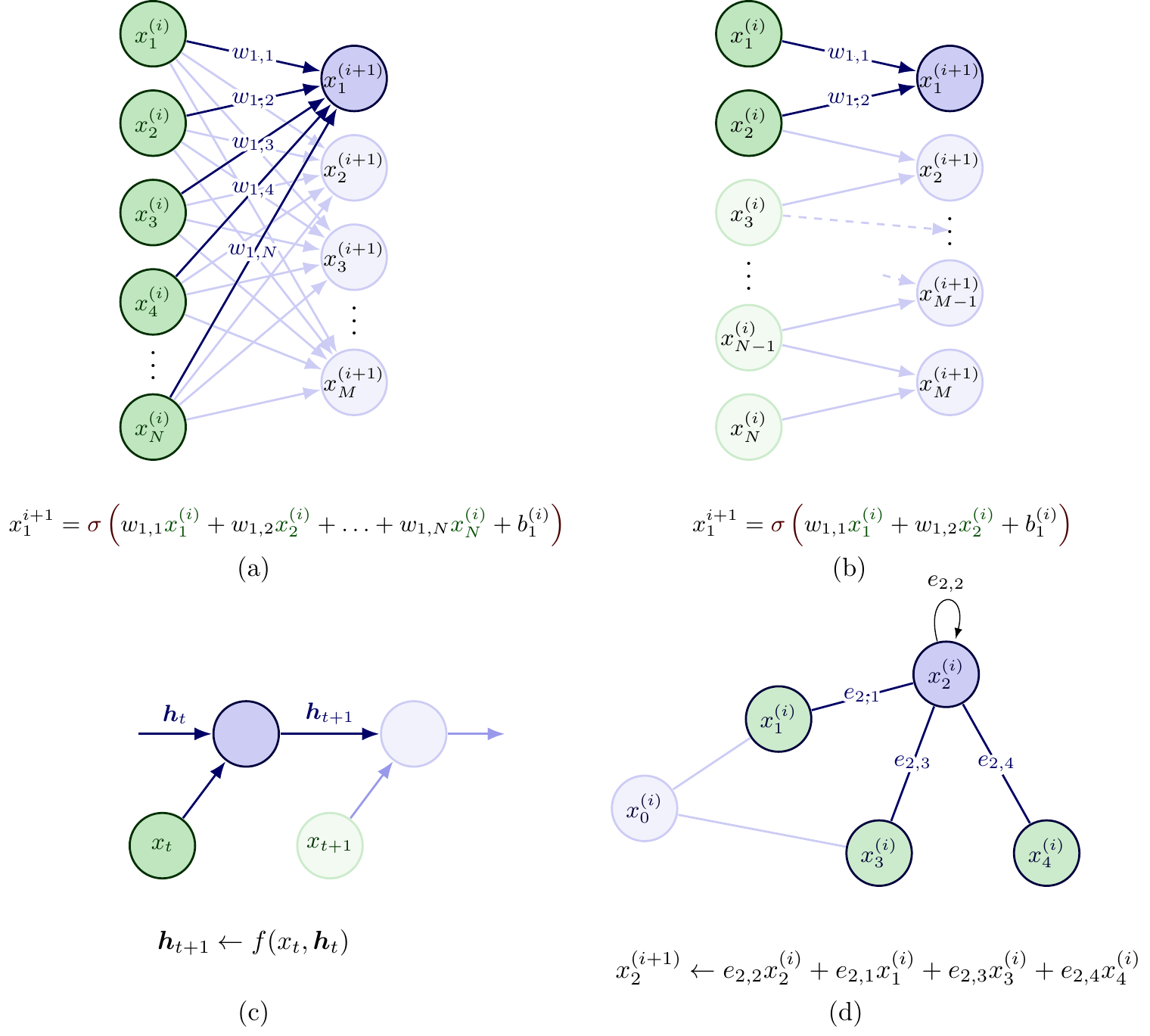}
	\caption{Three reusable and shareable deep neural network building blocks to model multiagent interactions. (a) Fully connected layers, (b) Convolutional layers, (c) Recurrent layers, and (d) graph layers.}
	\label{fig:deeplearningmodule}
\end{figure}

\subsection{Deep Neural Networks for Interaction Encoding}
\label{subsubsec:deepmodules}

Deep learning is a powerful tool with good potential for multi-agent interaction modeling. Nowadays, the neural network can be really complex. For example, GPT-3 \citep{brown2020language}, a large model for natural language processing tasks, has a capacity of 175 billion model parameters in its full version. However, essentially there are four types of fundamental neural network building blocks, based on which sophisticated networks can be developed for specific applications. Different building blocks pose different inductive biases\footnote{In a learning process, there can be multiple solutions that are equally good. An inductive bias is a prior assumption on the data, the solution space, or the problem structure that allows the learning algorithm to prioritize one solution (or interpretation) over others \citep{battaglia2018relational}.} on the data and solution space. People choose different building blocks to inject different inductive bias into the learning process for better performance, and a mismatched inductive bias could inversely result in suboptimal or poor performance. These modules may have been off-the-shelf to many deep learning practitioners, and more complex variants and operations in each of these layers are being developed. We here briefly introduce the basic structure and the central property for each type of layer to better remind us of their pros and cons and the design principle behind complex networks. For detailed descriptions, please refer to \cite{battaglia2018relational}.

\paragraph{Fully Connected Layers} The fully connected layer \citep{rosenblatt1961principles} is perhaps the most common module. In practice, both the input and output are usually a vector representing physical or hidden features. As shown in Figure~\ref{fig:deeplearningmodule}(a), each element (also known as entry or unit) in the output vector is the product of the input vector and a weight vector, followed by an added bias term and a non-linear activation function. As the name implies, fully connected networks (FC, also called multi-layer perceptron) connect all entries in the input vector to all entries to the output vector. In other words, \textit{all} input entries allow interacting with and contributing to \textit{any} output entry. Thus the inductive bias posed by the fully connected layer can be very \textit{weak}.

\paragraph{Convolutional Layers} The convolutional layer \citep{fukushima1982neocognitron} is realized by convolving an input vector or tensor with a kernel, adding a bias term, and applying a point-wise non-linear function, as shown in Figure~\ref{fig:deeplearningmodule}(b). The inputs could also be the vector values as in a fully connected layer. However, the connection between layers is sparser, as only entities \textit{nearby} could directly interact with each other and contribute to the output entry in each convolution operation. The same kernel is repeatedly utilized for all convolution operations in different regions of the input. Consequently, one of the main differences between a convolutional layer and a fully connected layer is the inductive bias of \textit{locality and translation invariance}. The convolutional layer is usually assumed to be suitable for capturing the spatial relations.

\paragraph{Recurrent Layers} The recurrent layer \citep{elman1990finding} is usually implemented over a temporal data sequence, as illustrated in Figure~\ref{fig:deeplearningmodule}(c). 
At each processing step, the new hidden state (e.g., $\boldsymbol{h}_{t+1}$) is generated by taking the current step's input (e.g., $x_{t}$) and the previous-step hidden state ($\boldsymbol{h}_{t}$). The same recurrent cell repeats in different time steps. Thus the recurrent mechanism poses the inductive bias of \textit{temporal invariance} and is assumed to be suitable for capturing the temporal relations. Commonly used recurrent neural network cells include Long-short Term Memory (LSTMs, \cite{hochreiter1997long}) and Gated Recurrent Units (GRUs \citep{cho2014properties}).

\paragraph{Graph Layers} The graph layer \citep{battaglia2018relational} can capture more explicit relational reasoning over graph-structured representations. A typical graph consists of nodes, edges and global attributes, where the edges represent relations between nodes and the global attributes present the context information. The computation of graph layer is a graph-to-graph function (Figure~\ref{fig:deeplearningmodule}(d)), which proceeds with updating attributes of each edge, aggregating edge attributes for each node, updating attributes of each node, and aggregating edge and node to update global attributes. In such a computation procedure, the graph layer poses the inductive bias of \textit{order invariance} for explicit relationship reasoning. That is to say, unlike fully connected and recurrent layers, the order of the input in a graph layer does not impact the results. Besides, the graph layer can deal with a varying number of agents. With these advantages, graph layers are usually assumed to be good at explicitly capturing interactions and relationships in multi-agent settings.

\subsection{Feature Representation for Social Interactions}
The spatial-temporal state feature tensor, spatial occupancy grid, and the dynamic insertion area are three commonly-used feature representations when modeling interactions with deep learning models. We will discuss these three types of feature representations as follows.

\begin{figure}[t]
	\centering
	\includegraphics[width=\linewidth]{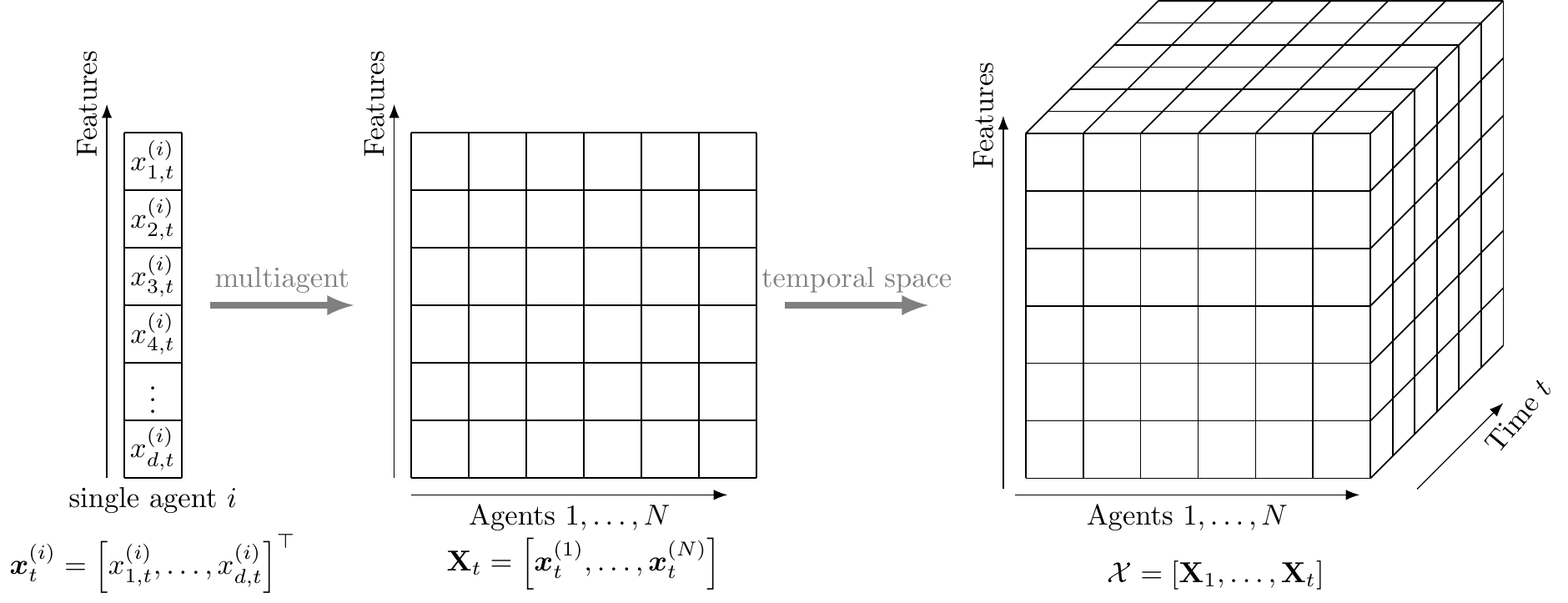}
	\caption{Construction of the list of state features, from single agent, to multi agent, to temporal space.}
	\label{fig:deeprepresentation}
\end{figure}

\paragraph{List of State Features} Assume that $N$ agents are interested in the traffic scenario and their state features\footnote{Here the term `state feature' is a general concept of sequential states, such as position, speed, and acceleration.} are measurable. Let's denote the $i$-th agent's state feature at time step $t$ as a vertical vector $\boldsymbol{x}_{t}^{(i)}\in\mathbb{R}^{d}$, as shown in Figure~\ref{fig:deeprepresentation}. Thus, we can shape all $N$ agents' states feature at a time step $t$ as a matrix with size $d\times N$\footnote{If a single agent's state at each time step is a vector, i.e., $\boldsymbol{x}_{t}^{(i)}\in\mathbb{R}^{d}$ with $d\geq 2$, then $\mathbf{X}_{t}$ is a matrix in the space $\mathbb{R}^{d\times N}$. If each agent's state is a scalar, then $\mathbf{X}_{t}$ will be a vector.}

\begin{equation}
	\mathbf{X}_{t} = \left[\boldsymbol{x}_{t}^{(1)}, \boldsymbol{x}_{t}^{(2)}, \dots, \boldsymbol{x}_{t}^{(N)}\right]
\end{equation}
Further, extending the matrix to cover a time horizon $T$ would generate a three-dimensional matrix $\mathcal{X}\in\mathbb{R}^{d\times N \times T}$. This representation is efficient because it uses the smallest amount of information necessary to represent the interaction scene. However, it has two limitations. First, the feature size of $\mathcal{X}$, the number of vehicles $N$ and each vehicle's valid time step $T$, could vary over time and space, which is problematic for learning approaches that expect constant-sized inputs. Second, this type of feature representation is \textit{permutation-variant}, i.e., dependent on the order in which the interactive agents are listed. For instance, simply switching feature entries of agent $i$ and agent $j$ in $\mathcal{X}$ would result in a different feature representation. A commonly-used way to avoid these limitations is to use an occupancy grid map, discussed as follows. 

\begin{figure}[t]
	\centering
	\includegraphics[width=0.7\linewidth]{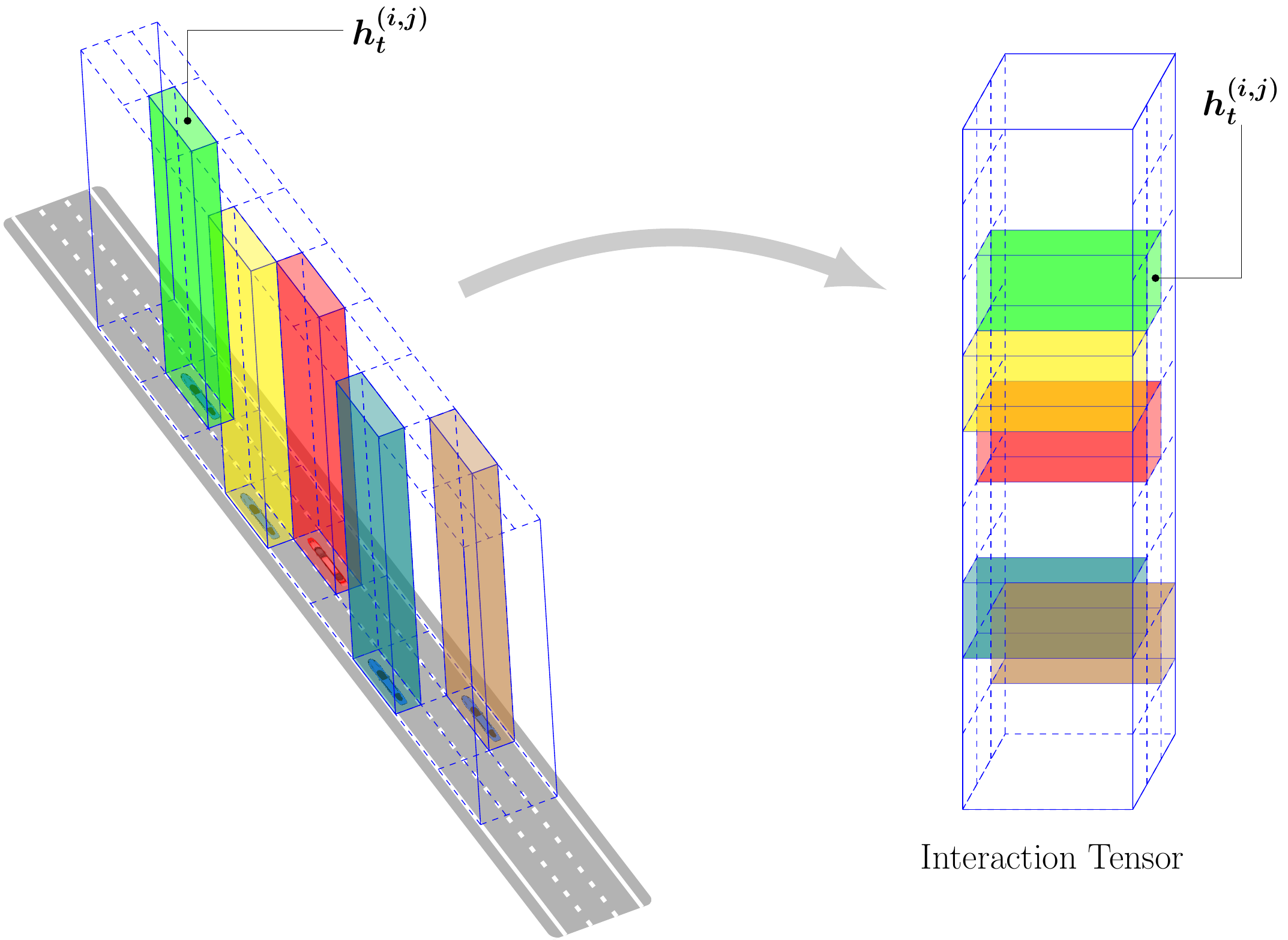}
	\caption{Left: Populating the spatial occupancy grid map with hidden features for multiagent interaction representation. Right: Populated spatial occupancy grid map representation. }
	\label{fig:occupygrid}
\end{figure}

\paragraph{Occupancy Grid Map} The occupancy grid map defines a spatial grid around the ego agent (i.e., ego-centric) or in a particular fixed region (i.e., scene-centric). The occupancy grid map can deal with a varying number of agents in the region of interest (RoI). The occupancy grid map can represent the interaction scene by being populated with either raw states (e.g., position, velocity, acceleration) or encoded states (e.g., hidden states output by fully connected layers). The spatial relationship between agents is naturally captured in the grid map layout. Besides, as shown in Figure~\ref{fig:occupygrid}, it can also capture the information over both spatial and temporal space if the grid, $\boldsymbol{h}_{t}^{(i,j)}\in\mathbb{R}^{d}$, (represented by color bars) is populated by the hidden features independently encoded from each vehicle's historical trajectory of $\tau$ time steps:

\begin{equation}
	\boldsymbol{h}_{t}^{(i,j)} = f( \boldsymbol{x}_{t-\tau:t}).
\end{equation}
where the function $f(\cdot)$ could be, for example, LSTMs, and $\boldsymbol{x}_{t-\tau:t}$ represent the single agent's historical trajectory from time step $t-\tau$ to the current time step $t$. Note that the shape of the grid map depends on the scene. For example, the grid map could be rectangle grids for highway interactions \citep{deo2018convolutional} or log-polar grids at the roundabout \citep{lee2017desire}. The spatial grid representation is raster-size and agent-order invariant but, in return, suffers from an accuracy-size trade-off \citep{leurent2019social} since the size of the tensor is related to the covered area size and the grid resolution.

\begin{figure}[t]
	\centering
	\includegraphics[width=0.9\linewidth]{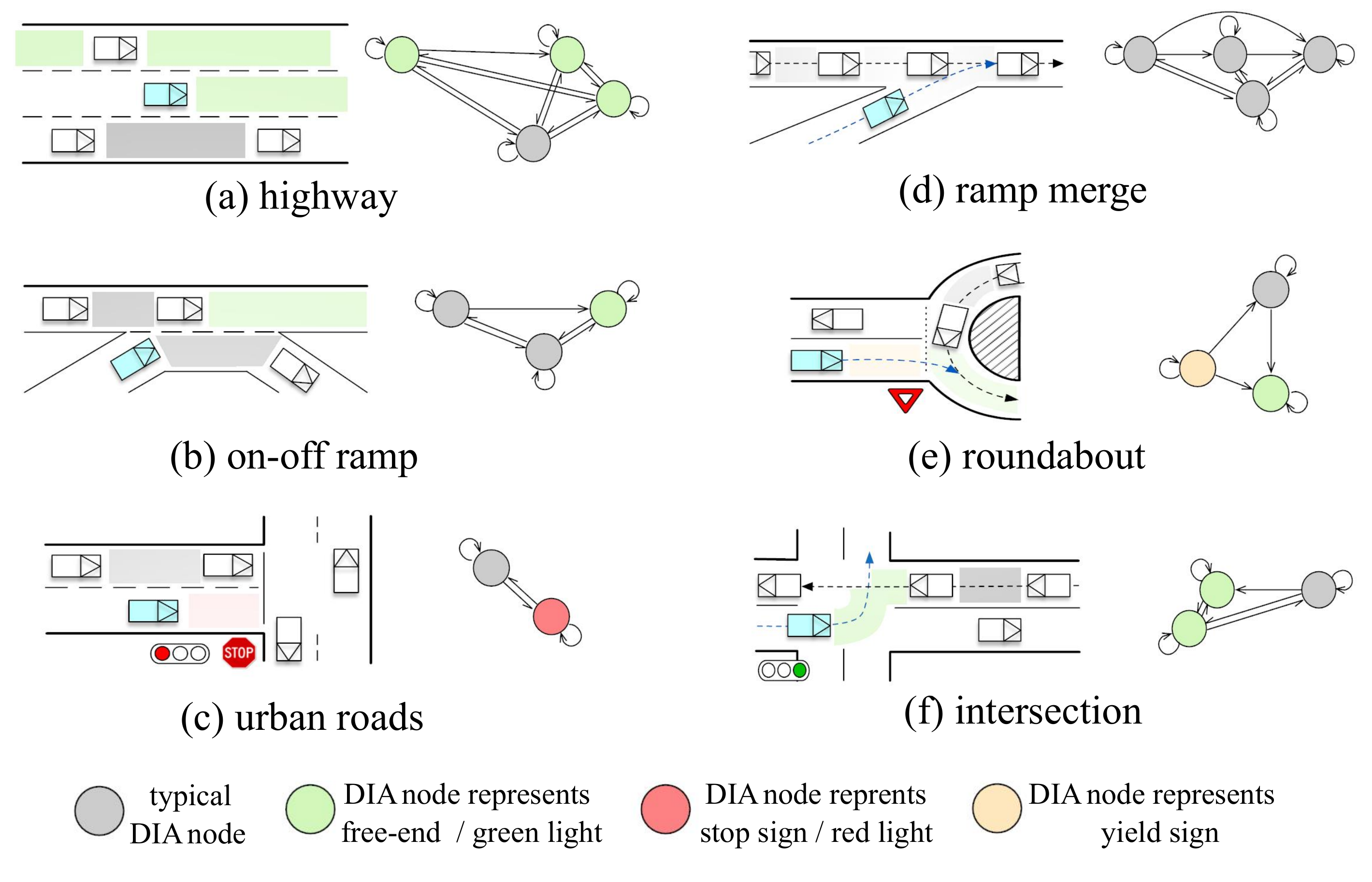}
	\caption{The extraction of dynamic insertion area (DIA) and the construction of semantic graph on diverse scenarios \citep{hu2020scenario}. The DIA and SG can serve as a unified and generic representation of the driving scene.}
	\label{fig:DIA}
\end{figure}
\paragraph{Interaction Graph}
Another representation is to use graphs by formulating entities in the driving scene as graph nodes $v_i$ and relationship among entities as graph edges $e_i$. All the nodes and edges in a time horizon $T$ then construct a spatial-temporal interaction graph $G$, which allows explicit interaction and relationship reasoning. There are two types of graph, varying in the definition of the entity/node. The first one is \textbf{agent-based graph}, which formulates agents in the scene as the nodes. The agent refers to road participants such as vehicles, pedestrians, motorcycles. Each node is represented by the state features or encoded features of each agent. While the agent-based graph concentrates on representing agents, another graph called the \textbf{area-based graph} focuses on representing the intention of vehicles. Many approaches classify driving intentions at the level of maneuvers such as lane keeping, lane changing, and turning. However, these maneuvers are highly conditioned on the driving scene. For example, the maneuvers could be quite different in highways or roundabouts due to distinct road topologies. In order for autonomous vehicles to drive through dynamic interactive traffic scenes in real life, a unified and generic definition on driving intention is necessary. Toward this end, one promising representation is \textit{dynamic insertion area} (DIA) \citep{hu2018probabilistic,hu2020scenario}, namely the available gaps on the driving scene that vehicles can insert into. 
As shown in Figure~\ref{fig:DIA}, when extracting DIAs from a scene, both static elements (road topology such as Frenet frame coordinates, road marks such as stop signs) and dynamic elements such as moving vehicles in the scene are leveraged. Thus the DIA can serve as a unified representation of the dynamic environment, covering all types of driving intentions and interactions on all road settings. Formally, the $i$-th DIA at time step $t$ is defined as $\mathbf{A}_i^t=(X_{\mathrm{front}}, X_{\mathrm{rear}}, X_{\mathrm{ref}})$, comprising information of the front boundary $X_{\mathrm{front}}$, the rear boundary $X_{\mathrm{rear}}$, and the reference path $X_{\mathrm{ref}}$ that the DIA lies on. The front and rear boundaries are formed by vehicles or road marks. All DIAs in a time horizon $T$ then can construct a temporal-spatial semantic graph $\mathbf{G}_t$,
 
\begin{equation}
	\mathbf{G}_t = \left[\{\boldsymbol{A}_1^t, \boldsymbol{A}_2^t, \dots, \boldsymbol{A}_N^t\}, \dots \{\boldsymbol{A}_1^{t+T}, \boldsymbol{A}_2^{t+T}, \dots, \boldsymbol{A}_N^{t+T}\} \right].
\end{equation}
where each DIA serves as a node of the graph.
The intention and interaction among agents then can be encoded and reasoned through manipulations on the spatial-temporal semantic graph \citep{hu2020scenario,wang2021hierarchical,wang2022transferable}.

In the above, we have discussed three types of common feature representation. In the next two subsections, we will introduce how these features are used to model interactions among agents.

\subsection{Social Interaction Encoding via Deep Learning Layers}
\label{subsubsec:interaction_encoding}
While various neural network architectures have been proposed, the critical parts of encoding interactions are ultimately traced back to basic deep learning modules/layers. In what follows, we will introduce how different types of neural network layers are used to encode agent interactions.

\paragraph{Fully Connected Layer Interaction Encoding} The idea of fully connected interaction encoding is that all features from different agents are flattened and concatenated into a single vector and fed into a fully connected layer. Ideally, the interaction among agents is encoded and processed by the information exchange between the stacked fully connected layers. However, the interaction modeling in these architectures is usually assumed to be pretty weak and implicit due to the lack of exploiting data structures and posing inductive bias in the models. For example, since all agent's features are concatenated into a single vector, the relative spatial relationships among agents are hardly reflected. It can be challenging for the neural network to distinguish features from different agents. Also, arranging agents in different orders would lead to different outputs, which may be contradictory as we are considering the same scene and agents. Thus, the fully connected layer is frequently used to model single agents' motions and intentions \citep{hu2018probabilistic} but rarely utilized to model inter-agent interactions.

\begin{figure}
	\centering
	\includegraphics[width = 0.85\linewidth]{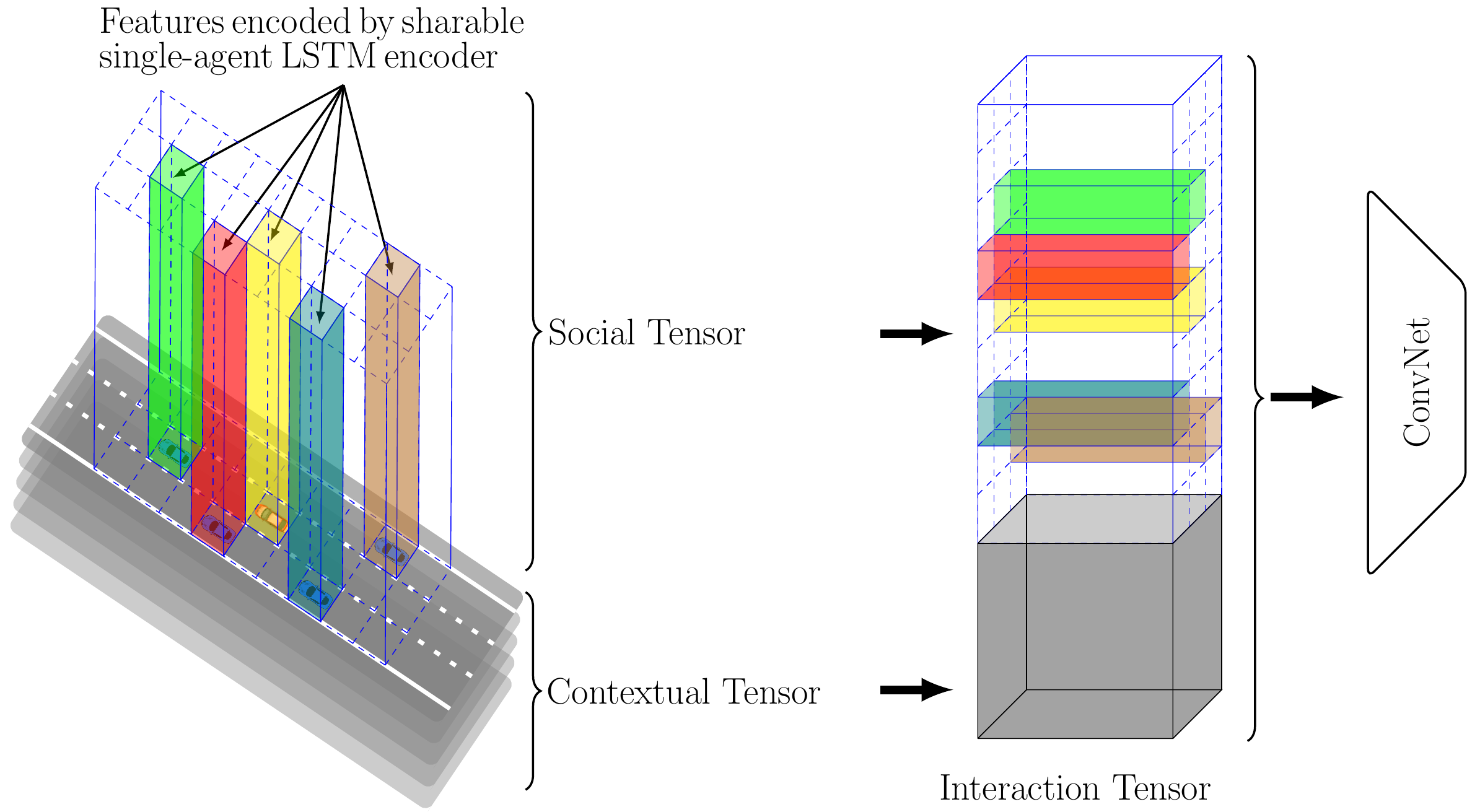}
	\caption{Illustration of a general structure of interaction encoding via convolutional operations.}
	\label{fig:socialpooling}
\end{figure}

\paragraph{Convolutional Layer Interaction Encoding} The idea of convolutional interaction encoding is feeding the spatial-temporal features (e.g., a tensor of state feature (Figure~\ref{fig:deeprepresentation}) or occupancy grid map (Figure~\ref{fig:occupygrid})) into a convolutional neural network for interaction reasoning. For example, \cite{su2021convolutions} populates the grid map with the estimated physical state of the detected agent within each grid at the current time step, as shown in Figure~\ref{fig:occupygrid}. Then, a series of convolutional layers are applied to learn the spatial inter-dependencies of agents. Instead of populating with raw features at one time step, \cite{deo2018convolutional} first encoded the individual vehicle's historic dynamics into a hidden state vector through a shared LSTM module. Each agent's hidden state vector then populates the spatial occupancy grid map, as illustrated in Figure~\ref{fig:socialpooling}, after which convolutional operations are applied.

Ideally, the convolutional interaction encoding could better leverage the spatial relationship among agents for deeper interaction reasoning. 
However, actual interaction can be local, non-local, and selective, depending on the specific driving situations:
\begin{itemize}
	\item Local interaction refers to the interaction with the ego vehicle's neighbors, the close agents in some specific directions (e.g., front, left front, and left behind) of the ego vehicle. 
	\item On the contrary, non-local (i.e., global) interaction refers to the interaction with \textit{all} the surrounding vehicles in a specific region of interest.
	\item Selective interaction refers to the interaction with relevant agents selected from certain rules \citep{niv2019learning}.
\end{itemize}
Thus it remains a question whether applying convolution operations on the spatial grid can cover sufficient interaction information and reasoning. \cite{su2021convolutions} detailedly explored how the grid map size, layout, and resolution can impact interaction modeling performance. Besides, \cite{messaoud2019non, messaoud2020attention} extended the convolutional social pooling approach by adding a non-local social pooling module with a multi-head attention mechanism to consider long-distance interdependencies. \citet{zhang2020multi} also proposed a dilated convolutional social pooling architecture to capture global context with larger receptive field.

\paragraph{Recurrent Layer Interaction Encoding} Recurrent layers is expected to better deal with temporal reasoning during interaction encoding. There are two typical schemes.

\begin{figure}[t]
	\centering
	\includegraphics[width=\linewidth]{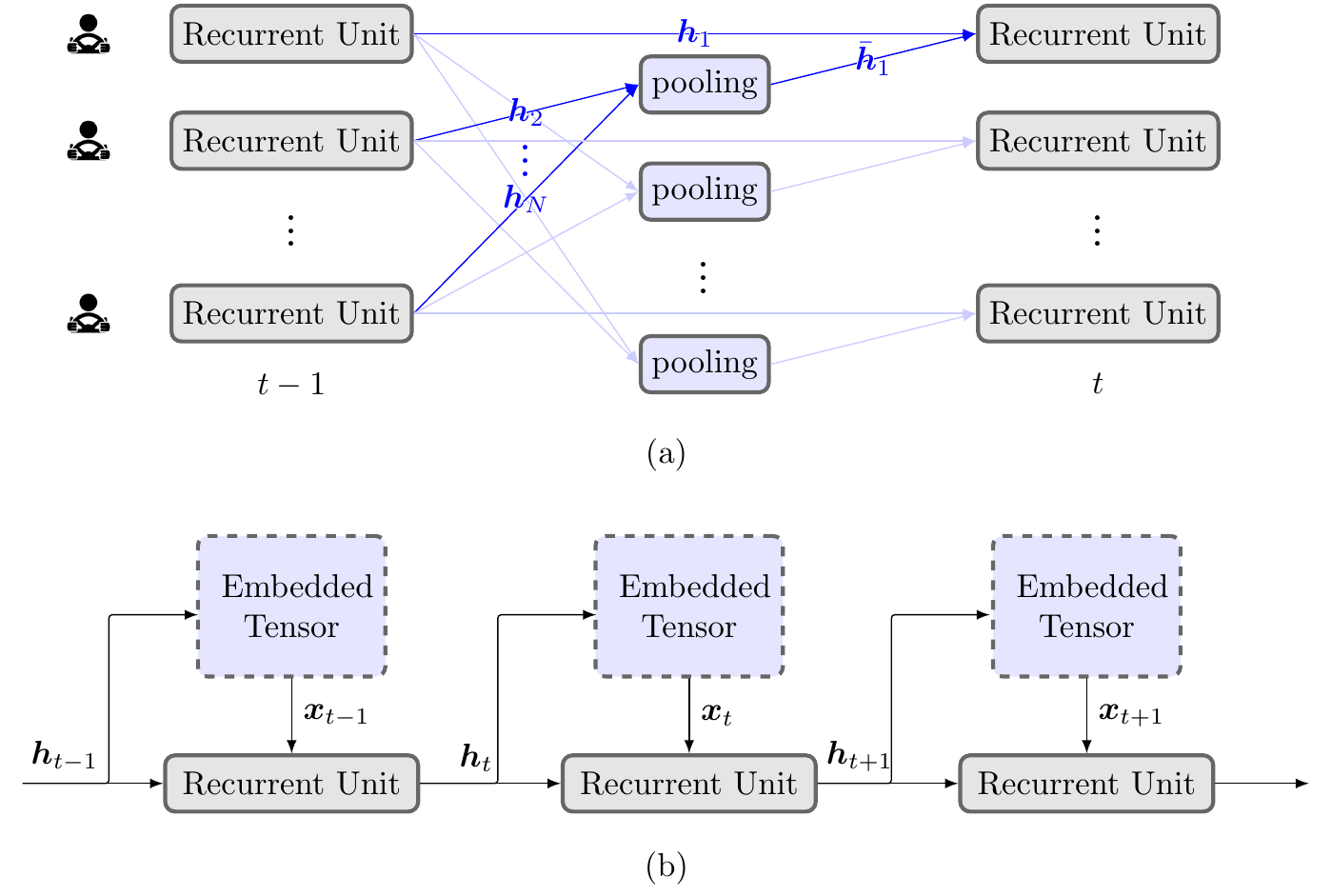}
	\caption{Illustration of interaction encoding via recurrent layers. Two schemes are shown: (a) interaction encoding in hidden states; (b) interaction encoding in input vectors.}
	\label{fig:recurrentpooling}
\end{figure}

\begin{itemize}
	\item A popular way is to encode interactions in the hidden state of recurrent units, following the concept of social LSTM proposed by \cite{alahi2016social}. As shown in Figure~\ref{fig:recurrentpooling}(a), at every time step, this approach pools the LSTM unit's hidden states for each agent with the hidden states of other adjacent agents according to their spatial relationship. Then the pooled hidden states are sent to the LSTM unit at the next time step. In practice, it is efficient to only pool the agent's neighbors with rich impacts on the ego agent \citep{deo2018multi}, instead of all agents in the scene. Depending on the applications, the pooling tricks can be max \citep{alahi2016social} or average \citep{lee2017desire}. Many works applied this or a similar schemes to predict vehicle trajectories \citep{dai2019modeling,hou2019interactive}.
	
	\item Figure~\ref{fig:recurrentpooling}(b) shows the other approach, where the interaction is processed in the input of recurrent units. At each time step, the adjacent agents' features are first concatenated into a vector tensor and then embedded. The embedded tensor at each time step is then fed into recurrent units as inputs for further temporal reasoning. In such a procedure, the embedded tensor at each time step is expected to capture the spatial information among agents, and the recurrent mechanism can process temporal information of agents. Moreover, this scheme allows additional raw sensory information of environments such as road structures. For example, \cite{lee2017desire} developed a scene context fusion unit over an RNN framework to capture the interactions. 
	
\end{itemize}

\paragraph{Graph Layer Interaction Encoding} The graph interaction encoding is expected to better handle relational reasoning in multi-agent settings. Usually, the agents are represented as the graph nodes with attributes. The relationships among agents are represented by undirected or directed edges between nodes. The nodes and edges together construct spatial-temporal graphs. The created graph is then fed into graph layers for interaction learning in a message-passing scheme, where each node aggregates their neighboring nodes' features to update their own node attributes. Such a graph interaction encoding enjoys two benefits:
\begin{itemize}
	\item \textbf{Varying number of agents.} It can deal with variable number of agents. Considering the fact that traffic interaction is usually dynamic and the number of interactive vehicles can fluctuate, interaction encoding among variable number of agents is usually necessary.
	\item \textbf{Permutation invariant.} It enables explicit relationship reasoning independent of the input ordering of the vehicle. In contrast, if we were to use a full connected layers for interaction encoding, feeding the vehicle feature into the model with different ordering would not necessarily generate the same results, which is contradictory since we were considering the same interaction scene.
\end{itemize}
In such a computing procedure, researchers usually combine the recurrent layers and graph layers to process temporal information better. For example, \cite{tang2019multiple} designed the attributes of each node as the hidden embedding of each agent's history information. Note that context information is sometimes taken into account by creating a context node \citep{li2020evolvegraph} or embedding a context feature from images \citep{salzmann2020trajectron++}. Besides, instead of directly formulating vehicles as nodes, \citet{hu2020scenario,wang2021hierarchical,wang2022transferable} formulated dynamic insertion area (DIA), a slot that vehicles can insert into. These DIAs can be extracted from the driving scene and then regarded as nodes to construct semantic graphs. In this way, the context and scene information is naturally incorporated into the graph representation, leading to easier learning and better generalizability/transferability. Attention mechanisms are also exploited to demonstrate agents' relative importance or interaction intensity \citep{hu2020scenario,wang2022transferable}. Heterogeneous encoding is also achieved by applying different embedding functions to different types of agents \citep{li2020evolvegraph}, adding an online adaptation module \citep{wang2021online}, or designing an extra category layer \citep{ma2019trafficpredict}.

\subsection{Social Interaction Encoding via Attention}
Except for the neural network layers above, another popular idea of encoding the social interaction among agents is attention -- a mechanism that quantifies how one feature influences the others, thereby representing the relationship among the features. 
In real traffic, humans drive vehicles in interactive scenarios by selectively accounting the spatial and temporal influences of other traffic agents. Each agent should pay attention to other spatially-located agents' driving behavior in history, at-present, and future. For instance, one driver tentatively changing lanes on highways would pay more attention to the vehicles in the target lane than the vehicles on other lanes. So, \textbf{how do we formulate the attention mechanism and design attention modules to capture such influences?} 

\begin{figure}[t]
	\centering
	\includegraphics[width=\linewidth]{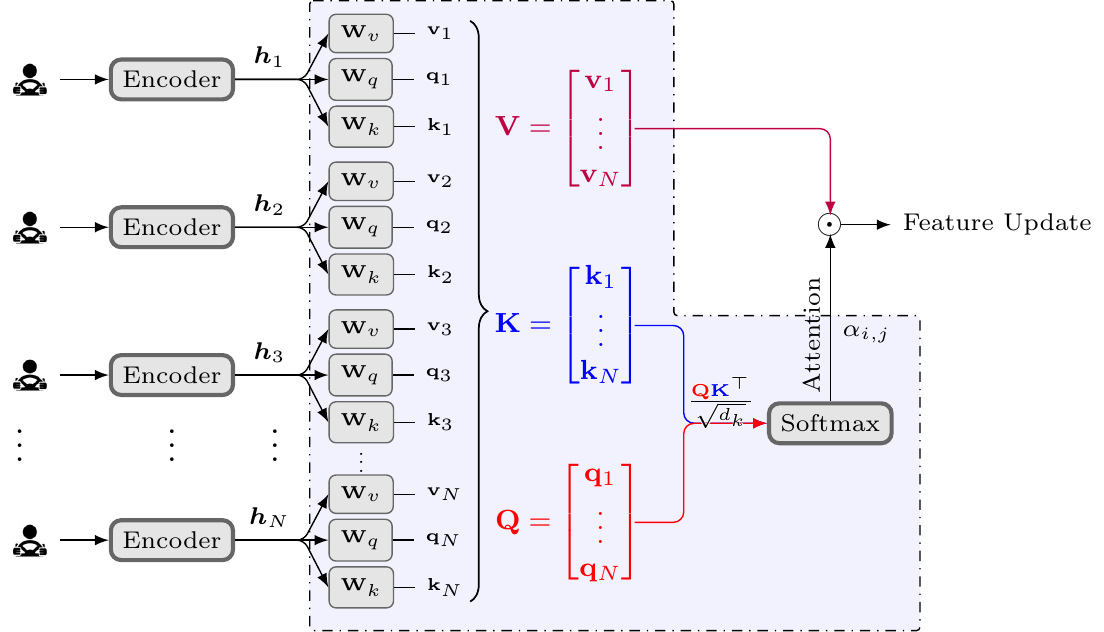}
	\caption{Illustration of the scaled dot-product attention with a single head.}
	\label{fig:dot-product-attention}
\end{figure} 

Without loss of generality, a straightforward idea is to vectorize each entity and then compute the attention level between them with certain functional measurement. Take a simplified two-agent interaction scene (agent $i$ and agent $j$ are the entities) for example, each agent's behavior is characterized by an independent vectorized feature (denoted as $\boldsymbol{h}_{i}$ and $\boldsymbol{h}_{j}$, respectively). Agent $i$ has a strong influence\footnote{Here we use the term \textit{influence} instead of \textit{interaction} since attention in nature measures the social factors of how much dependence of one agent's behavior on others. The important identified relations between agents are leveraged into interactions, forming social interaction.} on agent $j$ if the functional measurement outputs a large attention value \citep{graves2014neural,bahdanau2014neural, luong2015effective, vaswani2017attention}. In general, for $N$-agent ($N>2$) scenarios, the unified form of computing the influence on agent $i$ from the other one $j$ is defined as:

\begin{equation}
	\alpha_{i,j} =  f(\boldsymbol{h}_{i}, \boldsymbol{h}_{j})
	\label{eq:attention cal}
\end{equation}
where $i$ is the index of an agent (in \textit{spatial}, \textit{temporal}, or \textit{spatiotemporal}) whose influence is to be estimated, and $j$ is the index of agent $i$'s possible neighbors ($j\in \mathcal{N}_{i}$) in the scene. The pairwise measure function $f$ returns a scalar, representing the relationship such as affinity or influences between agents $i$ and $j$. An early and simple idea on designing the functional measurement $f$ is using dot product (see Fig. \ref{fig:dot-product-attention}) to evaluate the similarity between the pair of entities \citep{mercat2020multi,leurent2019social, wu2021hsta, guo2022vehicle,chen2022intention}. But more complicated functional measurements are also designed to quantify attentions more comprehensively \citep{wang2022transferable,hu2020scenario,li2020evolvegraph}, such as learnable general attention or concatenation attention. Appendix~\ref{app:attenion} summarizes some commonly used functional measurements. 

Note that in the examples above, each entity is representing one agent. However, under the most general definition, the attention mechanism can quantify the influence between different types of entities over \textbf{temporal space} (short-term and long-term) and \textbf{spatial space} (distant and local). We here provide five commonly used entity representation.

\begin{enumerate}
	\item \textbf{Temporal Attention.} In real traffic, human agents use information in history, at-present, and future to make decisions. For example, one vehicles' sudden brake behavior over the past one second delivers yield intention or emergency alert, which can help to predict the vehicle's future behavior. Thus, the attention mechanism in the temporal space is designed to attend to information of different time step \citep{chen2019attention}. To this end, agent $i$'s information in the time step $t$ is encoded into a hidden vector $\boldsymbol{h}_{i}^t$, usually via recurrent layers and fully connected layers. Each of the hidden vectors $\{\boldsymbol{h}_{i}^t\}$ represents one entity and the attention between them can be calculated according to Equation~(\ref{eq:attention cal}). For example, predicting vehicle $i$'s behavior for the future time step $t+1$ needs to consider its own historic behaviors in the past $r$ time step:
	\begin{equation}
		\boldsymbol{h}_{i}^{t+1} = \sum_{T=t-r}^{t} \alpha_i^{t,t'} \boldsymbol{h}_i^{t'}
	\end{equation}
	where $\alpha_i^{t,t'}$ is current time step $t$'s normalized attention on the historic time step $t'$ of the vehicle $i$. Note that such a scheme can be easily extended to temporal attention between different agents by considering the attention between vehicle $i$'s state at time step $t$ and vehicle $j$'s state at time step $t'$ \citep{wu2021hsta,ngiam2021scene}.
	
	\begin{figure}[t]
		\centering
		\includegraphics[width=\linewidth]{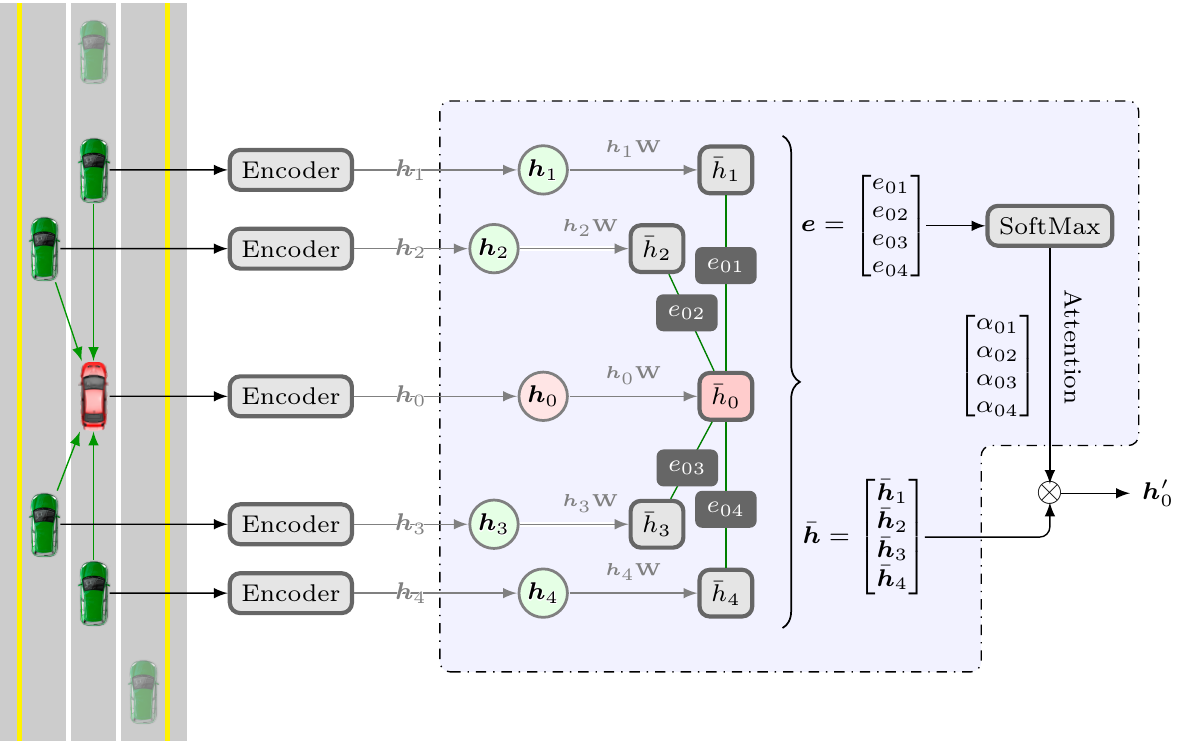}
		\caption{Illustration of generalized graph attention for capturing one agent's (red car) interaction with its neighboring agents (green cars).}
		\label{fig:graphattention}
	\end{figure}
	
	\item \textbf{Agent Pairwise Attention.} In real traffic, each agent negotiates with others while paying attention to others at different levels. Thus an intuitive representation is to formulate each agent as the entity and calculate the attention among them \citep{ding2019predicting,leurent2019social,mcallister2022control}. To this end, each agent's trajectory (or historical trajectory) is independently encoded into a hidden vectorized representation, $\boldsymbol{h}_{i}$, through a shared encoder such as RNNs. The feature of target agent $i$ is then updated as an attention-weighted sum of all other agents $j$ in the scene:
	
	\begin{equation}
		\boldsymbol{h}_{i}^{\prime} = \sum_{j} \alpha_{i,j} \boldsymbol{h}_{j}
	\end{equation}
	where $\alpha_{i,j}$ is the normalized attention weight capturing the agent $j$'s effect on the target agent $i$, calculated by Equation~(\ref{eq:attention cal}). Such an update can capture the target vehicle's relationship/interaction with its neighbor agents, considering the attention to each neighbor agent at different levels. Figure~\ref{fig:graphattention} provides an example to illustrate the encoding procedure using agent pair-wise attention. 
	
	\begin{figure}[t]
		\centering
		\includegraphics[width=0.95\linewidth]{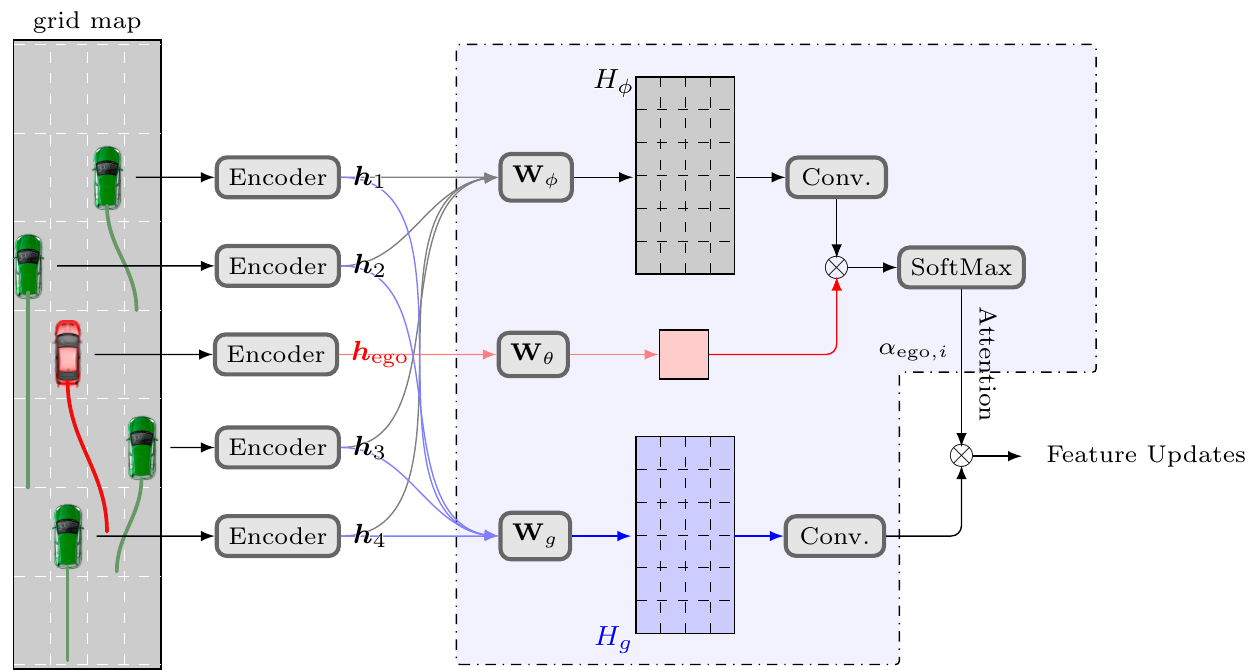}
		\caption{The structure of spatial grid-based attention for capturing one agent's (red car) interaction with its surrounding agents (green cars).}
		\label{fig:spatialgridattention}
	\end{figure}
	
	\item \textbf{Spatial Grid-based Attention.} The basic idea of spatial grid-based attention is similar to the agent-based attention, but it additionally leverages the information of the spatial grids $H$, which can help to preserve the spatial relationship between vehicles and the structured information of scene \citep{messaoud2021trajectory, chen2022intention, guo2022vehicle}. In this scheme, the scene is rasterized as a spatial grid $H$ (see Figure~\ref{fig:spatialgridattention}), say $M\times N$ grid, and each cell is populated by the encoded hidden states $\boldsymbol{h}_{j}$ of each agent:
	\begin{equation}
		H(m,n) = \sum_{j\in \mathcal{N}_{i}}\delta_{mn}(x_{j}, y_{j}) \boldsymbol{h}_{j}
	\end{equation}
	where $\delta_{mn}(x_{j},y_{j})$ is an indicator function equal to $1$ if and only if $(x_{j},y_{j})$ is in the cell $(m,n)$, $\mathcal{N}_{i}$ is the set of agent $i$'s neighbors.
	The spatial grid $H$ is then embedded via convolution layers to aggregate spatial information \citep{messaoud2019non, wang2018non, zhang2020spatial}, resulting in an embedded spatial grid $H'$.
	Thus, the influence of the $M\times N$ embedded spatial grid $H'$  on target agent $i$ is evaluated:
	\begin{equation}
		\mathbf{h}_{i}^{\prime} = \sum_{m,n}^{M,N} \alpha_{m,n} H'_{m,n}
	\end{equation}
	where $\alpha_{m,n}$ denotes the normalized attention weight between agent $i$'s hidden vector with the cell $(m,n)$ of the embedded spatial grid, calculated according to Equation~(\ref{eq:attention cal}).  
	The spatial grid form is flexible to integrate other modules to consider the influence of, for example, non-locally positioned agents in the scene \citep{messaoud2019non, messaoud2020attention} and the potentially available positions in the space \citep{ding2021ra} on the target agent. 
	
	\item \textbf{DIA-based Attention.} In dynamic scenes, drivers constantly recognize and pay attention to drivable areas/gaps, called dynamic insertion areas (DIAs) \citep{hu2018probabilistic, hu2020scenario}(see Figure~\ref{fig:DIA}), and then choose one of them to drive through. Thus, an intuitive scheme is to view DIAs as entities among which the attention is formulated. This scheme concentrates on modeling the relationship among potential DIAs, rather than among motion states of vehicles. The DIA representation can characterize the dynamic environment generally and cover diverse types of driving situations. In practice, DIAs are formed by vehicles and road marks, thus the relationship among DIAs can explicitly represent the interaction among agents and the environment. Each DIA is extracted from the scene and then encoded into a hidden vectorized $\boldsymbol{h}_{i}$ using recurrent and fully connected layers. The $i$-th DIA's feature is then updated with the attention-weighed sum of all other DIAs in the scene:
	\begin{equation}
		\boldsymbol{h}_{i}^{\prime} = \sum_{j} \alpha_{i,j} \boldsymbol{h}_{j}
	\end{equation}
	where $\alpha_{i,j}$ are the normalized attention weights, capturing the $j$-th DIA's effect on the $i$-th DIA, which can be calculated by Equation~(\ref{eq:attention cal}). Such a update is conducted on all DIAs in one graph layer, and multiple graph layers are usually exploited for better information aggregation \citep{wang2021hierarchical,wang2022transferable,hu2020scenario}.

	\item \textbf{Graph-based Attention.} In this scheme, the core idea is to treat the multi-agent interaction as a graph-structured scene --- a scene graph, $\mathcal{G}_{\mathrm{scene}} (\mathcal{V},\mathcal{E})$, composed of a set of nodes $\mathcal{V}$ and edges $\mathcal{E}$ (in spatial, temporal, or spatiotemporal). The scene graph can be generated via a graph construction process by taking (different types of) agents in the scene as nodes and influences/relationships between agents as edges (more graph-related preliminary information refers to Appendix~\ref{app:graphmodels}). Such graph-structured formulation allows injecting graph structure information into attention computation, for example, only accounting for agent $i$'s neighbors in the graph to compute the attention coefficients \citep{velivckovic2017graph}. The \textit{interaction} or \textit{relationship} among agents is reflected in updating the node features of the scene graph by accounting for the influence of other nodes \citep{salzmann2020trajectron++,li2020evolvegraph,li2019grip,cao2021spectral,kosaraju2019social}. Specifically, given a scene graph $\mathcal{G}_{\mathrm{scene}}$ and the corresponding node features $(\boldsymbol{h}_{1}, \dots, \boldsymbol{h}_{N})$ for agents in the scene, the graph attention is to learn node-wise attention and output the updated node features $(\boldsymbol{h}_{1}^{\prime}, \dots, \boldsymbol{h}_{N}^{\prime})$:
	\begin{equation}
		\boldsymbol{h}_{i}^{\prime} = \sum_{j} \alpha_{i,j} \boldsymbol{h}_{j}
	\end{equation}
	where we get node $i$'s updated feature $\boldsymbol{h}_{i}^{\prime}$ as the attention-weighted sum of neighbor agents' node features, $\alpha_{i,j}$ denotes the node $i$'s attention on the node $j$. This computation procedure is illustrated in Figure~\ref{fig:graphattention}, wherein the features of edges between the ego vehicle (agent $0$) and its surroundings (agent $i$) in the graph are denoted as $e_{0i}$. Such computation then can be conducted on all agents, and a complete form of all agents' updated output representations is
	\begin{equation}
		\begin{split}
			\begin{bmatrix}
				\boldsymbol{h}^{\prime}_{1} \\
				\boldsymbol{h}^{\prime}_{2} \\
				\vdots\\
				\boldsymbol{h}^{\prime}_{N} \\
			\end{bmatrix}
			& =  \underbrace{
				\begin{bmatrix}
					\alpha_{1,1} & \alpha_{1,2} & \cdots & \alpha_{1,N} \\
					\alpha_{2,1} & \alpha_{2,2} & \cdots & \alpha_{2,N} \\
					\vdots & \vdots & \ddots & \vdots \\
					\alpha_{N,1} & \alpha_{N,2} & \cdots & \alpha_{N,N} \\
			\end{bmatrix}}_{\mathrm{Attention \ Matrix}}
			\begin{bmatrix}
				\boldsymbol{h}_{1} \\
				\boldsymbol{h}_{2} \\
				\vdots\\
				\boldsymbol{h}_{N}
			\end{bmatrix}
		\end{split}
	\end{equation}
	As one can tell, such a graph-based attention mechanism is closely related to the agent pairwise attention mechanism. The major difference is that the graph-based attention mechanism 1) emphasizes the graph properties; 2) conducts feature update on all agents instead of only on the target agent for better information aggregation; 3) is usually combined with multiple graph layers to conduct multiple aggregations. 
\end{enumerate}

\subsection{Model Parameter Learning}
The above sections demonstrate how deep learning modules and attention mechanisms can be used to model the interactions (or influences/relationships) among agents. After these modelling, a training process is required to calibrate model parameters using data, which is often achieved with gradient-based optimization algorithms. According to the objective of the task, different loss functions are designed to back-propagate and adjust the model parameter. For example, for a driving behavior prediction task, the loss function usually minimizes the error between prediction and ground truth \citep{wang2021hierarchical,salzmann2020trajectron++}, while for a reinforcement learning task, the loss function encourages the agent to get higher rewards \citep{schmidt2022introduction,chen2019attention}. Besides, the loss function design also depends on the representation of the model output. Many existing works have the model output deterministic values, such as the predicted positions or actions \citep{alahi2016social,vemula2018social,jain2016structural}, where the training procedure is regarded as a deterministic regressor. There are also methods taking generative probabilistic approaches, outputting a probability distribution for these actions and intentions \citep{deo2018convolutional,hou2019interactive,lee2017desire}. Then, the training procedure is a probability maximization process.

\subsection{Summary}

Deep learning-based approaches provide a flexible network architecture for \textit{representing} and \textit{learning} interactions among traffic agents, benefiting from the modularized layers and abundant data nowadays \citep{krajewski2018highd, interactiondataset, krajewski2020round, bock2020ind, exiDdataset, yao2019egocentric, kang2019test, wang2017much}. Deep learning-based approaches have already shown their power and promise by occupying the top ranks of many driving-related challenges, competitions, and leaderboards. In the future, for safe and large-scale deployment on real autonomous driving systems, many challenges still remain, such as (i) improving interpretability while guaranteeing performance and (ii) enhancing the generalizability in diverse driving entities, scenarios, and situations. One safety-guaranteed way for deep learning could be realized by using the output of a RNN in the cost function of a trajectory optimization method \citep{brito2022learning}.

\begin{figure}[t]
	\centering
	\includegraphics[width=0.8\textwidth]{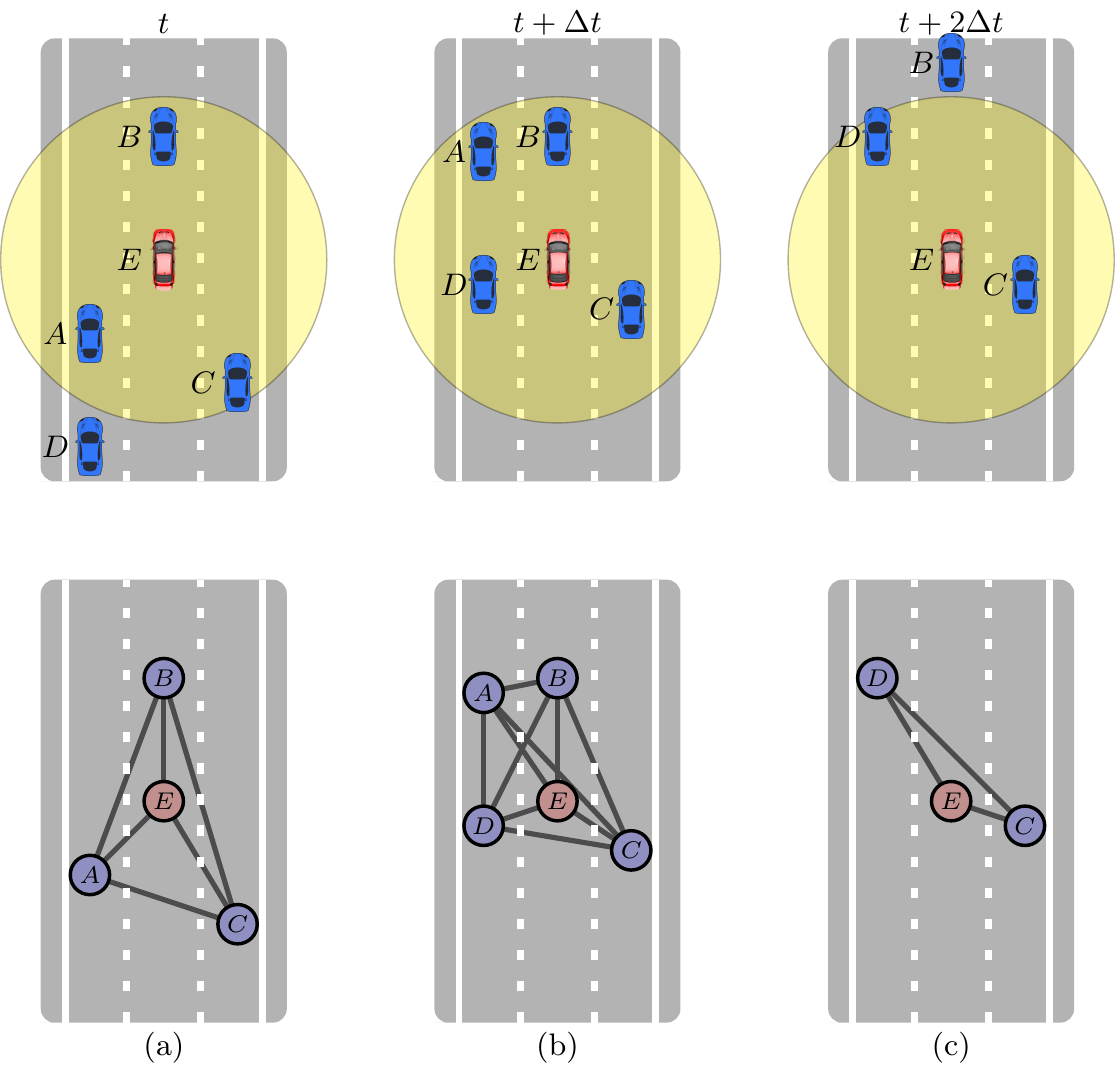}
	\caption{Illustration of the evolution of a complete dynamic graph of vehicle-to-vehicle interactions in a driving task: (a) $\rightarrow$ (b) represents adding nodes and edges, and (b) $\rightarrow$ (c) represents removing nodes and edges. The yellow shadow circle is the Region of Interest (RoI) hand-crafted by researchers, and only the vehicles inside of it are interactive agents.}
	\label{fig:graphs}
\end{figure}

\section{Graph-based Models}
\label{subsec:graph_based_model}

\paragraph{Why Graphical Models?} Interactions among road users in daily traffic scenes are structured; they dynamically change along with the spatiotemporal space with uncertainty, in which the interrelated parts (e.g., agents, road lines, obstacles) are organized and evolve with constraints of traffic rules and social norms. With fact, graph-based models provide a natural tool for dealing with the complexity and uncertainty of interaction behaviors \citep{murphy2012machine}. Graphical models provide an intuitively appealing interface by which researchers can model highly-interacting sets of variables and the data structure. Moreover, graph-based models are usually explainable to  represent the structured relationship between human agents. Using graphical models enables us to envisage new models tailed for a particular environment. For example, graph neural networks assign their nodes (vertices) as human agents \citep{girase2021loki}, instances \citep{ma2019trafficpredict}, or decision-related states \citep{hu2020scenario, wang2021hierarchical}, and the edges (arcs)\footnote{If the graphical network is undirected, i.e., for every connection from each pair nodes of $i$ to $j$ has a connection node $j$ to node $i$, the links are called edges. Otherwise, directed connections are termed arcs. More detailed definitions of graph models refer to Appendix~\ref{app:graphmodels}.} between nodes as the interactions between them, such as the agents' velocities and relative positions \citep{girase2021loki}. Figure~\ref{fig:graphs} provides a case of graphically representing the interactive relationships among agents with behaviors evolved over time. In what follows, we will mainly introduce three graph-based approaches to model interactions in traffic scenes: graph neural networks (Section~\ref{subsubsec:GNN}), Bayesian networks (Section~\ref{subsubsec:bayesian networks}), and topological models (Section~\ref{subsubsec:topology}).

\subsection{Graph Neural Networks (GNNs)}
\label{subsubsec:GNN}
A GNN-based model for capturing traffic agents' interaction could have different names such as \textit{interaction graph}, \textit{agent graph}\footnote{In \cite{cao2021spectral}, the authors distinguished the \textit{environment graph} that encodes the traffic context information and the \textit{agent graph} that captures the interactive behavior of human agents.} \citep{cao2021spectral}, \textit{scene graph} \citep{girase2021loki}, or \textit{traffic graph} \citep{chandra2020stylepredict, chandra2020cmetric, kumar2020interaction} in different works. In this paper, we collectively called these names \textbf{interaction graphs} for notational convenience. In the model, $N$ agents in the interaction scenario are usually represented by a graph with $N$ agent nodes (also called vertices) and $N\times N$ edges characterizing their interactions with each other. The interaction graph can have some node attributes (e.g., human drivers' states) and edge attributes (e.g., relations between human drivers). In addition to viewing human agents as graph nodes, other features such as the potential target positions (e.g., the insertion areas between vehicles ) can also be treated as nodes of a graph \citep{ding2021ra}. Before introducing the works of utilizing GNN-based models to capture the interactions between human drivers, we need to define graphs and networks clearly. Appendix~\ref{app:graphmodels} summarizes the definitions of graphs, digraphs (also called directed graphs), and networks (node weighted or edge weighted graphs, or both). 

The designed graph can be static or dynamic. For instance, \cite{li2019grip} formulated the interactions between vehicles via a static graph convolutional model, consisting of convolutional layers to independently capture the temporal information (i.e., motion pattern) of each vehicle  and graph operations to capture the inner-relation between vehicles over spatial space. \cite{cao2021spectral} built a \textit{fully} connected dynamic graph to model agents' interactions by adding a weighted edge for each pair of agent nodes. However, assuming all vehicles within a predefined region of interest have interactions with each other and adding edges for each pair of vehicles (i.e., creating a fully connected network between vehicles) might make models over complicated.

With the benefit of dynamic graphs, the model can leverage different interactions with associated structures of graphs and different model inputs. Before creating an efficient and effective interaction graph, some related questions should be in mind, including
\begin{itemize}
	\item How to enable the interaction graph to capture agents' dependencies over spatial and temporal spaces?
	\item How to make the learned interaction graph transferable?
	\item How to make the interaction graph embraceable for heterogeneous agents in traffic?
	\item How do we integrate environmental information into the interaction graph?
	\item How can the interaction graph be compatible with an arbitrary number of agents?
\end{itemize}
In what follows, we will discuss the essential questions raised above.

\begin{figure}[t]
	\centering
	\includegraphics[width=0.9\linewidth]{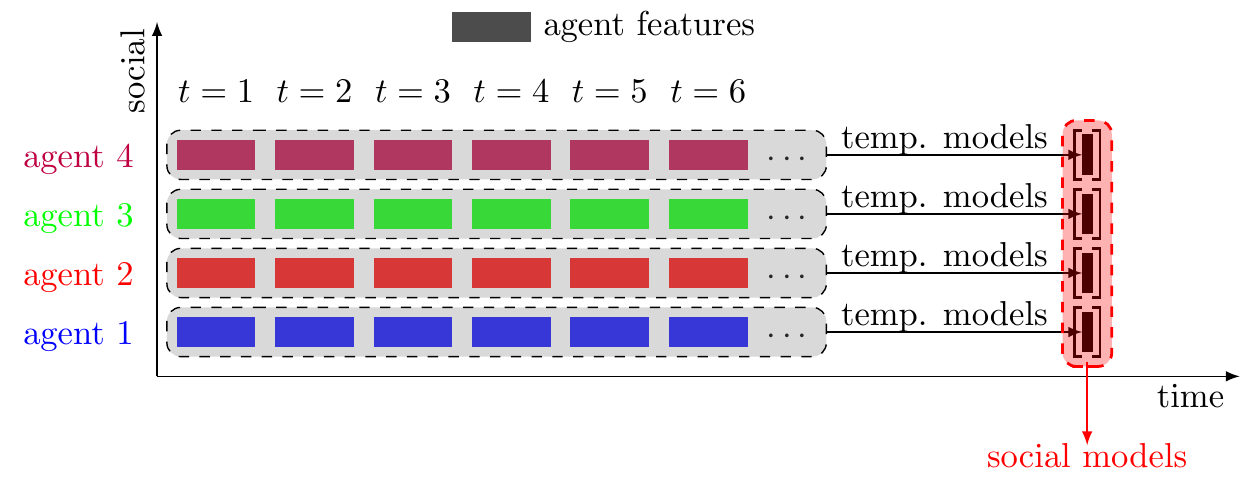}
	\caption{The typical structure of social-temporal models for multiple agents.}
	\label{fig:social_temporal}
\end{figure}

\paragraph{Spatiotemporal Information} Human agents' dependencies exist in two dimensions: one is the time dimension, and the other one is the spatial dimension (also called the social dimension). The time dimension mainly encodes how the past states of human agents influence their current and future states, while the social dimension encodes how one agent's states affect others in the future. The historical information is usually encoded via an LSTM network \citep{hochreiter1997long}, RNN \citep{chung2014empirical}, Transformers, or attention \citep{vaswani2017attention, li2020end} to human drivers' interactions \citep{li2019grip, deo2018convolutional, ma2019trafficpredict, girase2021loki, ma2021multi,zhao2020gisnet, roy2020detection, hou2019interactive} in the time dimension. To learn the long-term interval interaction patterns, operating a graph convolution in the spectral domain will eliminate the temporal dependency at different time steps in the time domain, thus capturing the global information of history trajectories \citep{cao2021spectral}. Besides, the attention mechanism \citep{vaswani2017attention} can capture and learn the interactions between agents in the past time domain \citep{song2021learning}. For instance, \cite{messaoud2020trajectory} applied multi-head attentions to a combined interaction tensor representation (i.e., agent-agent and agent-map tensor representations) to capture the interaction between human drivers. \cite{li2020social} utilized a graph double-attention network over spatiotemporal graphs to generate abstract node attributes representing the interaction between human drivers.

In the social domain, the interaction refers the spatial interdependencies between agents and can be computed by pooling \citep{deo2018convolutional, lee2017desire} and convolutional \citep{li2019grip, zhao2020gisnet} techniques. Usually, researchers first use temporal models to summarize trajectory features
over time for each agent \textit{independently} and then feed the
temporal features to social models to obtain socially-aware (i.e., spatial-interaction)
agent features \citep{hou2019interactive,choi2019drogon,cho2019deep}. Figure~\ref{fig:social_temporal} illustrates the inputs and outputs of the used temporal models and then social models cascadingly. The spatial information between human drivers is integrated at the \textit{last} step by abandoning the previous interactions. This approach is sub-optimal because independently encoding features from either time or social domain may result in loss of information. In the real world, one human driver's behavior at the current time step would influence other human drivers' behavior at a future time. \cite{yuan2021agentformer} proposed an agent-aware transformer (AgentFormer) to simultaneously model the interaction in both time and social domains by introducing a time encoder and a new agent-aware attention module. \cite{he2020ust} tackled this problem by unifying the \textit{spatial} and \textit{temporal} context into one single high-dimensional representation in a 3-D space spanned by 2-D location and time. Besides, the information encoded  between human drivers should be shareable over the time-space. Using the graph Fourier transformer \citep{kipf2016semi} can realize the encoding by taking the graph convolution operations in the spectral domain, such as in the Fourier space \citep{cao2021spectral, zhao2020gisnet}.

\paragraph{Transferable Interactions} Human drivers can conduct efficient and effective negotiations with surroundings in new traffic scenes in light of their cognitive ability to abstract complex scenarios into simple but efficient \textit{representations} that can be stored and reused in new scenarios. By this, \cite{hu2020scenario} proposed a semantic graph reasoning framework to learn the interactions between human drivers by developing a generic representation of the dynamic environment with semantics. \cite{wang2021hierarchical} then extended this concept. Besides, \cite{ma2021multi} claimed that the correct feature representation of interaction is the key to capturing human's ability to predict surroundings' behaviors in many different seen and unseen scenarios. The authors utilized self-supervised learning techniques to attain effective representations of interactions at the Fren\'{e}t coordinate systems for the graph's node and edge features.

\paragraph{Heterogeneous Agents} In natural traffic scenes, especially in urban environments, the interaction often occurs between different human agents (i.e., human drivers, cyclists, and pedestrians), forming a heterogeneous interactive traffic condition. Different types of human agents have different dynamic properties and motion rules, and only human agents in the same type would share the behavior attributes. Inspired by this, \cite{ma2019trafficpredict} proposed a heterogeneous traffic graph consisting of two layers. One is the \textit{instance layer} to capture the dynamic attributes of each type of agent (i.e., vehicles, cyclists, and pedestrians) using LSTM separately, and the other is the \textit{category layer} to measure the similarity between instances. Similarly, \citep{li2021hierarchical} designed category and instance layers to tackle the interaction among heterogeneous agents. \cite{girase2021loki} designed an information-shareable graphic framework for different agents using a daisy-chained process via the attention mechanism and claimed that the message passing with such a frame allows agents to share their past trajectory, goal, and intention information. However, it fails to justify what is the goal or intention specifically. Researchers also claimed that the same types of agents (e.g., vehicles, pedestrians) would share an identical encoder and designed different encoding channels. Thus, the interaction between heterogeneous agents is modeled via a directed edge-featured graph \citep{mo2021heterogeneous} with an integrated graph attention mechanism \citep{vaswani2017attention}. Besides, the interactions between heterogeneous agents can be formulated using the social tensor (Figure~\ref{fig:socialpooling}) by aggregating the generated embeddings of each agent \citep{chandra2019traphic,ivanovic2021heterogeneous}. 

\paragraph{Interactions with Environmental Information} Traffic physical constraints influence the interactions between human drivers. For instance, an unprotected left turn vehicle should yield to oncoming traffic while two spatially nearby vehicles driving on opposite lanes barely, even no, interact with each other. Therefore, map (or traffic contextual) structures are critical to modeling and learning interactions. A general pipeline to leverage map information into interactions encodes the map into a feature vector based on a rasterized map and aggregates it with a social tensor to shape a high-dimensional feature representation \citep{messaoud2021trajectory, zhao2019multi, hong2019rules, wang2021multiple, bahari2021injecting, song2021learning, gao2020vectornet}, as summarized in Fig. \ref{fig:socialpooling}. However, representing the map/contextual information of a large region of space with a single feature vector is difficult. To overcome this issue, a local-global hierarchical structure, i.e., a local layer and a global layer, is usually used to represent rich structural information of traffic context and maps. \cite{zeng2021lanercnn} proposed a graph-centric motion forecasting model (LaneRCNN) to learn a \textit{local} lane graph representation for each human agent and the \textit{global} topology of local maps. The interaction of human-driven vehicles with road infrastructures (e.g., lanes and traffic signs) can be measured using the Euclidean features (e.g, semantic grid maps, \cite{djuric2018short}) or non-Euclidean features \citep{pan2020lane}. Similarly,  \cite{zhang2021trajectory} proposed a dual-scale approach to modeling traffic context by integrating the geometrical (i.e., local) and topological (i.e., global)  features. The inter-layer (local-global layers) connectivity is then evaluated via attention \citep{vaswani2017attention, zhang2021trajectory} or information propagation via convolution operation \citep{zeng2021lanercnn}.  

\begin{figure}[t]
	\centering
	\includegraphics[width = 0.34\linewidth]{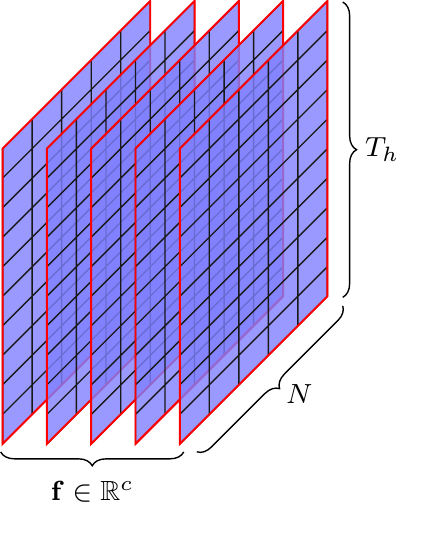}
	\caption{The tensor-input of spatio-temporal graphs. Each plane represents the extracted features, the row of each plane represents the states of one agent over a fixed time horizon $T_{h}$, and the column represents the states of $N$ agents at a specific time.}
	\label{fig:graphillustration}
\end{figure}

\paragraph{An Arbitrary Number of Agents} The nature of GNNs has limitations on the number of interaction agents. The number of surrounding vehicles  should usually be fixed over space and time. However, the setting is counterfactual. Figure~\ref{fig:graphs} illustrates one common scenario wherein different surrounding vehicles interact with the ego vehicle over time and space. Directly applying graphic representations with agents as nodes and relationships as edges would lead to varying-structural graphic models that are computationally intractable. Three ways can overcome this problem to adapt the dynamic interactive scenarios. 
\begin{itemize}
	\item \textbf{Grid representation.} The most straightforward way is to mesh the environments (or the world fixed with the ego vehicle) into a grid-based representation of which the cell would be assigned to nonzero state values if agents occupy it \citep{deo2018convolutional,messaoud2019non, messaoud2020trajectory}. This approach is compatible with the social-pooling techniques (Figure~\ref{fig:socialpooling}) and adaptable to scenarios with an arbitrary number of agents. However, it is sensitive to the world's granularity level and requires high-computational resources.
	
	\item \textbf{Agent selection.} Another way is to simplify the interaction scenarios by selecting a fixed or limited number of target vehicles while neglecting others. All these fixed or selected vehicles' states shape one tensor (as shown in Figure~\ref{fig:graphillustration}) as the inputs of neural networks over space and time. Agents can be selected by \textit{manually} defining the relevant vehicles with the prior knowledge of interaction scenarios \citep{cao2021spectral} or \textit{automatically} learning which agents are more relevant to the decision-making, such as through an attention-based architecture \citep{chen2020midas}. Attention mechanisms allow focusing on the surrounding agents that are highly relevant to their maneuvers, thus enabling them to handle an unordered, arbitrary number of agents in the ego vehicle's observation range.
	
	\item \textbf{Agent adaption.} The last way is to use a general functional metric to identify the interaction types. \cite{chandra2020stylepredict, chandra2020cmetric} developed a new scalar-value measure, CMetric, to quantify the aggressiveness level of driving behaviors with computational graph theory and social traffic psychology. Compared with grid representation and agent selection, the CMetric seems more robust and can directly operate on raw sensory trajectory data automatically without requiring manual adjustments such as parameter-tuning. 
\end{itemize}

\subsection{Bayesian Dynamic Models}
\label{subsubsec:bayesian networks}

\begin{figure}[t]
	\centering
	\includegraphics[width=\linewidth]{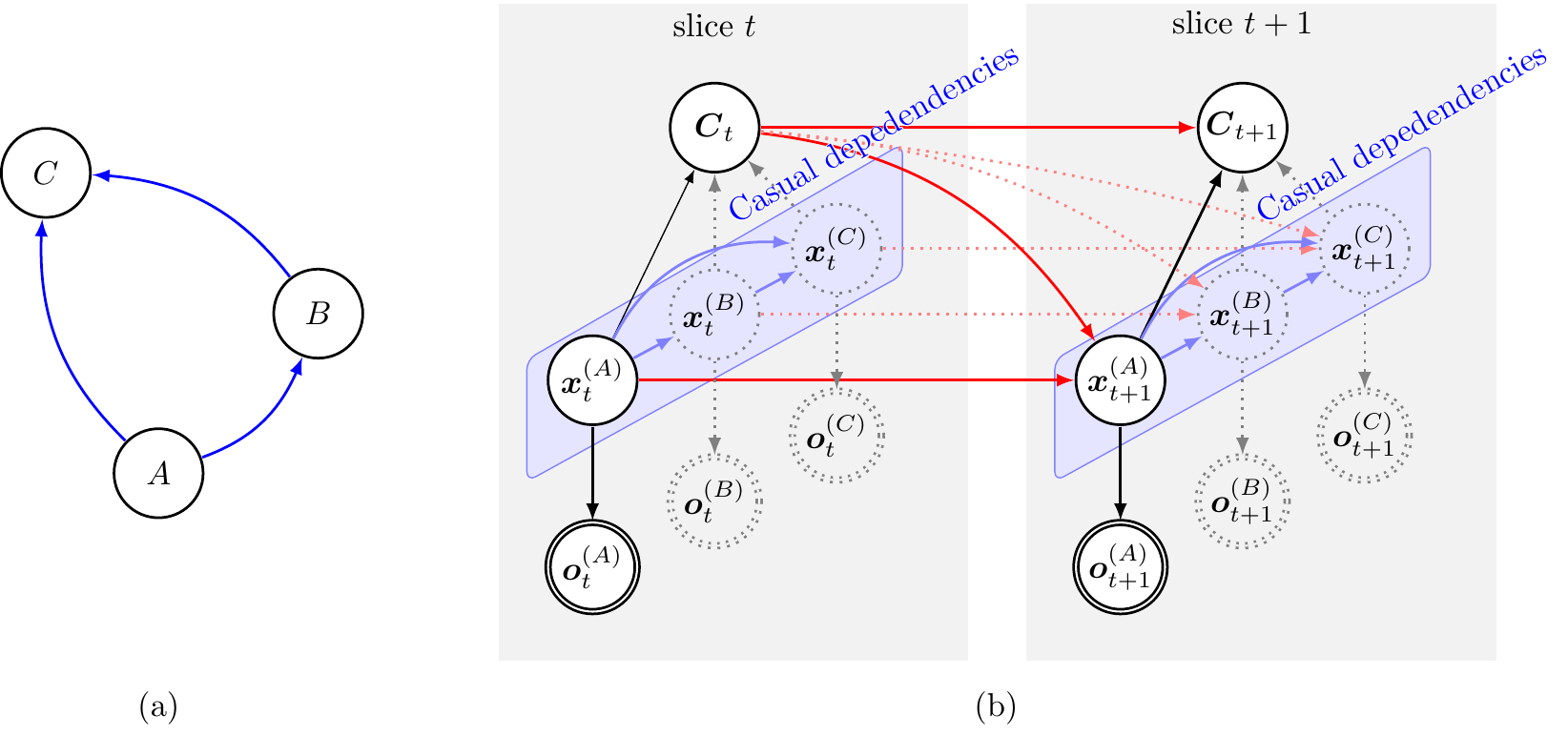}
	\caption{Illustration of (a) vanilla Bayesian networks with a directed acyclic graph for casual dependencies among agents (e.g., agents $A$, $B$, and $C$) and (b) dynamic Bayesian networks for casual and temporal dependencies through two adjacent time slices. At a single time slice, all the agents could generate the shareable context ($\boldsymbol{C}_{t}$) that is treated as the conditions for their states ($\boldsymbol{x}_{t+1}$) at the next time slice, measured as observations ($\boldsymbol{o}_{t+1}$) through sensory techniques. Double circles represent the observations and single circles represent the random variables; directed edges represent stochastic relationships (dependencies) between associated random variables.}
	\label{fig:DBNs}
\end{figure}

Bayesian networks (BN; \cite{murphy2002dynamic}) are a type of probabilistic graphical model (PGM; \cite{koller2009probabilistic}). Typically, PGM is served as a standard graphical tool to visualize large probability distribution with a high degree of structure \citep{bahram2016combined}. Unlike the GNN-based interaction model,  the nodes of BNs represent the random variables (e.g., the agent's states or sensory information with noise or discrete maneuvers), and the edges represent the \textit{scholastic causal dependencies} between nodes in conditional probability distributions. These dependencies can be schematically represented using a directed acyclic graph, and the structure of the dependencies often reflects the underlying hierarchical generative process, similar to conditional behavior prediction \citep{tolstaya2021identifying, tang2022interventional}. 
Vanilla BNs provide a probabilistic graph architecture that explicitly describes the \textit{causal dependencies} among agents at a single time slice \citep{bahram2016combined} but excludes \textit{temporal dependencies}, as shown Figure~\ref{fig:DBNs}(a).

\paragraph{Parametric Bayesian Dynamic Models} Dynamic Bayesian networks (DBNs), as a type of Bayesian dynamic model, are BNs that include temporal dependencies of nodes through, briefly stated, repeating the dependency structure of vanilla BNs over time, as illustrated in Figure~\ref{fig:DBNs}(b). Some dependencies of DBNs' nodes across these time-slices are usually recognized as temporal \textbf{dynamics} of interactions. DBNs are appropriate for agent interactions as a general framework to build up probabilistic models that describe dynamic processes with uncertainty.  In applications, the temporal dependency in a DBN is often realized either as a deterministic recursive process (e.g., in RNNs) \textit{or} as a first-order Markovian process (e.g., state-space models (SSMs), \cite{gonzalez2017interaction,li2018generic}). With this view, some SSMs and RNNs can be treated as special cases of DBNs.

In real traffic, interactive behaviors of road users are nonlinear and context-dependent and evolve through time. The observed sensory states are the generations and realizations of their inner models (e.g., latent/hidden states) that are usually unmeasurable. \cite{agamennoni2012estimation} formulated the agent interactions using a probabilistic joint distribution over all relevant states composed of dynamic states and the context classes of vehicles. The value estimation of any states from data boils down to performing statistical inference under the framework of probabilistic models.  Similarly, \cite{gindele2010probabilistic} applied DBNs to explicitly model the dependencies and influences between situation context (i.e., discrete behaviors) and continuous trajectories. Besides, \cite{gindele2015learning} also proposed a hierarchical dynamic Bayesian model to capture the interactions between environments, road networks, and traffic participants. This model can describe the physical relationships and the driver's behaviors and plans. \cite{gonzalez2017interaction} applied a switching SSM to represent the interactions among human drivers with a factorized term $p(\boldsymbol{x}_{t}^{\mathrm{target}}| m_{t-1:t}^{\mathrm{target}}, \boldsymbol{x}_{t}^{1:N})$, to describe the dynamic evolution of the target's state ($\boldsymbol{x}_{t}^{\mathrm{target}}$) given the distribution over maneuvers at current and previous time steps ($m_{t-1:t}^{\mathrm{target}}$), and the previous states of all vehicles ($\boldsymbol{x}_{t}^{1:N}$). 
\cite{xu2018aware} built a probabilistic dynamic model based on an architecture of DBNs to capture the evolution procedure of the human driver state encoded as a set of discrete latent variables when interacting with a leading car. 
\cite{tang2022interventional} investigated a coherent interactive prediction and planning framework with a dynamic Bayesian network in which the planned trajectory of the ego agent is treated as an intervention to other agents. 

The BN-based interaction models are a powerful tool to capture the complex interactions in a structurally explicit way of embracing physical constraints, unobservable states, and uncertainties of observations. Usually, using DBNs to model interactions allows reliable longer-term  prediction because they consider the mutual influences (or dependencies) between human drivers' motions and decisions dynamically. However, the model performance strongly depends on the correctness of the model assumptions. For example, a DBN-based model with a risk-minimizing assumption would not correctly capture the conditional behaviors in an actual dangerous traffic situation. On the other hand, DBNs would suffer computational complexity, which usually grows exponentially with the number of agents involved in the interactive scenes. Thus, DBNs are currently conducted with offline evaluations and laboratory testing based on existing datasets.

\paragraph{Nonparametric Bayesian Dynamic Models} Understanding multiple traffic agents' interaction processes offers significant advantages for algorithmic design. The social insights of the interactive process could be asked by
\begin{itemize}
	\item What are the fundamental representations (also known as modes, components, blocks, patterns, or primitives) of interactions at the behavioral-semantic level?
	\item How does the interaction behavior dynamically evolve over space and time?
	\item What are the fundamental topological modes constituting dynamically interactive traffic scenarios of multiple road users?
\end{itemize}
Revealing the interaction process can be initially conducted straightforwardly using traditional statistical analysis over temporal space statically \citep{wang2021social}, which fails to consider the dynamic procedure. Moreover, the high dimension and large scale of collected observations over spatial and temporal space make it intractable to do manual analysis. Fortunately, advanced machine learning techniques automatically allow us to do informative analysis for complicated multi-agent interactions with less effort. There are two types of graph-based approaches to understanding the interaction procedure: Bayesian nonparametric dynamic networks and topological methods. We first introduce the former in the paragraph immediately following and then the latter in Section~\ref{subsubsec:topology}.

\begin{figure}[h!]
	\centering
	\includegraphics[width = \linewidth]{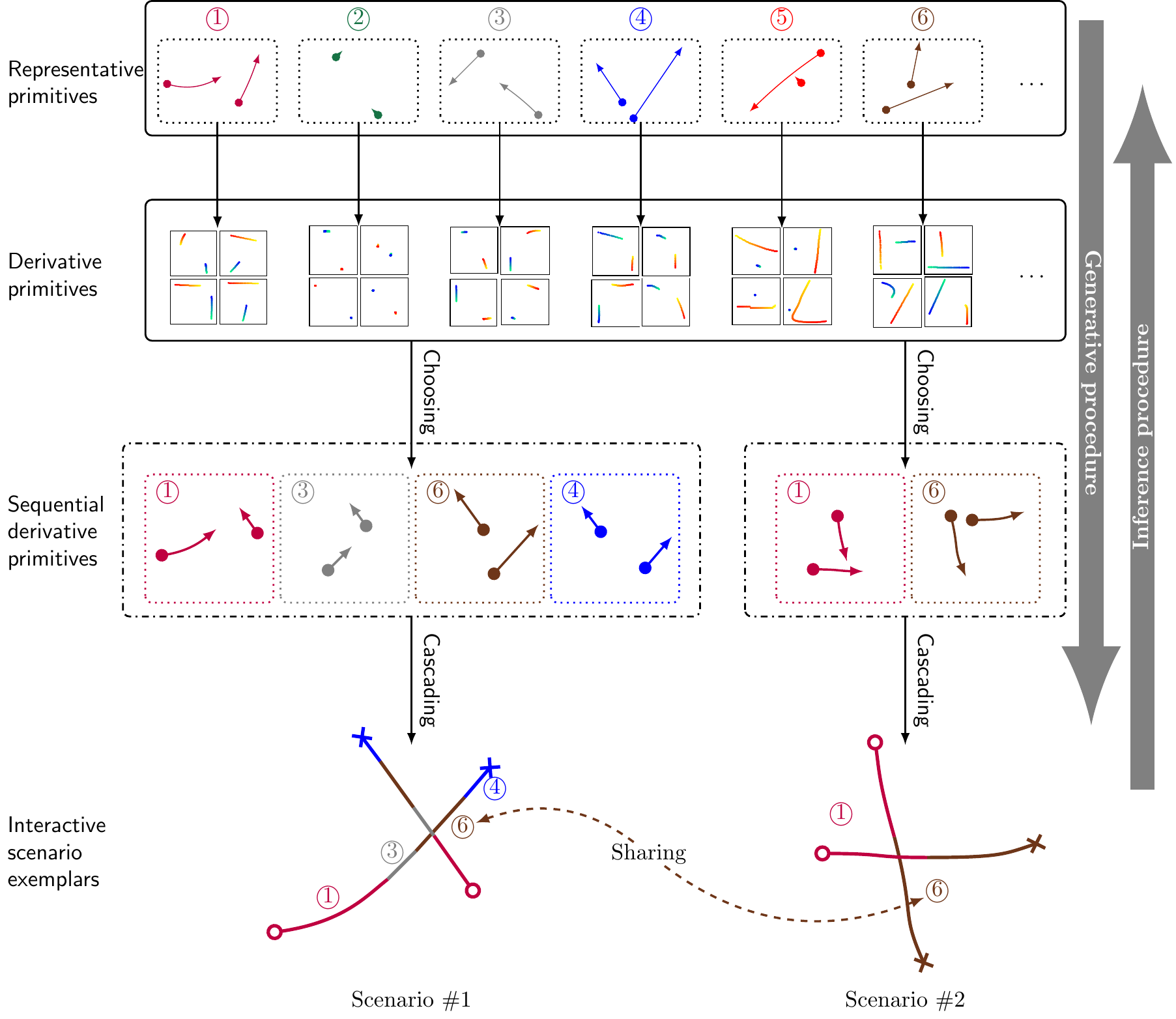}
	\caption{An example of the procedure of generating complex interaction behaviors (e.g., between two human drivers) from driving primitives of moving trajectories. Each color represents one type of primitives. Dots denote the start point of primitives, and arrows indicate the moving directions. Circles and crosses denote the start and end points of the generated interactive scenarios, respectively. Black arrows with solid lines represent the generative procedure.}
	\label{fig:NPBL}
\end{figure}

DBNs describe interactions using the \textit{predefined} and \textit{fixed} spatial edges of dynamic networks of agents that evolved over spatial-temporal space dynamically. In contrast, the nonparametric Bayesian dynamic networks can model interactions but do not require specific prior knowledge, such as the fixed number of interaction patterns. To implement the networks efficiently, researchers view the interactive behaviors as a sequential decision-making process over a countable infinite set of building blocks, also called driving primitives, traffic primitives \citep{wang2018extracting}, and the `driving DNA' \citep{fugiglando2017characterizing} akin to the essential units (e.g., DNA and molecular) for organisms. These primitives represent interaction at a low granularity level over temporal space. They allow the analysis of diverse interactions in an explainable way and summarize the complex behaviors across different interactive scenarios.

On the other hand, theoretically, any complex interaction behavior is generated from or reconstructed by some primitives at different granularity. For example, changing lanes can be decomposed into several sequential discrete primitives at the operation or tactical level \citep{atagoziyev2016lane, do2017human}. Figure~\ref{fig:NPBL} conceptually illustrates the basic idea of generating complex interaction behaviors from a library of driving primitives (represented by colors) in a hierarchical architecture. The top-to-bottom framework represents the generative procedure, which is detailed as follows. First, the basic information of \textsf{representative primitives} (i.e., primitives \textcircled{1}, \textcircled{2}, ...) is given in a priori, such as their distributions and transition probabilities from one type of primitives to others. We can then carefully draw (e.g., via sampling) their corresponding sets of derivations (called \textsf{derivative primitives}) from these representative primitives with such information. The number of representative primitives could be finite or countably infinite. Second, we choose one set of the derivative primitives at random with probability given by their distributions, for example, choosing the set  of primitives {\color{purple} \textcircled{1}}. Then, we \textsf{choose} an instance from corresponding group of derivative primitives. We choose the following primitives according to the transition probabilities $p(\cdot|\textcircled{1})$ using the already instantiated primitives {\color{purple}\textcircled{1}}. For example, the next selected primitives could be {\color{gray}\textcircled{3}} or {\color{auburn}\textcircled{6}}, corresponding for the generated Scenario \#1 or Scenario \#2, respectively. Third, we cascade the obtained pool of derivative primitives to build up the exemplars of interaction scenarios.  Driving scenarios are finally generated through a hierarchical generative framework, enabling the information of these primitives (e.g., primitive {\color{auburn}\textcircled{6}}) to be shared among different driving scenarios. The driving primitives shared across different scenarios could be realized via a global discrete stochastic process, such as by adding a hierarchical Dirichlet process (HDP, \cite{teh2006hierarchical})  to the dynamic networks.  Inversely, the bottom-to-top procedure represents the inference procedure that enables us to recognize primitives from raw driving sequential data with sampling and optimization techniques.

\begin{figure}[h!]
	\centering
	\includegraphics[width = \linewidth]{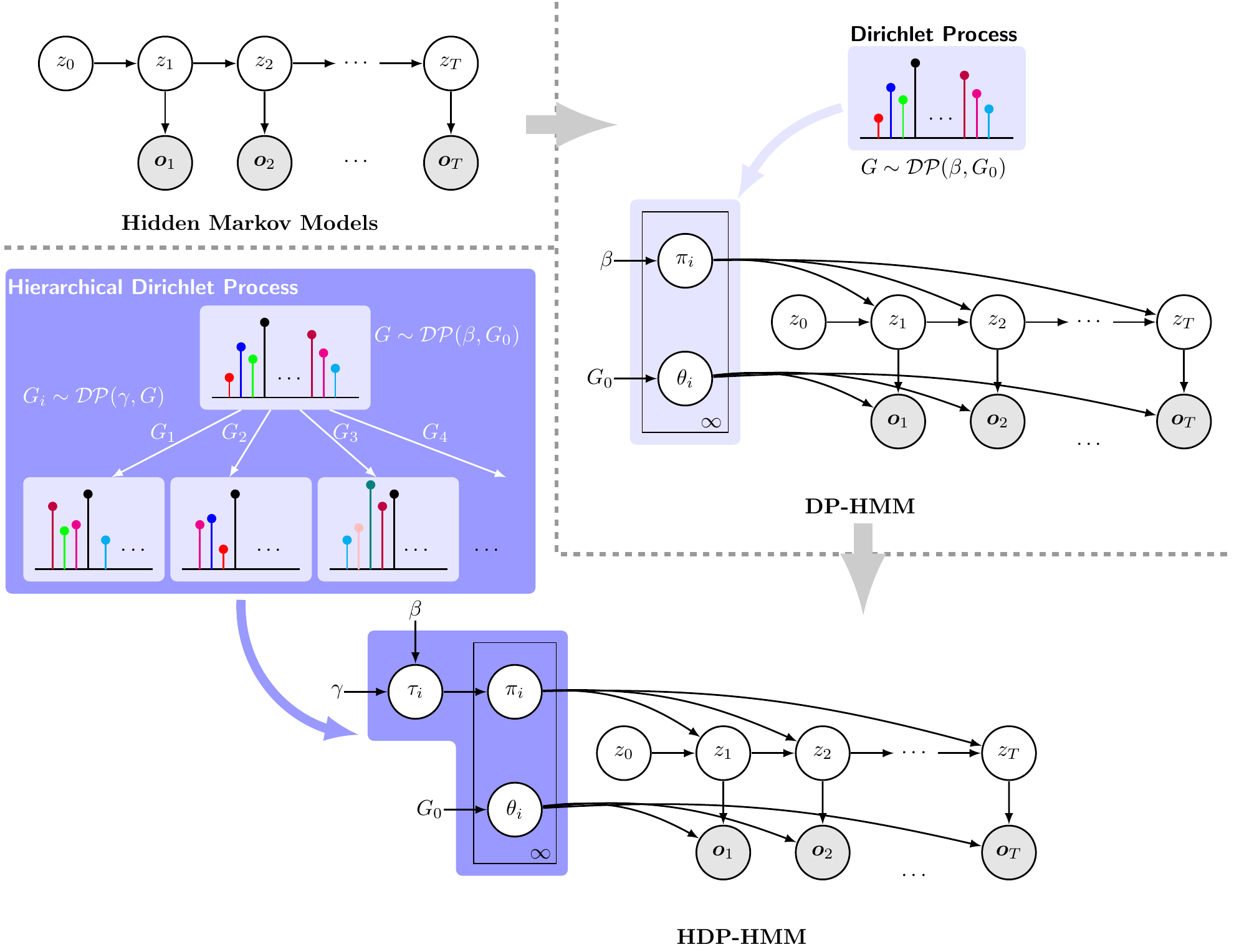}
	\caption{The derivation procedure from a regular hidden Markov model (HMM) to Bayesian nonparametric dynamic networks. Upper left: A typical graphical illustration of HMM with a finite number of discrete hidden states marked using unshaded circles, denoted as $z_{t}$, which are unobservable. Shaded circles represent observations, denoted as $\boldsymbol{o}_{t}$. Upper right: The HMM with infinite hidden states, which is derived by adding a Dirichlet process (DP) as a priori to its state transition model and emission model. The DP is built with a concentrated parameter $\beta$ and a base measure, $G_{0}$, denoted as $G\sim DP(\beta,G_{0})$. Lower: A hierarchical DP with hidden Markov frame. It is capable of sharing features among transition probabilities by assigning a hierarchical DP from which base measures $G_{j}$ are drawn from a discrete measure generated by a DP.}
	\label{fig:NPBL_method}
\end{figure}

Bayesian nonparametric dynamic networks can automatically learn the associated patterns from multi-variate sequential observations without requiring prior knowledge of the number of patterns. As a result, they have been applied to analyze individuals'  \citep{taniguchi2014unsupervised,taniguchi2015sequence, hamada2016modeling} and multi-vehicle interactive behaviors \citep{zhang2021spatiotemporal, zhang2019learning, wang2020understanding}. Usually, the underlying essential module of capturing the temporal evolution of sequential behaviors is based on a first-order Markovian process such as hidden Markov models and state-space models \citep{fox2009bayesian,fox2011bayesian}. The interactive process of driving behavior is assumed to obey the Markov properties and then captured using HMM, as shown by the top-right plot in Figure~\ref{fig:NPBL_method}. Each hidden state represents the label or type of driving patterns, modes, or maneuvers (e.g., yield and go, turning right/left). The conventional HMM requires predefining and fixing the number of hidden states \citep{wang2020uncovering}, which is counterfactual to the intuition that human drivers will increase their knowledge of the patterns (i.e., increasing the number of hidden states) when encountering new scenarios they have never seen. To enable the number of states to grow as more data get observed and share the knowledge of hidden states over different scenarios, researchers introduced a hierarchical Dirichlet process (HDP) as the prior distribution over these hidden states and their dynamics, forming an HDP-dynamic network.

With the benefits of HDP-HMM, researchers semantically analyzed the interaction process for complex driving behaviors in traffic using the information of connected vehicles. \cite{wang2020understanding} applied a sticky HDP-HMM to investigate the behavioral interaction primitives between two vehicles semantically with a compact representation. Then, \cite{zhang2019learning} also used the Bayesian nonparametric dynamic networks to learn the semantic interactive driving primitives at intersections, revealing the relationship between road intersection shape and behavior types. These two works provide insights into how two vehicles dynamically interact. Moreover, \cite{zhang2021spatiotemporal} leverages continuous (i.e., Gaussian processes) and discrete (i.e., Dirichlet processes) stochastic processes into the sticky HDP-HMM to reveal underlying interaction patterns of the ego vehicle with other nearby vehicles in lane-change scenarios on highways and the transition probability between these patterns.

Although the Bayesian nonparametric dynamic networks employed above enable the extraction of primitives constituting the sequential behaviors, it does not directly reveal the dynamic decision-making process of human drivers in terms of rewards and actions. This is because their basic dynamic models (e.g., HMMs, SSMs) do not consider the rewards and actions. 


\subsection{Topological Models}
\label{subsubsec:topology}

\paragraph{Why Topological Model?} In interactive scenarios, humans are general-purpose agents; the mechanisms of human action interpretation are teleological \citep{csibra1998teleological}. Moreover, people's decision-making succeeds in complex interactions by benefiting from their constructed mental representations \citep{ho2022people}, such as the topological abstractions \citep{konidaris2019necessity}, which effectively simplifies inference. As a result, humans abstracted their observations into high-level \textit{embeddings} invariantly shareable and reusable cross-interaction scenarios and behavior. So, \textit{how to define and find these embeddings?} Although the Bayesian nonparametric dynamic networks can analyze high-dimensional sequences by extracting primitives at a low level of granularity, they can not consider the multiagent interaction's behavioral and geometric structure and the teleological reasoning. For multiagent behaviors, two fundamental questions are naturally raised
\begin{itemize}
	\item Is there an approach that may encode any complex spatiotemporal, multiagent interactions into a compact representation algebraically and geometrically? 
	
	\item Is there a measure that may allow identifying any topology-preserving deformations of agents' trajectories?
\end{itemize}
Topological models provide a preliminary solution to these two questions, which can embed the structured information into an abstract formalism. Two topological models are frequently used: topological braids for the first question and topological invariance for the second one. 

\begin{figure}[t]
	\centering
	\includegraphics[width=0.7\linewidth]{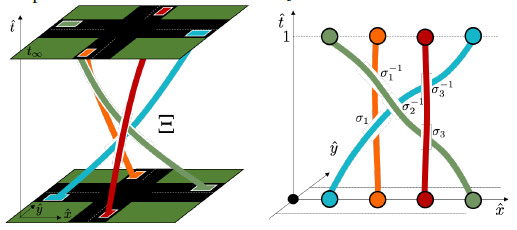}\\
	\vspace{1cm}
	\includegraphics[width=0.9\linewidth]{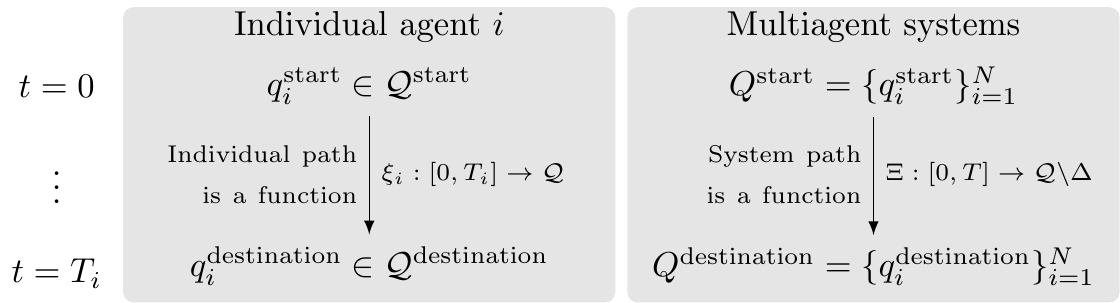}
	\caption{ Upper \citep{mavrogiannis2020implicit}: Representation of four-vehicle interactions using topological braids  with (left) corresponding trajectories of agents as they navigate an intersection over space and time, $x\times y \times t$, and (right) braid, $\sigma_{3}\sigma_{1}\sigma_{2}^{-1}\sigma_{3}^{-1}\sigma_{1}^{-1}$, capturing the topological entanglement of agents' trajectories. Lower: (left) The dynamics $i$-th agent's configuration ($q_{i}\in\mathcal{Q}\subset\mathbb{R}^{2}$) along time and (right) the dynamics of multiagent system configuration along time, where $\Delta$ represents the collision of agent paths.}
	\label{fig:topologicalBraids}
\end{figure}

\paragraph{Topological Braids}   One popularly used topological approach to modeling interactions in robotics is the topological braids (see Appendix~\ref{app:topologicalBraids}), which have succeeded in interpreting the interaction mechanisms of complicated dynamic phenomena in physics, such as fluids \citep{tan2019topological}. The topological braid may naturally encode an arbitrarily complex, spatiotemporal, and multiagent interaction in any environment into a compact representation of dual algebraic and geometric nature \citep{mavrogiannis2019multi}. For example,  \cite{mavrogiannis2020implicit,mavrogiannis2021analyzing} applied the formalism of topological braids to analyze the interactive behaviors between non-communicating rational agents at unsignalized intersections by collapsing the cooperative behaviors from a high-dimension space into a low-dimension space of modes. Figure~\ref{fig:topologicalBraids} illustrates an example of representing interaction trajectories of four vehicles at intersections using topological braids. Denote $\xi_{i}: [0, T_{i}]\rightarrow \mathcal{Q}$ the individual path for agent $i$, $i=1,2,\dots$, with corresponding start position ($q_{i}^{\mathrm{start}}\in \mathcal{Q}^{\mathrm{start}}$) and destination ($q_{i}^{\mathrm{denst}}\in \mathcal{Q}^{\mathrm{denst}}$). The system path $\Xi=\{\xi_{i}\}\in \mathcal{X}\times\mathcal{Y}\times\mathcal{T}$ was partitioned over $\mathcal{T} $ into a set of classes of homotopically equivalent system paths corresponding to a path of permutations. Each such class has distinct topological properties, forming a distinct \textit{joint strategy}\footnote{Conceptually, a joint strategy refers to a sequence of strategy profiles of all small interaction cycles in multiagent interaction systems. For more details, refer to Appendix~\ref{app:jointstrategy}.} that the agents follow to reach their destination without collisions. 

The topological braids have many advantages compared to interaction models using pure deep learning. The braid formalism enables abstracting interaction behaviors into topological modes which are identified symbolically in a compact and interpretable fashion. Besides, the topological approaches can also provide the complexity analysis of interactions, which Bayesian dynamic networks can not reach. Analog to the topological braids, other analytical methods have also been used to evaluate the priority preference of human drivers, which is then embedded into game-theoretic frameworks to model interactions. For instance, \cite{zanardi2021urban,zanardi2021posetal} formulated the interactions between human drivers as urban driving games and embedded each player’s problem complexity in the lexicographic preference or the prioritized metrics of the agent over the outcomes.

\begin{figure}[t]
	\centering
	\includegraphics[width=\linewidth]{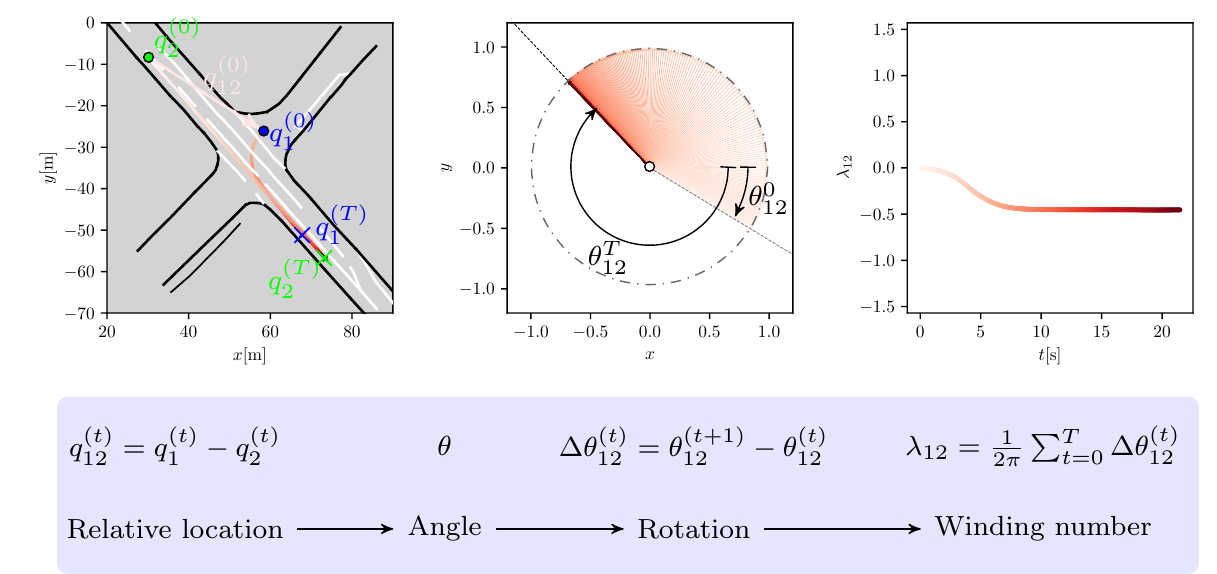}
	\caption{Upper: Identified Interaction modes with topological invariance at intersection crossing selected in the inD dataset \citep{bock2020ind}. (Left) Bird view of two vehicles' trajectories. (Middle) Unit winding vector trajectories (darkness increases with time). (Right) Winding number covargence. Lower: Computing procedure of the topological invariance.}
	\label{fig:toplogical-invariance}
\end{figure}

\paragraph{Topological Invariance} In interactive scenarios, the geometric structure of the driving environment and the incentive of human agents to negotiate efficiently without collision with each other compress the space of possible multiagent trajectories, effectively simplifying inference. The space compression can be represented by a finite set of \textit{modes} of joint behavior through topological invariance (see Appendix~\ref{app:topologicalinvariance}), enabling the trajectories belonging to the same mode to keep the same topological attributes. \cite{roh2020multimodal} designed a multiple typology to predict multi-vehicle trajectory based on the modes that are formalized using a notion of topological invariance. Figure~\ref{fig:toplogical-invariance} illustrates their idea, where the \textit{winding number}, $\lambda_{12}$, is a topological invariant\footnote{A topological invariant refers to a function which remains unchanged when a specified transformation is applied}. Given a fixed pair of agent's endpoints, any topology-preserving deformations of their trajectories would be identified using the same winding-number value. Theoretically, the definition of topological invariance (see Appendix~\ref{app:topologicalinvariance}) provides an intuitive explanation: Any trajectories of one vehicle that encircle the other vehicle the same number of times maintain the same winding number value. A specific winding-number value can represent many topologically-close interaction scenarios. Moreover, the sign of the winding number corresponds to the rotation direction (i.e., clockwise or counterclockwise) of the winding vector $q_{12}$, indicating the side on which the two agents pass each other. Specifically, we have 

\begin{equation}
	w_{12} = \mathrm{sign}(\lambda_{12})=
	\begin{cases}
		\mathrm{Right-side \ passing}, & \mathrm{if \ positive} \\
		\mathrm{Left-side \ passing}, & \mathrm{if \ negative}
	\end{cases}
\end{equation}
This topological invariance of two agents can be extended into a multiagent interaction scenario. 

\subsection{Summary}
The learned spatiotemporal interaction between human drivers can infer the vehicle relationships via graph models with nodes as vectorized spatiotemporal features \citep{choi2019drogon}. Although GNNs and social pooling with tensor fusion are convenient to use auxiliary information to train a network in an end-to-end manner, they cannot ensure whether the desired physical or social factors are actually captured and learned (i.e., lack of interpretability). Bayesian networks provide an interpretable way to build models flexibly, allowing for an explicit description of dependencies among human agents and their sequential interactive behaviors. Topological models could provide an analytical way to consider multiagent interaction's behavioral and geometric structure across different driving tasks. 

Most works on multi-agents behavior and trajectory prediction using graphical models are toward the safe path-planning and control design of socially-compatible autonomous vehicles by carefully using these predictions. However, the fidelity of the interaction models derived in the absence of autonomous agents is questionable when applying them in human-autonomy mixed settings since human drivers might act differently on autonomous cars. These models neglect the influence of autonomous systems on human agents. Furthermore, GNNs provide an intuitive understanding but not quantitatively-precise insights, which is insufficient for safety-critical applications such as autonomous driving. A further discussion is conducted in Section~\ref{sec:discussion}.

\begin{figure}[t]
	\centering
	\includegraphics[width=0.75\linewidth]{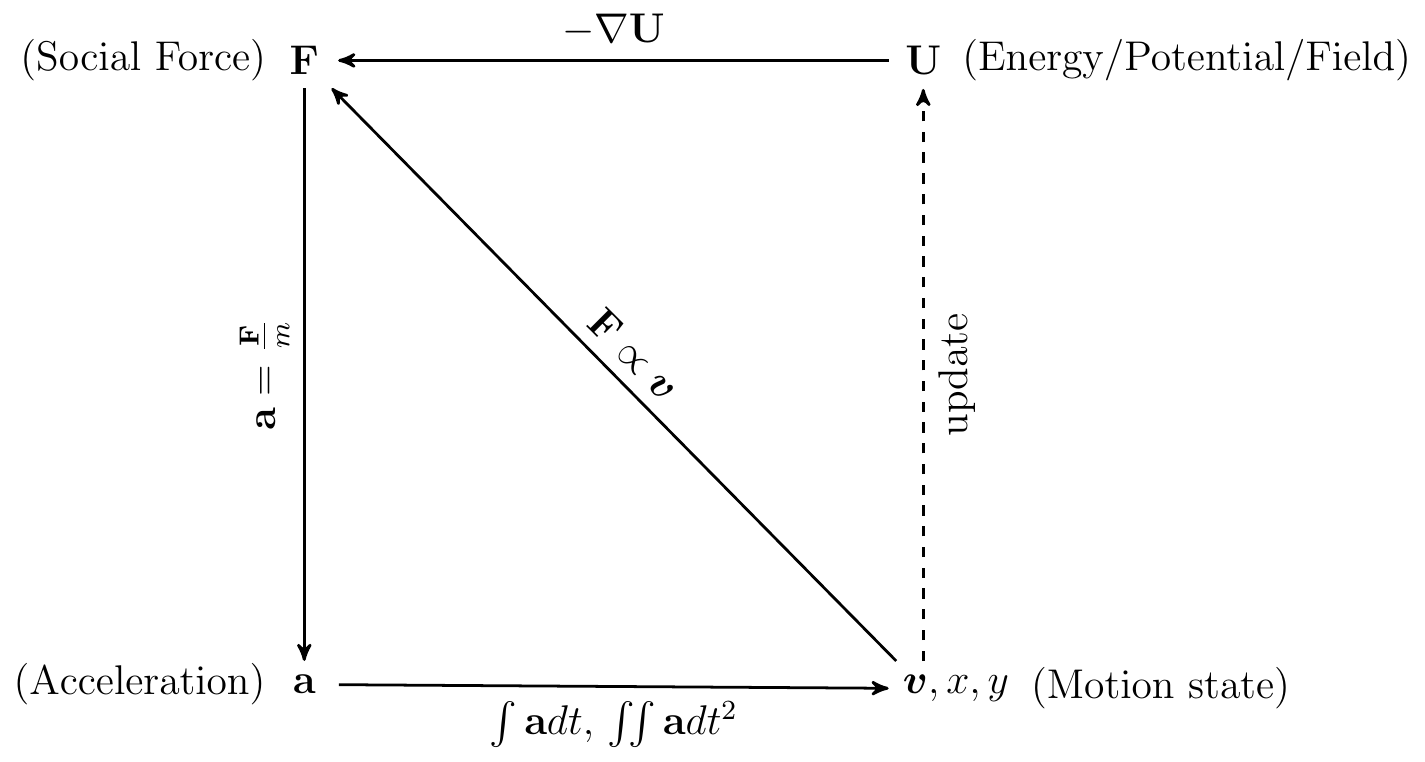}
	\caption{The diagram of the relationship between social forces, social potentials/fields, and acceleration for agent interactions.}
	\label{fig:social_force}
\end{figure}

\section{Social Fields \& Social Forces}
\label{subsec:social_force_based_model}
While modeling human driving interactions given the complicated (e.g., environmental, internal, and social) stimulus of motion, it is essential to consider the virtual forces actively created in the human mind when perceiving the traffic environment like road boundaries and sidewalks. On the other hand, it is equally critical to consider the influence of other human drivers on social rules to drive. For instance, when a human driver approaching the intersection and taking an unprotected left turn sees an aggressive forthcoming vehicle, the driver will change their decision before colliding with the vehicle.  In this process, the upcoming vehicle does not apply a contacted force to the ego vehicle, but the human driver reacts as if a force exists. This type of virtual force that does not exist but can intuitively describe and explain the social interactions is the social force model \citep{helbing1995social}. Besides, forces are the governing and unifying factor of all interactions and motions \citep{su2019potential}, and social force-based reward design can improve interaction performance for multi-robot navigation \citep{gil2019effects}. Therefore, it is reasonable to assume that human driver-related movements (e.g., move forward and steering) are governed and driven by a virtual force generated from the internal mind of humans (e.g., internal motivation, desired speed and destination) and external constraints (e.g., traffic regulation, obstacles, and moving agents). The detailed social force models will be discussed in Section~\ref{sec:social_force}.

\begin{figure}[t]
	\centering
	\includegraphics[width=0.6\linewidth]{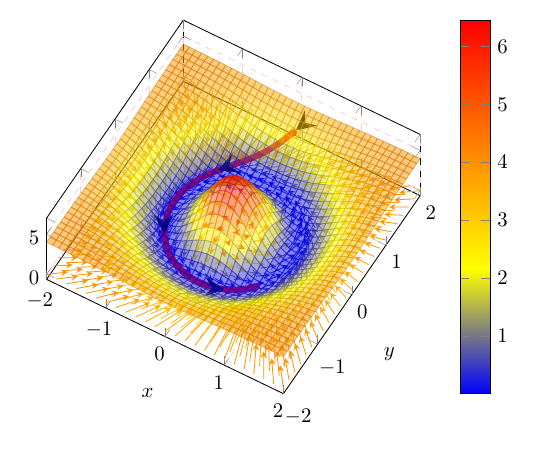}
	\caption{Illustration of fields and trajectories compelled by forces.}
	\label{fig:fields}
\end{figure}

From the other view of fields and potentials, the underlying reason a human driver follows a lane and interacts with other vehicles is that the driver restricts themselves within a field formed by the influences of traffic environments and other agents. Human drivers generate trajectories and take actions by balancing costs and rewards, which leads to a total utility characterized by a potential field. \cite{gibson1938theoretical} pioneered a theoretical field analysis of automobile driving from psychology and termed `\textbf{Fields of Safe Travel}'. The virtual forces compel human agents to move towards the lower field values, analogous to a charged particle in an electrical field in physics. When discussing social forces, the corresponding integrals known as potentials, fields, and energy cannot get around, as shown in Figure~\ref{fig:social_force} and Figure~\ref{fig:fields}. The field-related models will be discussed in Section~\ref{sec:fields}. 

On the other hand, human-driven vehicles' motion states (i.e., speed and position) are derived from acceleration over time, as shown in Figure~\ref{fig:social_force}. Therefore, directly learning the acceleration can capture the same interactive effects of social force.

\begin{table}\centering
	\caption{Examples of the Commonly-Used Kernel Functions for the Sub-Fields based on Relative Distance ($d\geq 0$) and Velocity ($v$) for Two Agents}
	\label{tab:kernel_functions}
	{\small
	\begin{tabular}{lll}
		\hline
		\hline
		Type of Sub-fields & Formulation & Parameters \\
		\hline
		Moving cars & $ \mathbf{U}_{\mathrm{moving}} = \frac{\exp(-\alpha_{1} d)}{1 + \exp(-\alpha_{2} v d)} $ & $\alpha_{1}, \alpha_{2}$ \\
		& & \\
		Static obstacles & $\mathbf{U}_{\mathrm{static}} = A_{\mathrm{car}} \frac{\exp(-\alpha d)}{d} $ & $A_{\mathrm{car}}, \alpha$  \\
		& & \\
		Road edges & $\mathbf{U}_{\mathrm{edge}} = \beta \left(\frac{1}{y - y_{0,i}}\right)$ & $\beta$  \\
		& & \\
		Road lines & $\mathbf{U}_{\mathrm{lane}} = A_{\mathrm{lane}}\exp\left( -\frac{(y - y_{i})^{2}}{2\sigma^{2}} \right)$ & $A_{\mathrm{lane}}, \sigma^{2} $ \\
		& & \\
		Dynamic traffic control & $\mathbf{U}_{\mathrm{vel}} = \gamma (\boldsymbol{v} - \boldsymbol{v}_{\mathrm{des}})$ & $\gamma$ \\
		Social driving factors & &  \\
		\hline
		\hline
		\end{tabular}
		}
\end{table}

\subsection{Theory of Social Fields}
\label{sec:fields}
In physics, the distribution of a physical quantity (e.g., velocity, temperature, electric, and magnetic) in a specific region of space is referred to as a \textit{field}. Analogously, the distribution of a traffic-related quantity such as risk or safety levels in an area of interest has termed a risk or safety field, known as the fields of safe travel \citep{gibson1938theoretical}. Field theories provide a unified framework to describe environmental constraints and collision risks from which (sub)optimal trajectory and motion planning can be achieved for mobile robots. Human drivers interact with surroundings via a perception-action loop: assessing the risk level of environments in real-time, predicting the motions of other agents, and then generating reactions to the dynamic environments.
The field theory that reveals the underlying mechanisms of interactions by longitudinal and lateral vehicle operations while driving on the road is a special field \citep{gibson1938theoretical}. The relevant constructed fields have also been used to represent interactions between human drivers and traffic environments \citep{kruger2020interaction}. In multi-vehicle interactive traffic scenarios, existing works adopted many different terminologies relevant to the field theory, including risk fields/maps or driver's risk field \citep{kolekar2021risk, tan2021risk, liu2022interactive}, driving risk potential fields \citep{li2020dynamic, akagi2015stochastic}, artificial potential fields \citep{wolf2008artificial, kim2017local, kim2018trajectory, yi2020using, kruger2020interaction}, occupancy risk cost \citep{pierson2018navigating}, or driving safety field \citep{wang2015driving}. Although the used terminologies are diverse, they share the common points: \textbf{artificially designed} and \textbf{risk information-based}. In what follows, we classify the aforementioned social fields into three groups: risk/safety fields, potential fields, and occupied fields. 

\paragraph{Risk/Safety Fields} The fields of safe travel are characterized by a scalar quantity of \textit{risk values}\footnote{Traffic scenarios can be evaluated by the \textit{risk} level or, conversely, the \textit{safety} level. A field quantified with risk levels is called a risk field, and a field quantified with safety levels is called a safety field: The less risk, the safer. Therefore, here we reviewed and discussed the safety/risk field together.} over a predefined space, forming a risk field. Many factors can influence the risk level. Traffic psychologists \citep{gibson1938theoretical} hold that the field of safe travel is composed of three sub-fields: the field of the human driver, the field of other human drivers, the field of the car itself. However, they failed to consider the influence of fixed traffic structure (e.g., road edges and lane lines) and traffic control (e.g., stop signs and traffic lights). The risk level is a superposition of several disparate and independent functions related to the following aspects:

\begin{itemize}
	\item Moving objects such as other vehicles. A symmetric circle-based or Ellipse-based field can evaluate the influences of a moving car on its surrounding area \citep{pierson2018navigating, wolf2008artificial, tan2021risk}. However, moving cars will have different impacts on their front and back, left and right directions. A forward-moving car with positive acceleration would have stronger influences on its front than its back areas, and a left lane-changing car has more potent influences on its left than its right. Other moving attributes (i.e., velocity and moving direction) can be embedded into the field functions. \cite{wang2015driving} applied a symmetric circle-based field considering speed to form the field of moving cars, which was then extended to an elliptic-based field by \cite{li2019shared}. Also, \cite{pierson2018navigating} developed an asymmetric risk field with a Gaussian peak multiplied by a logistic function. 
	
	\item Static obstacles such as street-parking cars. The influence of static obstacles on the surrounding region depends on their shape, mass, and other attributes \citep{wolf2008artificial, tan2021risk}. A parking truck or a supercar (e.g., Aston Martin) would have a more decisive social influence on other surrounding human drivers than a regular car, compelling surrounding human drivers to slow down and pass away with ample space. Such interactions come from humans' social cognition, which still leaves challenges in modeling.
	
	\item Road constraints, including road edges and lane lines. These physical factors should be considered when modeling interaction behaviors \citep{wolf2008artificial, tan2021risk, kruger2020interaction} since social interactions in traffic strongly depend on the environment structure.
	
	\item Dynamic traffic controls such as  traffic lights. Traffic lights are used to reduce, even remove, the uncertainties of human agents' interactions, forming central coordination \citep{tan2021risk}. Human drivers can achieve challenging driving tasks by directly following traffic light guidance. 
	
	\item Social driving factors. Human drivers have different social driving factors, ranging from the desired velocity and acceleration  \citep{wolf2008artificial} to hazard anticipation for the occluded region (e.g., a blind corner) \citep{akagi2015stochastic}. These social driving factors diversify the human driver's subjective estimation of danger resulting from perceptions and anticipations, generating different decisions and control policies.
\end{itemize}

Some of the above factors can be summarized, simplifying the risk/safety field formulations. \cite{wang2015driving} developed a driving safety field consisting of three sub-fields: a potential field for static obstacles and road constraints, a kinetic field for moving objects, and a behavior field for social driving factors, without considering the dynamic traffic control information. Moreover, many existing works tried to define a simplified risk field by only considering the static obstacles (including road constraints) and moving vehicles \citep{li2019shared, li2020dynamic, li2020risk}, such as a well-defined probabilistic driving risk field considering the probability of other surrounding vehicles' motion prediction \citep{mullakkal2020probabilistic}. On the other hand, the mixture of different types of fields can benefit interaction behavior modeling. \cite{mullakkal2020probabilistic} formulated road boundary objects as a potential field and surrounding vehicles as a kinetic risk field to depict how the driver behaves in uncertain environments.

Researchers first qualitatively analyze human drivers' typical perception-reaction in interactions and then design each sub-field function with associated constraints. Table~\ref{tab:kernel_functions} lists several commonly-used kernel functions for each sub-field formulation, most of which are Gaussian-based due since they are explainable and differentiable \citep{pierson2018navigating, kim2017local, kim2018trajectory}. All the designed sub-fields are dynamic and vary over time and space except the static obstacles and road constraints. \cite{tan2021risk} developed an artificial continuous function with value constraints $[0,1]$ by adding constant terms and the reciprocal operation of functions. Inspired by the Navier–Stokes equations in fluid dynamics, \cite{zhu2018field} and \cite{li2020fluid} developed a computational fluid-inspired risk field by assuming each vehicle on the road emits invisible smoke. The density of smoke responds to vehicle motion, providing a dynamic update of risk fields, and a higher density of smoke indicates a closer distance to the nearby vehicle. There are also other alternatives for formulating the interactions. For example, one straightforward idea is to use pure data-driven approaches such as neural networks \citep{hossain2022sfmgnet,kreiss2021deep}, but they usually lack explainability.

\paragraph{Potential Fields} In physics,  explicit and intuitive relations exist between potential and fields, e.g., electric fields and electric potentials --- the fields are the minus of the differential of the potential with respect to distance. Unlike the physical potentials having rigorous, objective proof and validation, the safe driving potential functions are heuristically and manually designed according to experts' insights and understanding of human driving behavior. \cite{khatib1985real} artificially created the artificial potential fields\footnote{In physics, \textit{potentials} and \textit{fields} are two different basic concepts and have explicit formulations. In traffic, researchers usually mix them together and refer them to as \textit{potential fields}.} for real-time obstacle avoidance for manipulators and mobile robots, which then were widely extended to multi-vehicle interactions toward collision-free path planning and control \citep{wolf2008artificial, bahram2016combined, semsar2016cooperative, woo2016dynamic} and human driving behavior modeling in dynamic traffics \citep{gonzalez2018modeling}. The potential field can be dynamic and evolving, forming an adaptive potential field by leveraging the future information of surrounding vehicles \citep{kim2017local, kim2018trajectory, tan2021risk} or the obstacle vehicles' speed and acceleration information under road coordinate systems \citep{lu2020adaptive, li2020novel}.

Like the risk fields, the potential fields are aggregation or weighted summation of several independent sub-potential fields. However, researchers have different views of the definition of potential fields. Some researchers believe that moving and non-moving objects can influence the potential field. \cite{su2019potential} built a potential field with three sub-potential fields (i.e., potential environmental field, potential inertial field, social force field) to capture the effects of the stimuli from environmental, internal, and social factors. \cite{li2020dynamic} developed a dynamic driving risk field by weighting three sub-potential fields: lane marks, road boundaries, and moving/non-moving vehicles. Generally, for multiagent scenarios, denote $\boldsymbol{\xi}_{i}=(x_{i},y_{i})$ the position of each human agent. The occupancy risk cost $\mathbf{U}$ can capture the density of agents at any query location $\boldsymbol{\xi}^{\ast}$ \citep{pierson2018navigating,wang2020learning}, which can be evaluated by considering the influences of moving direction, i.e., the direction of velocity

\begin{equation}
	\mathbf{U}(\boldsymbol{\xi}^{\ast}) = \sum_{i=1}^{n}\frac{\exp\left(-(\boldsymbol{\xi}^{\ast} - \boldsymbol{\xi}_{i})^{\top}\Omega(\boldsymbol{\xi}^{\ast} - \boldsymbol{\xi}_{i})\right)}{1+\exp \left( - \alpha \dot{\boldsymbol{\xi}}_{i}^{\top} (\boldsymbol{\xi}^{\ast} - \boldsymbol{\xi}_{i}) \right)},
\end{equation}
where $\alpha$ is the parameter of skewing the position by the velocity, and $\Omega$ is the diagonal matrix of the inverse square of the standard deviation. In the $x-y$ coordinates, $\Omega=\mathrm{diag}\{1/\sigma^{2}_{x},1/\sigma^{2}_{y} \}\in\mathbb{R}^{2}$. 

On the other hand, some researchers hold that the potential fields only stem from non-moving objects. For instance, \cite{li2019shared} treated the potential field as a field formed only by \textit{static} objects such as obstacles, road lanes/edges, and traffic signs. They then developed a potential field with two sub-potential fields: stationary obstacles, as formulated in \cite{wang2015driving}.

\paragraph{Parameter Estimation of Social Fields} The main difficulty inherent to the field theoretic-based approach is parameter estimation. Most researchers \citep{wolf2008artificial, wang2015driving, li2020dynamic} determined the field parameters according to their subjectively qualitative judgment and evaluation of human driver interactions. Choosing the optimal hyper-parameters is tedious, which is an $\mathcal{NP}$-hard problem. Although estimating field parameters seems intractable, some promising progress has been made by carefully designing a computationally tractable field for specific driving behaviors. \cite{akagi2015stochastic} developed a simplified potential field over risk levels to model braking behavior at unsignalized urban interactions, which allows formulating the parameter estimation as an optimization problem. However, this achievement is at the cost of the generality of models.

In general, the interaction among human drivers is complicated, and the mixture of potential and risk fields could provide an efficient solution to behavior modeling for the specific implementation. To mimic human driving behavior, \cite{guo2017humanlike} developed a hybrid potential map with trajectory-induction potentials and risk-potential fields for each category of cars recognized via Bayesian networks.

\subsection{Social Forces}
\label{sec:social_force}
\cite{helbing1995social} applied the concept of social forces to model pedestrian interaction and showed promising performance. Then, its derivations have been proposed for modeling the influences between pedestrians \citep{pellegrini2009you,chen2018social} and interactions between heterogeneous road users, such as vehicle-crowd interactions \citep{anvari2015modelling, pascucci2015modeling, rinke2017multi, ma2017two,yang2021sub}. In this paper, we will not discuss the interaction of human drivers with other types of road users (i.e., pedestrians, cyclists, and motorcyclists); instead, we are only concerned with the interactions between human-driven vehicles on roads. 

\begin{figure}[t]
	\centering
	\includegraphics[width = 0.5\linewidth]{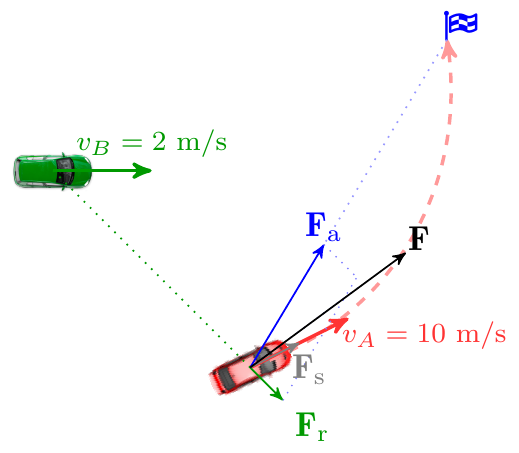}
	\caption{Illustration of two-vehicle social interaction with the social force model, composed of three forces: self-driven ($\mathbf{F}_{\mathrm{s}}$), repulsive ($\mathbf{F}_{\mathrm{r}}$), and attractive ($\mathbf{F}_{\mathrm{a}}$). The synthesized social force ($\mathbf{F}$) finally acts on the red car and changes its motion states.} 
	\label{fig:social force example}
\end{figure}

In natural traffic scenes, a complex sensory stimulus (i.e., internal, environmental, and social) causes a behavioral reaction that depends on the human driver's aims and is chosen from a set of behavioral alternatives to maximize utility. A rational and experienced human driver usually adapts to the situations they are familiar with according to their habits in the brain \citep{dolan2013goals}; thus, the driver's reaction is relatively automatic and determined by their experience of which reaction will be the best. In such a view, researchers mathematically recover the behavioral rules and the effects between human agents by a vectorized quantity, called social force. The social forces generated between human agents and the physical limitations are the sources leading to behavior changes, which generally are composed of three types of forces corresponding to internal, environmental, and social stimuli:

\begin{itemize}
	\item \textbf{Self-driven force.} The self-driven force, $\mathbf{F}_{\mathrm{s}}$, is an intrinsic force of human drivers and the reflection of internal stimulus, which pushes the driver to reach their destination (e.g., desired velocity) as comfortably as possible by changing motion. The self-driven force can be linearly proportional to the difference between the current velocity $\boldsymbol{v}_{t}$ and the desired velocity $\boldsymbol{v}^{\mathrm{desired}}$, formulated by
	
	\begin{equation}
		\mathbf{F}_{\mathrm{s}}(t) = \alpha (\boldsymbol{v}^{\mathrm{desired}} - \boldsymbol{v}_{t}).
	\end{equation}
	where $\alpha$ is the coefficient. Some researchers interpret this underlying self-compelling force as the gravity force in traffic flow \citep{ni2011unified}. Note that the desired velocity could be constant (e.g., driving on highways and keeping speed close to a speed limit of $70$ km/h) or time-varying (e.g., approaching an intersection by slowing down).  
	
	\item \textbf{Repulsive force.} Neural evidence reveals that human brains shape peripersonal and extrapersonal spaces \footnote{The \textit{peripersonal space} is the space that directly surrounds us and with which we can directly interact, and the \textit{extrapersonal
	space} is the space far away from the subject. The body cannot  directly act upon that.} when socially interacting with other agents \citep{clery2015neuronal}. Each human driver keeps a safe and comfortable distance from static obstacles and other drivers depending on the speed, acceleration, and hazard perception level. A human driver usually feels increasingly uncomfortable when getting closer to the other moving vehicle at a higher speed. This results in \textit{repulsive effects} on the driver. This repulsive force, $\mathbf{F}_{\mathrm{r}}$, stems from the influence of other human drivers and the reaction to environmental stimuli to avoid collisions. Besides, the repulsive effects can also come from other natural physical boundaries (e.g., road edges, static obstacles, trees, and houses.) and virtual information constraints (e.g., stop-sign, road markers).
	
	\item \textbf{Attractive force.} Human drivers are sometimes attracted by other human drivers, objects, and targets (e.g., destination) socially, for example, to get the right of way and avoid the cut-in behavior of other vehicles by closely following a leading car (see examples in the \href{https://www.youtube.com/watch?v=UEIn8GJIg0E}{traffic video}). These attractive effects are termed as attractive forces, $\mathbf{F}_{\mathrm{a}}$, guiding the vehicle to follow the other agent.
\end{itemize}
The synthesized social force is computed by 

\begin{equation}\label{eq:social_force}
	\mathbf{F} = \sum_{\mathcal{F}_{\mathrm{s}}} \mathbf{F}_{\mathrm{s}} + \sum_{\mathcal{F}_{\mathrm{r}}} \mathbf{F}_{\mathrm{r}} + \sum_{\mathcal{F}_{\mathrm{a}}} \mathbf{F}_{\mathrm{a}},
\end{equation}
with $\mathbf{F}_{s}\in\mathcal{F}_{\mathrm{s}}$, $\mathbf{F}_{r}\in\mathcal{F}_{\mathrm{r}}$, $\mathbf{F}_{a}\in\mathcal{F}_{\mathrm{a}}$, where $ \mathcal{F}_{\mathrm{s}}$, $\mathcal{F}_{\mathrm{r}}$ and $\mathcal{F}_{\mathrm{a}}$ are the sets of self-driven forces, repulsive forces, and attractive forces, respectively. Figure~\ref{fig:social force example} illustrates an example of a two-vehicle interaction scenario. The red car moves at a slightly high speed (e.g., $v_{A}=10$ m/s) intends to pass in front of the green car and approaches its destination. Without considering the influence of road structure, the social interaction between these two cars can be formulated via a social force model with a combination of the three types of forces. 

\begin{figure}[h!]
	\centering
	\includegraphics[width = 0.8\linewidth]{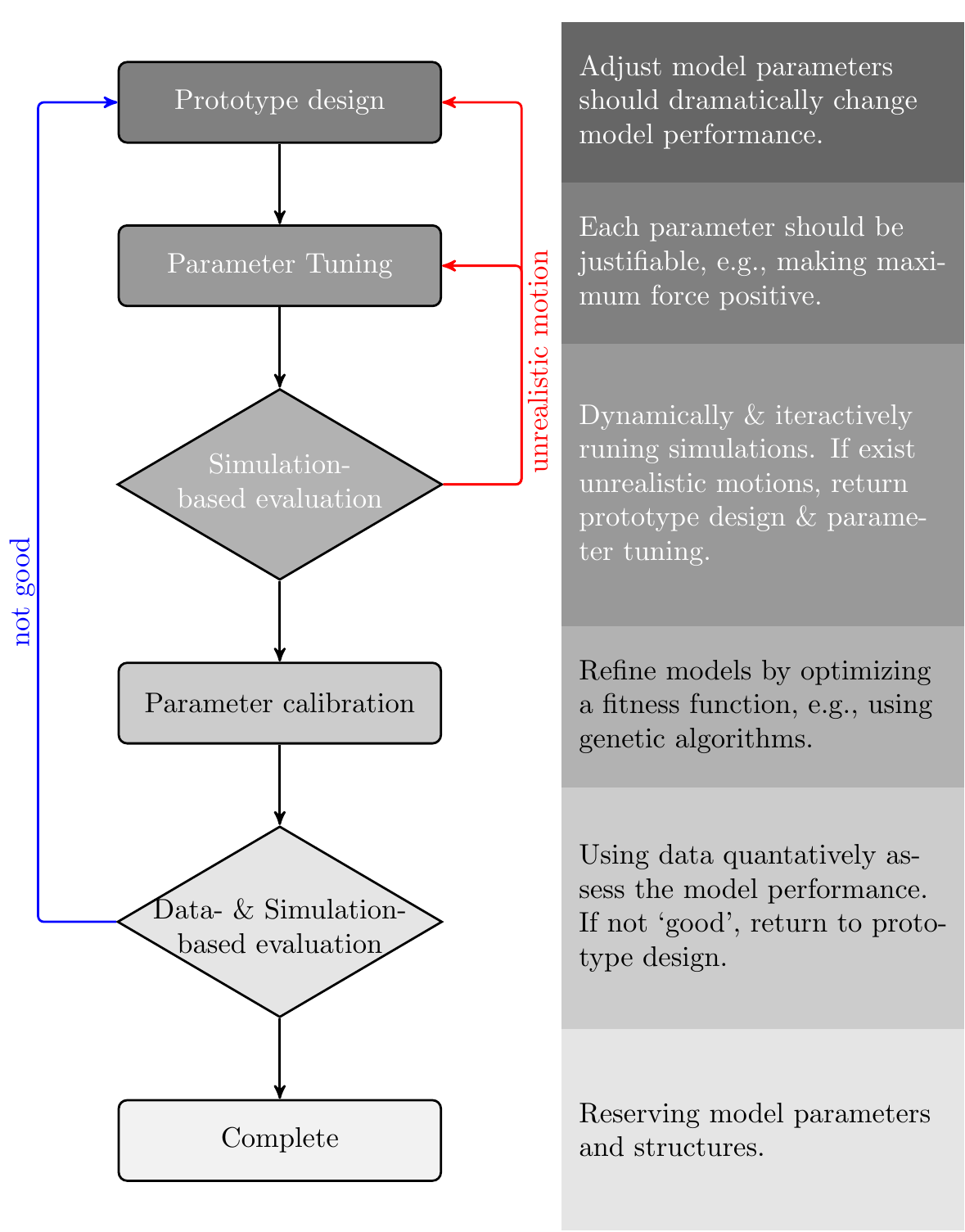}
	\caption{Process of designing and evaluating a social force model.}
	\label{fig:social force design}
\end{figure}

\paragraph{Social Forces Design}
As mentioned above, the three classes of forces are derived from pedestrians or crowds, which basically covers all the fundamental social forces of interactive driving behaviors in traffic. However, it is necessary to modify these forces to make them suitable for specific applications due to the difference between pedestrians and vehicles. For example, the constraints on accelerations and angular velocities of the vehicles must be tighter than those of pedestrians due to the physical limitations. \cite{yang2018model} designed five sub-social forces (i.e., self-driven forces, following forces, repulsive forces, boundary forces, and passable-gap forces) and combined them to describe the dynamic interactive behaviors between vehicles at intersections with an island work zone. Besides, social forces allow leveraging domain knowledge of environment physical constraints and interaction behaviors into dynamic models of individuals. \cite{ding2016new} inserted the combination of repulsive and attractive forces between agents into the general dynamic model of a target tracking problem but only considered the longitudinal influence. In contrast, \cite{tian2020multi} considered both longitudinal and lateral influences with social forces. Although a straightforward design of complicated social force models sounds reasonable, it usually does not work as expected. A \textit{generic} modeling process is a recycled and iterative procedure \citep{yang2021sub} consisting of prototype design, parameter tuning, simulation-based evaluation, parameter calibration, and data- \& simulation-based evaluation, as shown in Figure~\ref{fig:social force design}. 

\paragraph{Model Calibration and Validation} Calibrating a good social force model is not trivial because (i) most model parameters do not have a specific immediate interpretation to be measurable directly --- most of them are abstract meaning, (ii)   one single parameter often impacts many aspects of driving behavior, although the social force model is assumed to be a combination of several independent sub-force models, and (iii) a specific aspect of driving behavior results from the values of more than one parameter. Most existing social force models calibrate model parameters heuristically and subjectively without specific explanation \citep{ding2016new, tian2020multi}. There is no unique standard way to tune these parameters or take model calibration using data-driven techniques. Fortunately, some methods exist to calibrate the social force models of pedestrians and vehicle-pedestrians \citep{kretz2018some}. 

When using data-driven techniques to do calibration, researchers usually classify the model parameters into two types: \textit{measurable} and \textit{non-measurable}. The measurable parameters can be directly obtained via sensors or calibrated via statistical data analysis and vehicle's physical limitations, such as the maximum steering angle and acceleration. While for the non-measurable parameters, the curve fitting algorithms are needed, such as binomial
logistic regression \citep{xu2019simulation} and nonlinear programming with genetic algorithms \citep{yang2018model, yang2021sub}.

Inspired by the relationship between acceleration and forces, directly learning acceleration is an alternative to result in social forces instead of estimating the affected forces between agents via field theory, as shown in Figure~\ref{fig:social_force}.  \cite{ju2020interaction} develops an encoder-decoder to extract the interaction-aware acceleration from the past traffic environment observations. The derived acceleration allows straightforward integration with vehicle dynamics models with explicit physical meanings and low-level controllers for autonomous vehicles. 

\subsection{Summary}

Borrowing the concept of fields from physics (e.g., the electric field or the magnetic field in the theory of electricity) to create a virtual field of driving itself is a rather peculiar sort of field leveraged with social factors in several aspects. The social fields and social forces provide a conceptually-unified framework explaining driver behavior in different scenarios \citep{kolekar2021driver}, thus can be used for risk evaluation \citep{kolekar2021risk, li2020risk}, optimal control, behavior prediction, and path planning. Moreover, they are flexible in leveraging other traffic factors. Field theory-based models are usually built through quantitative analysis according to researchers' subjective cognition and understanding of human driving behavior. Thus, many factors can be leveraged, such as road attributes (e.g., straight and curve roads), traffic conditions, vehicle attributes (e.g., shape and mass), and human factors (e.g., the driver's attention). Theoretically, the field-based models can tackle complicated scenarios but only have been validated in simple interaction scenarios, e.g., car-following and lane change interactions between two agents \citep{wang2015driving, li2020dynamic, li2020novel}, due to the overwhelming calibration efforts. 

The elements generating the fields are more selective. Not all but partial elements and agents in the environments are salient for task performance \citep{niv2015reinforcement}. This is because the pertinent elements that dominate motion and driving tasks are attended to while non-pertinent elements normally recede into the background \citep{langdon2019uncovering, niv2019learning, radulescu2021human}. Besides, the influence between two human drivers could be asymmetric to each other according to their roles in interactions. For instance, the surrounding cars could be a leader car, parked car, tail-end car, merging car, and others, and their potential influences on the ego car are different. Each category of cars may share an identical potential field, making a more compact potential field model. \cite{guo2017humanlike} considered the similarity and differences in surrounding agents by categorizing them into groups using Bayesian networks.

The field of safe travel is, in nature, objective or/and subjective representation. Some researchers \citep{wang2015driving} treated the driving risk field as a physical field and claimed that it is an objective feature that does not vary with a person's subjective willingness. This might be true from `the view of the god' --- analyzing the interaction behaviors from a bird's eye view. However, this conclusion might be wrong if analyzing the interaction behavior from the first-person view of the ego car with considerations of social factors. We argue that the field is a risk field actively and passively perceived by human drivers, which reflects the human driver's actions and reactions in social interactions, and therefore is in nature a subjective representation. 

\section{Computational Cognitive Models}
\label{subsec:cognition}
One of the powerful attributes enabling human drivers to interact with other rational agents safely and efficiently through implicit communications is the cognitive mechanisms.  All approaches to capturing human drivers' interactions have the associated behavioral foundations supported by behavioral and psychological behavioral cognition. Therefore, cognitively understanding the interaction is the prerequisite to designing efficient interaction modeling approaches. Addressing fundamental questions at the frontiers of cognitive science would be one solution to socially-compatible autonomous driving on social roads \citep{chater2018negotiating}.

Most existing works about the cognitive models mainly focused on an individual's driving behavior rather than on the interaction between agents. They are particularly appropriate for modeling a single agent's relationship between higher-level and lower-level behavior. The earliest work of cognitively modeling human driving behavior can be dated back to the 1980s \citep{wilde1972general, wilde1976social} but focuses on individual target drivers' path-tracking behavior, steering ability, and visual-action responses \citep{donges1978two, meyer1997computational}, without revealing the mutual influences among human agents. Later, \cite{salvucci2001toward, salvucci2006modeling} proposed a specific type of cognitive architecture --- the production-system architecture based on condition-action rules (e.g., Adaptive Control of Thought-Rational, called ACT-R \citep{anderson2014atomic} and see a review paper \cite{ritter2019act}), describing how the driver processes information and generates behavior through control, monitoring, and decision-making component processes. Although the ACT-R architecture has been then extended to model driver behavior \citep{liu2006multitasking, deng2019modeling, zhou2020driving}, providing a unifying account of the developmental data, its insights are restricted to qualitative phenomena. It does not explain quantitative variation in human reasoning as a function of actions, situations, or prior expectations of plausible mental states. As a result, there leaves a gap between the conceptually unified \textit{architecture}\footnote{A cognitive architecture specifies the basic building blocks for cognitive models, and how those components can be assembled \citep{danks2014unifying}. Cognitive architectures describe aspects of our cognitive structures that are relatively constant over time. Therefore, a single cognitive architecture encodes a set of cognitive models.} and the  applicable real-time \textit{models}\footnote{A cognitive model offers a computationally well-specified account of a specific aspect of cognition, which usually provides a specific function with well-specified inputs and outputs. In practice, almost all cognitive accounts that make concrete predictions are cognitive models.} for socially-compatible autonomous driving in interactive scenarios. 

Going beyond this limitation requires developing techniques using cognitively inspired computational modeling. In interaction processes, the driver needs to actively and passively perceive its surrounding vehicles, react to other vehicles' current behavior as well as the future potential behaviors anticipated by the ego agent according to their prior knowledge, and finally make an efficient decision based on the beliefs of their judgment of the environment. Inspired by the interaction procedure, we try to answer some fundamental questions from the perspective of the ego driver:
\begin{enumerate}
	\item How could the driver \textbf{react} to others' behaviors?
	\item How could the driver be able to \textbf{understand} others' behaviors?
	\item How does the driver \textbf{perceive}, i.e., process the perceptual information, while arriving to a decision?
\end{enumerate}
Each of the above four questions can be explained by a specific (but not unique) cognitive theory. In what follows, we will discuss them one by one. 

\subsection{Stimulus-Response}
For the first question, `\textbf{How do human drivers react to others' behaviors?}', a straightforward and common answer lies in the stimulus-response mechanism. From the perspective of traffic psychologists, human drivers make interactive responses to a given stimulus according to a specific law
\begin{equation}
	\mathrm{Response} = \mathrm{Sensitivity} \times \mathrm{Stimulus}.
\end{equation}
With such an assumption, many classical interaction models are developed to explain human drivers, such as the classical car-following models (see review papers \cite{brackstone1999car,saifuzzaman2014incorporating}) and lane-change interaction models (see review paper \cite{zheng2014recent}). As we claimed in Section~\ref{subsec:background}, stimulus-response interaction is classified as a \textit{simple} social interaction (e.g., reactive interaction in car-following behaviors), which we will not focus on.

\subsection{Theory of Mind (ToM)}
\label{sec:ToM}
Humans are natural mind-readers; human drivers are endowed with an inborn ability to put themselves in the position of other drivers and reason about their behavior and intentions. Unlike artificial machines, one essential attribute of human capabilities to interact with other human agents is described by the \textit{Theory of Mind} (ToM; \cite{premack1978does}) --- that human beings can reason about other human agents' mental states and actions. One classical realization is the human's capability via
mirror neurons \citep{gallese1998mirror} --- the ability to put ourselves in somebody else's shoes allows us to better interact with our environment and cooperate more efficiently and effectively with our peers \citep{cuzzolin2020knowing}. For example, a rational human driver on the highway would usually leave space and give way to the vehicle on ramps trying to merge. Thus, for the second question, `\textbf{How could the driver be able to understand others' behaviors?}', ToM, as a significant component of social cognition, could be one of the potential answers. In traffic psychology, ToM broadly refers to the human driver's ability to represent the mental states of others, including their desires, beliefs, and intents, without explicit communication. 

\paragraph{`Understanding' as a High-level Model} In natural traffic conditions, what does it actually mean to `\textbf{understand}' other human agents? Rational human drivers rely on their estimation of others' mental states at a high level to make decisions in most cases, which depends on the `understanding' of the environment. Although other human drivers' latent states and information processing procedures are almost entirely inaccessible to the ego human, we can drive efficiently and safely with remarkable adeptness by predicting others' future behavior and inferring what information they have about the world \citep{rabinowitz2018machine}. On the other hand, the ego driver does not refer to the other humans' underlying structures at a relatively micro level, such as the activity of others' neurons and the connectivity of others' prefrontal cortices, but does rely on high-level models of others \citep{gopnik1992child}. Thus, `\textbf{understanding}' other human drivers involves extracting abstract and underlying states at a high level, which can predict their future behavior while explaining for the observed behavior.

\begin{table}
	\centering
	\caption{Comparison of ToM, IRL/RL, and Bayesian ToM.}
	\label{table:ToMcomparison}
	\vspace{2pt}
	\begin{tabular}{lll}
		\hline \hline
		Theory of Mind & IRL/RL & Bayesian Theory of Mind \\
		\hline \hline
		Desires & Reward function & Desires\\
		Beliefs & World model &  Beliefs\\
		Intentions & Policy & Planning\\
		Outcome & State change & World/agent states\\
		Actions & Policy execution & Actions\\
		\hline\hline
	\end{tabular}
\end{table}

\begin{figure}[h!]
	\centering
	\includegraphics[width=\linewidth]{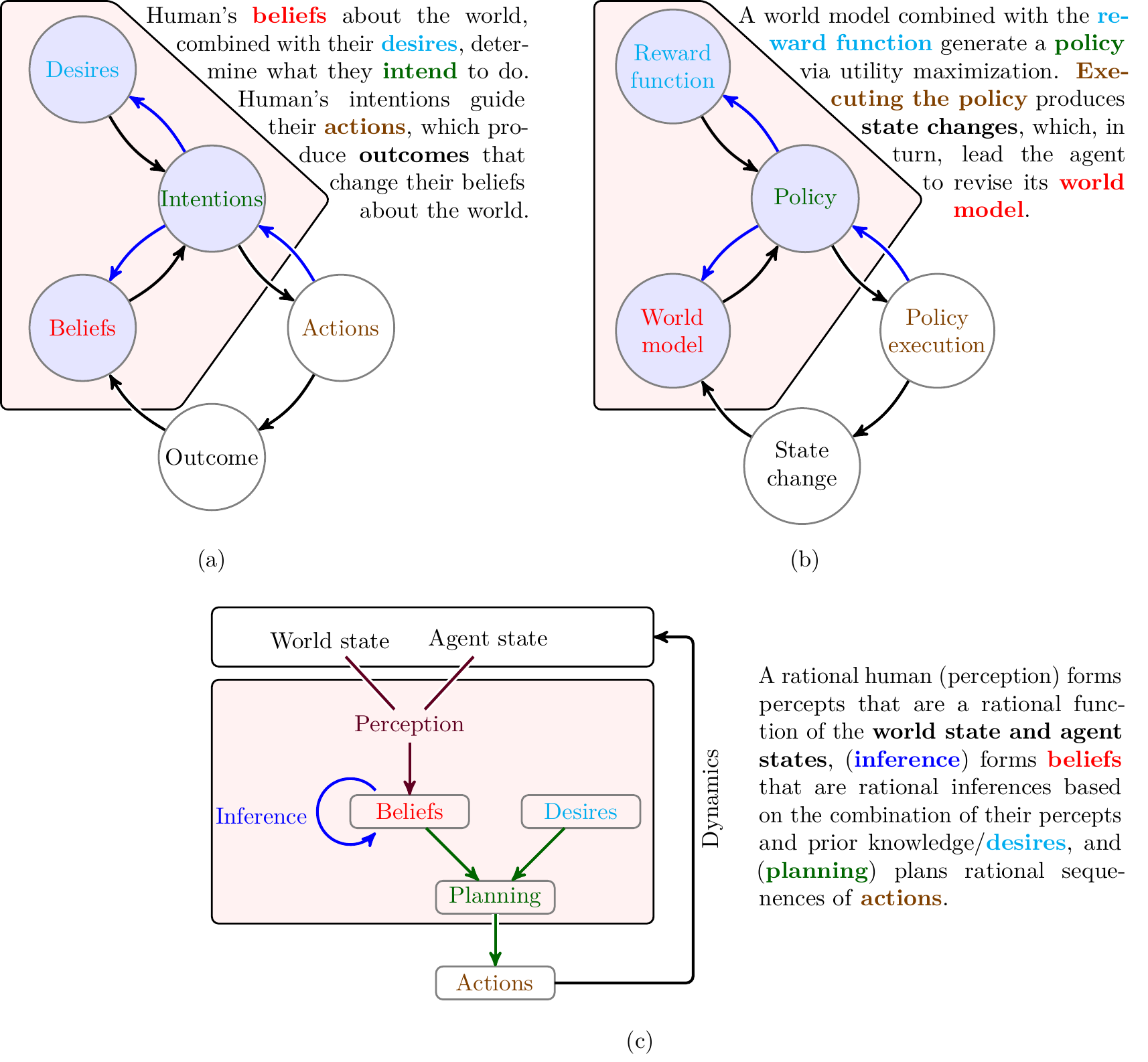}
	\caption{Illustration of how Theory of Mind (ToM) can be modeled as IRL and extended to Bayesian inference. (a) Schema for core ToM components. (b) Schema for core model-based IRL/RL. (c) Schema for Bayesian ToM \citep{baker2017rational}. Pink boxes represent the agent's mind and unobservable states.}
	\label{fig:ToM as IRL}
\end{figure}

\paragraph{Machine ToM (M-ToM)} ToM needs a computationally intelligent machine as a carrier when applying it to solve specific problems \citep{cuzzolin2020knowing}. Machine theory of mind (M-ToM) is the conceptual solution, which refers to a system that models other agents \citep{rabinowitz2018machine}. Several works have applied the M-ToM based on specific theories (e.g., IRL, Bayesian theory, and graphic models) to `understand' and predict other agents' behaviors in interactive traffic scenarios. 
\begin{itemize}
	
	\item \textbf{ToM via inverse reinforcement learning (IRL).} Learning the reward functions from demonstrations by drivers can be formulated via IRL when conceptualized as cognitive theories \citep{jara2019theory} --- reasoning about other drivers' mental states such as driving styles and intents. Many works infer human's latent states and diving styles using IRL \citep{sun2018probabilistic, ozkan2021socially, rosbach2020driving} and then integrate them with other modules such as game theory \citep{tian2021learning} and POMDP \citep{luo2021interactive}. For instance,  \cite{tian2021learning} utilized IRL to infer other drivers' latent intelligence levels and then integrated these levels into a game-theoretic framework to formulate the mutual influence between agents. The realization of IRL-based ToM has certain major limitations. It requires extensive training data, and the learned reward functions are unrealistically tailored to the observed scenarios.
	
	\item \textbf{ToM via Bayesian theory.} Humans estimate other human agents' beliefs and desires based on what they can see, which can be realized via Bayesian inference, given the observed trajectories and the prior probability distributions over their initial beliefs and desires \citep{baker2014modeling}. The Bayesian ToM assumes that the agent's behavior is generated by approximately rational belief-dependent planning and desire-dependent planning, maximizing their rewards in partially observable domains. The frequently used formal and computational model of recovering such mechanisms is POMDP \citep{baker2014modeling, baker2017rational}.
	
	\item \textbf{ToM via graph-theoretic.} An expected M-ToM should universally model human driving behavior in all traffic scenarios rather than domain-shifting issues.  \cite{chandra2020stylepredict,chandra2021using} developed a computational model for M-ToM based on graph theory, named graph-theoretic M-ToM, to interpret driving styles according to their trajectories. The graph-theoretic M-ToM is more robust than the IRL-based ToM without needing for parameter-tuning or manual adjustment.
\end{itemize}

Figure~\ref{fig:ToM as IRL}(a) and (b) compare the structures of ToM and IRL/RL and indicate that ToM can be specified under an IRL/RL structure. The blue arrows represent the human mental-state inference (i.e., inferring human's (unobservable) beliefs and desires given some observed actions, see Figure~\ref{fig:ToM as IRL}(a)), which corresponds to the problem in IRL (i.e., inferring agent's unobservable model of the world and reward function, given some observed policy executions, see Figure~\ref{fig:ToM as IRL}(b)). Moreover, the ToM can formalize human mentalizing as Bayesian inference about the unobservable variables (i.e., beliefs, desires, and percepts) given the observed actions (Figure~\ref{fig:ToM as IRL} (c)), which is a generative model for action. Table~\ref{table:ToMcomparison} also illustrates the connection between the core components of ToM, IRL/RL, and Bayesian ToM. 

On the other hand, the ToM also provides game-theoretic methods and assumptions to make them computationally tractable when modeling interactions. For example, \cite{schwarting2021stochastic} assume that any agent $i$'s belief over another agent $j$ is the same as that agent $j$'s belief about themselves, which thus enables interactive predictions of other agents in belief space while maintaining computational tractability in the scheme of stochastic games. 

\subsection{Information Accumulation Mechanisms}
For the fourth question, `{\textbf{How does the human driver process the perceptual information while arriving at a decision?}}',  one of the potential answers lies in the natural mechanisms of perceptual information accumulation. 

\paragraph{Evidence Accumulation} The drift-diffusion model (DDM; \cite{fudenberg2020testing}) is a sequential sampling model with diffusion signals. A decision-maker accumulates evidence until the process hits an upper or lower boundary and then stops and chooses the alternative corresponding to that boundary. One typical example could be the decision-making process while merging into a traffic flow at an unsignalized roundabout. A driver will wait at the roundabout on the secondary road until a target `gap' is acceptable. The waiting time is an accumulated indicator of how much probability the driver will take a risky decision to merge: The longer the waiting time, the more risky decisions would be made. Many existing works only focus on \textit{what} decisions the driver would make, such as behavior modeling and prediction \citep{pollatschek2002decision} but less on \textit{how} this process is operated and the decision is triggered with more and more information observed. \cite{zgonnikov2020should} borrowed the conceptional mechanisms of evidence accumulation from the fields of  behavioral and cognitive science to investigate \textit{how} a human driver processes the perceptual information of on-coming vehicles before making a socially-compatible left-turn decision in the unprotected left-turn interactive scenarios.

Note that for revealing how the perceptual information is used while arriving at a decision (i.e., perceptual decision making), there generally exist four decision theories in the behavioral sciences \citep{orquin2013attention}: rational models, rounded rationality, evidence accumulation, and parallel constraint satisfaction models. Until now, however, only the evidence accumulation has been used in explaining human driving behavior in traffic scenarios \citep{durrani2021predicting}.

\subsection{Summary}
Computational cognition brings a new view of understanding the essence of various cognitive functionalities in social interaction among human drivers. Some works have been done to explain human driver's decision-making process in interactions by borrowing the ideas and findings from cognitive computational neuroscience. However, on the one hand, existing approaches typically provide excessive detail, for example, describing low-level neurological phenomena that make models too large. On the other hand, some are too simplistic to be tractable in practice, for example, only testing in a laboratory setting and lacking principled investigations in complex driving contexts \textit{or} providing a high-level cognitive structure of making decisions. Therefore, a challenging gap exists between primary laboratory mechanisms and practical applications for implementing these mechanisms in real-time predicting human driving behavior.

\chapter{Discussions}
\label{sec:discussion}
We have reviewed works on how human drivers interact with others in interactive traffic scenes using implicit communications, defining the social interaction and showing how researchers developed interpretable models to formulate such social interactions. Next, we will outline some open questions and possible future directions. 

\section{Are the Social Interaction Models Really Socially-aware?}
Most works develop interaction models under a well-defined structure of mutual influences based on assumptions and prior knowledge of interactions, such as the attention mechanisms. These models finally provide expected performance (e.g., the low prediction errors of motion trajectories, driving maneuvers, and intents) in the light of various advanced AI algorithms. However, they failed to investigate the adversarial robustness and assess the alignment of those models and the social behaviors in terms of social understanding. \cite{saadatnejad2021socially} investigated the robustness of different neural network-based trajectory predictors by introducing a socially-attended attack. Social understanding can reveal the limitations of the current models, thus providing possible future directions. 

\section{Shifts between Model Assumptions and Data Sets.} Most researchers trained and calibrated their model parameters based on real-world data and obtained promising performance as expected in specific environments. However, the model robustness is still a core problem in many transportation applications. This might be caused by the mismatch between designed models and generated data, including two aspects. 
\begin{itemize}
	\item Interaction model design is \textbf{goal-oriented}, but the behaviors generated by human drivers are primarily \textbf{habitual}. The behavior is initially goal-directed but then becomes habitual over the course experience \citep{dolan2013goals}. For example, a new driving learner starts planning their actions with a series of \textit{specific} sub-goals decomposed by a teacher for a given driving task (e.g., parking). Then, the driver digests these steps and transfers them into habits without such specific sub-goals. Training a goal-directed model based on a habitually generated dataset might weaken the model's generalizability. One possible solution to the shift between model assumptions and data might be elaborating model-based and model-free frameworks.
	\item Interaction model design is social-directed, but behavior data are generated with a mixture of physical and social interactions. When developing interaction models, most works directly modeled their interaction trajectories, such as using graphical neural networks (GNN) but neglected the insights into why and how the interaction behavior is generated. Counterfactual only uses a social interaction-assumed model to approximate the social- and physical interaction behaviors. 
\end{itemize}
Therefore, we argue that insights into pertinent \textbf{data} to power learning algorithms could be one of the keys to making autonomous vehicles socially-compatible and robust. Note that we are not emphasizing that methods should be solely data-driven but suggesting that knowledge from data analysis would be critical for designing suitable theoretical frameworks and practical algorithms toward interaction. 

\section{Can Cognitive Science Help Make AVs Socially-compatible?} Perfect autonomous cars require intuitive psychology beyond path tracking, object detection, and collision avoidance.  Due to the technical and social barriers \citep{hutson2017matter, bonnefon2016social,martinho2021ethical}, it is too early to bring mindless machines such as autonomous cars without intuitive psychology on the roads and share road space with other human drivers \citep{lepore2009not}.  A socially-compatible autonomous vehicle should be able to safely and acceptably interact with other road users by uncovering other human agents' mental states and beliefs. In other words, the final goal is to build autonomous vehicles that think and learn like rational human drivers. However, genuinely human-like learning and thinking capabilities are beyond current engineering trends in both what they learn and how they learn it \citep{lake2017building}. The opening question is, `\textit{Can this be achieved?}' The positive answers rely on not only perception and control but the cognitive foundations of social interaction that are far more difficult \citep{chater2018negotiating}. Therefore, socially-compatible autonomous driving forces us to develop efficient tools to explain and understand the underlying decision-making process from observations using structured, computational cognitive models.

\section{The More Accurate a Trajectory Prediction Model, The Better?}
Agent behavior prediction and inference have been recognized as an indispensable part of safety-critical interactive system design; for example, autonomous vehicles negotiating in human environments need to leverage the future behavior of human-driven vehicles into their planning and decisions. More existing works are keen to pursue higher accuracy trajectory forecasting of moving cars in complex scenarios and have made significant progress at the beginning but small later. For example, designing a simple forecasting algorithm with a low effort can improve a lot (e.g., from $40\%$ to $80\%$) when starting to solve the prediction problem but now only improves performance by less than $3\%$ even with a significant effort to design and fine-tune \citep{li2021prediction,jia2021ide}. In practice, instead of blindly pursuing trajectory forecasting accuracy, we need carefully think about `\textit{how accurate do the models need to be for successful interaction?}'  It is essential to consider what aspects of human cognition and behavior have the most significant impact on interaction performance \citep{markkula2022accurate}, which requires dedicated empirical and modeling work. For different practical tasks, making a very high accurate trajectory forecasting may not benefit the whole interaction performance, although it might benefit the low-level controller design. Human drivers may not make a prediction as accurate as a machine (e.g., an autonomous vehicle) but they still make an efficient interaction. Here, we did not argue that an accurate trajectory prediction model is unnecessary but that figuring out which aspects are the key aspects of socially compatible interactions is more important; for instance, predicting goals/objectives and using them in game theoretic frameworks as an attention \citep{peters2021inferring}. 

\chapter{Conclusions}
\label{sec:conclusion}
Understanding how human drivers interact with others is a central problem in developing socially-compatible autonomous vehicles. Having quantitative models to predict these interactive behaviors has become increasingly important as autonomous vehicles interact more closely with other human agents on social roads. This paper first provided a clear definition of social interactions in road traffic. It then provided an inevitably selective review of methods and models of interactions between human drivers and associated applications, including rational utility-based models, deep learning-based models, graph-based models, social fields/forces, and computational cognitive models. Critical findings and open questions are finally posed, which could provide new directions for interactions of autonomous vehicles with human-driven vehicles. 

\begin{acknowledgements}
We would like to acknowledge support for this project from 2020 IVADO Postdoctoral Fellowship Awards (IVADO-PostDoc-2020a-5297372919) and IVADO MSc Excellence Scholarship (IVADO-MSc-2020-0841321446). The views expressed here are the authors' and do not reflect the funding bodies.
\end{acknowledgements}

\appendix
\chapter{Markov Decision Processes and Markov Games}\label{app:MarkovGames}
\vspace*{-1.2in}

Section \ref{sec:game_theory} discusses some works that utilize various game-theoretical frameworks to formulate the interaction among agents; each agent can be modeled by employing a Markov decision process (MDP). In what follows, we will briefly revisit MDPs and Markov games.

\begin{definition}[Markov Decision Processes]
	An MDP can be described by a tuple of key elements, $\langle s, a, p, r, \gamma\rangle$ with
	\begin{itemize}
		\item $s\in \mathcal{S}$ --- The environment states in the state space $\mathcal{S}$.
		\item $a\in \mathcal{A}$ --- Agent's possible actions in the action space  $\mathcal{A}$.
		\item $p:\mathcal{S} \times \mathcal{A} \rightarrow \mathcal{S}$ --- For each time step $t>0$, given an agent's action $a\in \mathcal{A}$, the transition probability from a state $s\in\mathcal{S}$ to the state $s^{\prime}\in\mathcal{S}$ in the next time step.
		\item $r: \mathcal{S} \times \mathcal{A} \times \mathcal{S} \rightarrow \mathbb{R}$ --- The reward function that returns a scalar value (i.e., $r\in \mathbb{R}$) to the agent for a state transition from $s$ to $s^{\prime}$ after taking an action $a$. 
		\item $\gamma \in [0,1]$ --- The discount factor that represents the value of time.
	\end{itemize}
\end{definition}

\begin{definition}[Markov Games or Stochastic Games]
	A stochastic game can be viewed as a multiplayer extension to the Markov decision process, constituted of a set of key elements $\langle N, s, \{a_{i}\}_{i=1}^{N}, P, \{r_{i}\}_{i=1}^{N}, \gamma \rangle$, with 
	\begin{itemize}
		\item $N\in\mathbb{N}^{+}$ --- The number of agents.
		\item $s\in \mathcal{S}$ --- The environment states shared by all agents, over the state space $\mathcal{S}$. 
		\item $a_{i}\in\mathcal{A}_{i}$ --- The $i$-th agent's action in its action space $\mathcal{A}_{i}$.
		\item $p: \mathcal{S}\times\mathcal{A}_{1}\times \mathcal{A}_{2}\times \dots \times \mathcal{A}_{N}\rightarrow \mathcal{S}$ --- For each time step, given agents' joint actions $\boldsymbol{a}=[a_{1}, a_{2}, \dots, a_{N}]$, the transition probability from state $s\in\mathcal{S}$ to state $s^{\prime}\in\mathcal{S}$ in the next time step. 
		\item $r_{i}: \mathcal{S}\times\{\mathcal{A}_{i}\}_{i=1}^{N}\times\mathcal{S}$ --- The reward function that returns a scalar value to the $i$-th agent for a transition from $(s, \boldsymbol{a})$ to $s^{\prime}$.
		\item $\gamma$ --- The discount factor that represents the value of time. 
	\end{itemize}
	Each agent $i$ aims to maximize its expected discounted total rewards with the starting state $s_{0}$ at time $t$
	\begin{equation}
		V_{\pi_{i}}(s) = \mathbb{E} \left[ \sum_{\tau=0}^{} \gamma^{\tau} r_{i}^{\tau + t} | s_{t} = s_{0}\right].
	\end{equation}
\end{definition}

\chapter{Graph Models}
\label{app:graphmodels}

\begin{figure}[t]
	\centering
	\includegraphics[width = 0.8\linewidth]{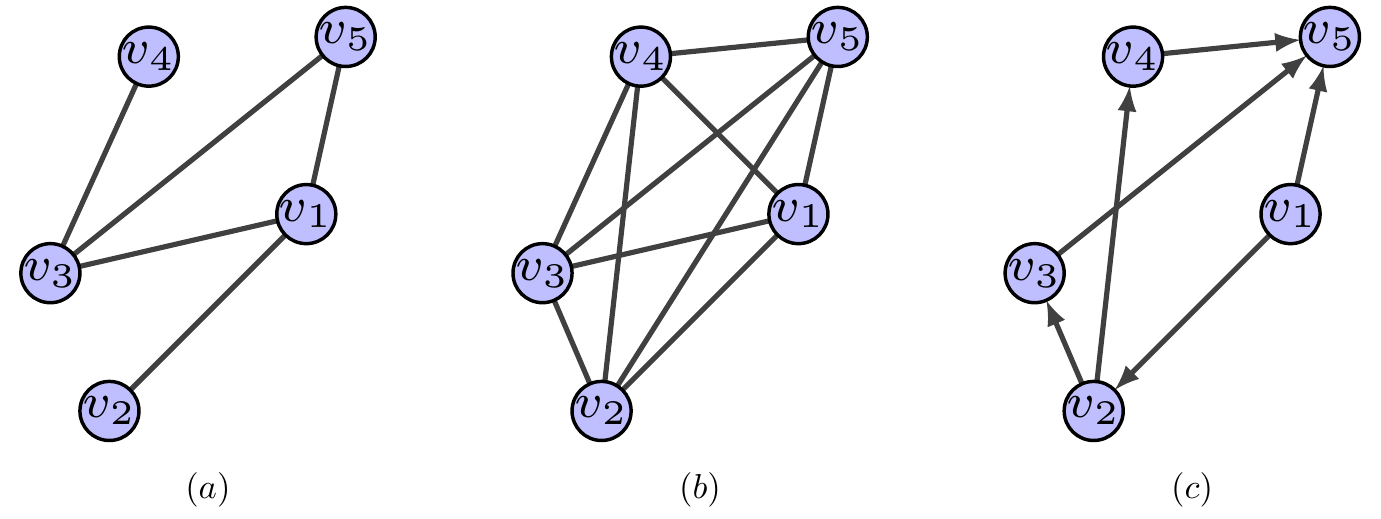}
	\caption{Examples of three types of graphs with five nodes. (a) incomplete undirected graph, (b) complete graph (or fully connected graph), and (c) directed graph.}
	\label{fig:graphmodels}
\end{figure}

\begin{definition}[Graph/Digraph] A graph $\mathbf{G}$ is a pair $(\mathcal{V}, \mathcal{E})$, where $\mathcal{V}$ is a finite set of vertices, and $\mathcal{E}$ is a set of edges. If each edge is an \textit{ordered} pair $(u, v)$ of nodes with $(u, v)\in \mathcal{V}\times\mathcal{V}$, $u\neq v$, the graph is called digraph.
\end{definition}

\begin{definition}[Node Network] A node network (or node weighted graph) is a triple of $(\mathcal{V}, \mathcal{E}, f)$, where $f$ is a function mapping \textit{vertices} to numbers, $f: \mathcal{V} \rightarrow \mathcal{N}$, where $\mathcal{N}$ is some number system, assigning a value (or weight) which may be real (or complex) numbers. 
\end{definition}

\begin{definition}[Edge network] A edge network (or edge weighted graph) is a is a triple $(\mathcal{V}, \mathcal{E}, g)$, where $g$ is a function mapping \textit{edges} to numbers. 
\end{definition}

In general, a graph or network consists of the following entities: a set of vertices $\mathcal{V}$, a set of edges $\mathcal{E}$, a function mapping vertices to numbers $f$, a function mapping edges to numbers $g$. A \textit{dynamic graph} is obtained when any of these four entities changes over time (\cite{harary1997dynamic}).

\begin{definition}[Edge network] A edge network (or edge weighted graph) is a is a triple $(\mathcal{V}, \mathcal{E}, g)$, where $g$ is a function mapping \textit{edges} to numbers. 
\end{definition}

\begin{definition}[Adjacency, degree, and Laplacian matrices] Figure~\ref{fig:graphmodels}(a) represents an undirected graph with five nodes $v_{1}\sim v_{5}$ and five edges $(v_{1},v_{2})$, $(v_{1},v_{3})$, $(v_{1},v_{5})$, $(v_{3},v_{4})$, and $(v_{3},v_{5})$. The adjacency ($\mathbf{A}$), degree ($\mathbf{D}$), and Laplacian ($\mathbf{L} = \mathbf{D}-\mathbf{A}$) matrices for the graph is 
	\begin{equation}
		\begin{split}
			\mathbf{A} & = 
			\begin{bmatrix}
				0 & 1 & 1 & 0 & 1 \\
				1 & 0 & 0 & 0 & 0 \\
				1 & 0 & 0 & 1 & 1 \\
				0 & 0 & 1 & 0 & 0 \\
				1 & 0 & 1 & 0 & 0 \\
			\end{bmatrix}, \\
			\mathbf{D} & = 
			\begin{bmatrix}
				3 & 0 & 0 & 0 & 0 \\
				0 & 1 & 0 & 0 & 0 \\
				0 & 0 & 3 & 0 & 0 \\
				0 & 0 & 0 & 1 & 0 \\
				0 & 0 & 0 & 0 & 2 \\
			\end{bmatrix}, \\
			\mathbf{L} & = 
			\begin{bmatrix}{r}
				3 & -1 & -1 & 0 & -1 \\
				-1 & 1 & 0 & 0 & 0 \\
				-1 & 0 & 3 & -1 & -1 \\
				0 & 0 & -1 & 1 & 0 \\
				-1 & 0 & -1 & 0 & 2 \\
			\end{bmatrix}.
		\end{split}
	\end{equation}
	
\end{definition}

Of course, the entries of the adjacency matrix $\mathbf{A} = \{a_{i,j}\}$ can be the measurement (i.e., a real value as a weight, formed a weighted matrix) of the interaction intensity over time, for example, by assigning the entries $a_{i,j}^{(t)}$ at the time frame $t$ as a function of the relative distance between two vehicles $i$ and $j$ (\cite{cao2021spectral})
\begin{equation}
	a_{i,j}^{(t)}=
	\begin{cases} 
		1 / ||\tau_{t}^{i} - \tau_{t}^{j}||_{2}, & i\neq j, \\
		0, & \mathrm{otherwise}.
	\end{cases}
\end{equation}
where $\tau_{t}^{i}$ and $\tau_{t}^{j}$ are the positions of agents $i$ and $j$ at time $t$.
The introduced matrices ($\mathbf{A}$, $\mathbf{D}$, and $\mathbf{L}$) allow us to capture the interactions between human drivers under a graph structure.

\chapter{Attention Measure}
\label{app:attenion}

Assume that individual human agent's behavioral information (or observed data) can be sufficiently encoded in a compact manner such as into a low-dimensional vector, denoted as $\boldsymbol{h}$, thus the relationship or influence between any two agents with corresponding vectorized entries ($\boldsymbol{h}_{i}\in\mathbb{R}^{d}$ for agent $i$ and $\boldsymbol{h}_{j}\in\mathbb{R}^{d}$ for agent $j$) can be quantified by a function $f$

\begin{equation*}
	\alpha_{i,j} = f(\boldsymbol{h}_{i}, \boldsymbol{h}_{j})
\end{equation*}
Generally, there are five frequently used quantification measures, and all of them are basically based on the operations on the \textit{projections} of the vectors $\boldsymbol{h}_{i}$ and $ \boldsymbol{h}_{j}$, i.e., dot production mathematically. 

\section{Content-based Attention}
\begin{figure}[h]
	\centering
	\includegraphics[width=0.5\linewidth]{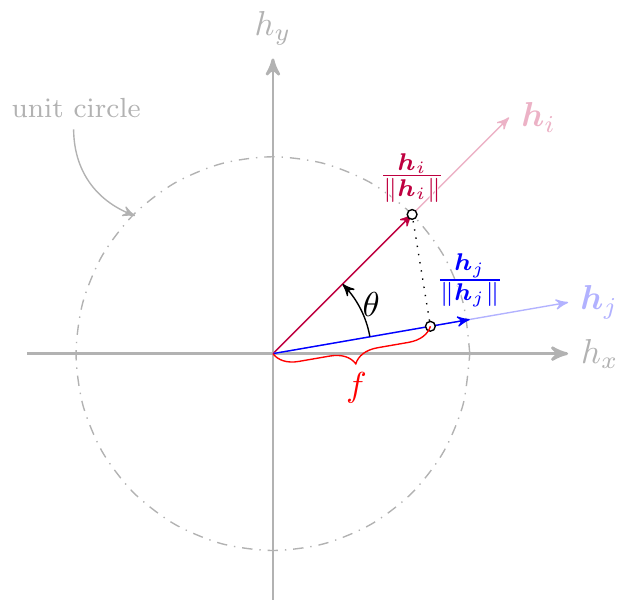}
	\caption{Illustration of content-based attention with $\boldsymbol{h}_{i}$, $\boldsymbol{h}_{j}$ in a 2D plane.}
	\label{fig:content}
\end{figure}

The idea of content-based attention is to quantify the similarity level using the cosine similarity by projecting one attention vector to the other one \citep{graves2014neural}:
\begin{equation}
	f(\boldsymbol{h}_{i}, \boldsymbol{h}_{j}) = \cos (\boldsymbol{h}_{i}, \boldsymbol{h}_{j}) = \frac{\boldsymbol{h}_{i}^{\top} \boldsymbol{h}_{j}}{\parallel\boldsymbol{h}_{i}\parallel \parallel\boldsymbol{h}_{i}\parallel} = \frac{\boldsymbol{h}_{i}^{\top}}{\parallel\boldsymbol{h}_{i}\parallel} \frac{\boldsymbol{h}_{j}}{\parallel\boldsymbol{h}_{j}\parallel} 
\end{equation}
The resulting similarity values $f\in[-1,1]$ with interpretation: $-1$ indicates exactly opposite, and $1$ indicates exactly the same, and $0$ indicates decorrelation, while in-between values indicate intermediate similarity or dissimilarity. Figure~\ref{fig:content} geometrically visualizes the resulting values of the content-based attention, where $\theta$ denotes the angle between two vectors $\boldsymbol{h}_{i}$ and $\boldsymbol{h}_{j}$.

\section{Concatenation Attention}

The idea of concatenation attention is projecting the normalized weighted summation of two hidden vectors $\boldsymbol{h}_{i}$ and $\boldsymbol{h}_{j}$ to a defined value $\mathbf{v}_{a}$ \citep{bahdanau2014neural}:

\begin{equation}\label{eq:concatentation attention}
	f(\boldsymbol{h}_{i}, \boldsymbol{h}_{j}) = \mathbf{v}^{\top}_{a} \tanh \left( \mathbf{W}_{a} [\boldsymbol{h}_{i} \oplus \boldsymbol{h}_{j}]\right)
\end{equation}
where $[\boldsymbol{h}_{i} \oplus \boldsymbol{h}_{j}]$ is the concatenation operation\footnote{In Luong's \citep{luong2015effective} and Bahdanau's \citep{bahdanau2014neural} notations, they used a semicolon operator in the formulas to denote concatenation, i.e., $[\boldsymbol{h}_{i} ; \boldsymbol{h}_{j}]$.} and $\mathbf{W}_{a}$ is a weight matrix that projects the concatenated vector to a different vector. 

On the other hand, the projection of concatenation of two vectors in (\ref{eq:concatentation attention}) have an equivlent form as

\begin{equation}
	\mathbf{W}_{a} [\boldsymbol{h}_{i} \oplus \boldsymbol{h}_{j}] = 
	\begin{bmatrix}
		\mathbf{W}_{a,i} & \mathbf{W}_{a,j} \\
	\end{bmatrix}
	\begin{bmatrix}
		\boldsymbol{h}_{i} \\
		\boldsymbol{h}_{j} \\
	\end{bmatrix}
	= \mathbf{W}_{a,i} \boldsymbol{h}_{i} + \mathbf{W}_{a,j} \boldsymbol{h}_{j}
\end{equation}
where $\mathbf{W}_{a} = [ \mathbf{W}_{a,i} \oplus \mathbf{W}_{a,j} ]$. The above equation directly follows from the definition of matrix multiplication. Therefore, the concatenation attention can also be presented in an equivalent form as 

\begin{equation}
	f(\boldsymbol{h}_{i}, \boldsymbol{h}_{j}) = \mathbf{v}^{\top}_{a} \tanh \left( \mathbf{W}_{a,i} \boldsymbol{h}_{i} + \mathbf{W}_{a,j} \boldsymbol{h}_{j}\right)
\end{equation}

\section{General Attention}
The idea of general attention is straightforward proposed by \cite{luong2015effective}, 

\begin{equation}
	f(\boldsymbol{h}_{i}, \boldsymbol{h}_{j}) = \boldsymbol{h}_{i}^{\top} \mathbf{W}_{a}  \boldsymbol{h}_{j}
\end{equation}
which can be viewed as the projection of a linear transformation of one vector $\boldsymbol{h}_{i}$ over matrix $\mathbf{W}_{a}$ to another vector $\boldsymbol{h}_{j}$. More specifically, we have follows

\begin{equation*}
	\begin{split}
		\boldsymbol{h}_{i}^{\top} \mathbf{W}_{a}  \boldsymbol{h}_{j} & = 
		\begin{bmatrix}
			h_{i1} &  h_{i2} & \dots & h_{id}
		\end{bmatrix}
		\begin{bmatrix}
			w_{11} & w_{12} & \dots & w_{1d} \\
			w_{21} & w_{22} & \dots & w_{2d} \\
			\vdots & \vdots & \ddots & \vdots \\
			w_{d1} & w_{d2} & \dots & w_{dd} \\
		\end{bmatrix}
		\begin{bmatrix}
			h_{j1} \\
			h_{j2} \\
			\vdots \\
			h_{jd}
		\end{bmatrix} \\
		& = 
		\underbrace{\begin{bmatrix}
				\sum_{\ell=1}^{d} h_{i\ell} w_{\ell 1} & \sum_{\ell=1}^{d} h_{i\ell} w_{\ell 2} & \dots & \sum_{\ell=1}^{d} h_{i\ell} w_{\ell d}
		\end{bmatrix}}_{\text{Linear transformation of}\ \boldsymbol{h}_{i}}
		\begin{bmatrix}
			h_{j1} \\
			h_{j2} \\
			\vdots \\
			h_{jd}
		\end{bmatrix} \\
		& = \sum_{k=1}^{d} \sum_{\ell=1}^{d} h_{i\ell}w_{\ell k} h_{jk}
	\end{split}
\end{equation*}

\section{(Scale) Dot-Product Attention}

\paragraph{Dot-product} The dot-product attention \citep{luong2015effective} is more implementation-friendly and can be viewed as a special case of the general attention with the transformation matrix $\mathbf{W}_{a}$ as an identity matrix, i.e., $\mathbf{W}_{a} = \mathbf{I}_{d\times d}$, obtaining 

\begin{equation}\label{eq:dot product}
	f(\boldsymbol{h}_{i}, \boldsymbol{h}_{j}) = \boldsymbol{h}_{i}^{\top} \boldsymbol{h}_{j}
\end{equation}

\paragraph{Scaled Dot-product} The scaled dot-product \citep{vaswani2017attention} is a variant of the dot-product attention with an additional scaling factor $\frac{1}{\sqrt{d}}$

\begin{equation}\label{eq:scaled dot product}
	f(\boldsymbol{h}_{i}, \boldsymbol{h}_{j}) = \frac{\boldsymbol{h}_{i}^{\top} \boldsymbol{h}_{j}}{\sqrt{d}}
\end{equation} 
For a small value of $d$,  (\ref{eq:scaled dot product}) performs similarly with the dot-product attention but will outperform (\ref{eq:dot product}) for a large value of $d$ when feeding into a softmax function.

\section{(Embedded) Gaussian Attention}

\paragraph{Gaussian} The Gaussian-based attention is operating the dot-product over a Gaussian function \citep{wang2018non}
\begin{equation}
	f(\boldsymbol{h}_{i}, \boldsymbol{h}_{j}) = e^{\boldsymbol{h}_{i}^{\top} \boldsymbol{h}_{j}}
\end{equation}

\paragraph{Embedded Gaussian}
A simple extension of the Gaussian function is to compute the similarity in an \textit{embedding space}, for example

\begin{equation}
	f(\boldsymbol{h}_{i}, \boldsymbol{h}_{j}) = e^{\theta(\boldsymbol{h}_{i})^{\top} \phi(\boldsymbol{h}_{j})}
\end{equation}
where $\theta(\cdot)$ and $\phi(\cdot)$ are embeddings. for example, linear transformations
\begin{subequations}
	\begin{align}
		\theta(\boldsymbol{h}_{i})  = & \mathbf{W}_{\theta} \boldsymbol{h}_{i} \\
		\phi(\boldsymbol{h}_{j}) = & \mathbf{W}_{\phi} \boldsymbol{h}_{j}
	\end{align}
\end{subequations}
Note that the embedded Gaussian attention can be viewed as re-shaping the similarity of the general attention over a Gaussian function since

\begin{equation}
	e^{\theta(\boldsymbol{h}_{i})^{\top} \phi(\boldsymbol{h}_{j})} = e^{(\mathbf{W}_{\theta} \boldsymbol{h}_{i})^{\top} \mathbf{W}_{\phi} \boldsymbol{h}_{j}} = e^{\boldsymbol{h}_{i}^{\top}\mathbf{W}_{\theta}^{\top}\mathbf{W}_{\phi} \boldsymbol{h}_{j}} = e^{\boldsymbol{h}_{i}^{\top}\mathbf{W}_{a}\boldsymbol{h}_{j}}
\end{equation}
with $\mathbf{W}_{a} = \mathbf{W}_{\theta}^{\top}\mathbf{W}_{\phi}$.

\chapter{Topological Braids}
\label{app:topologicalBraids}
\section{Braids}
\textbf{Braids} are topological objects with algebraic and geometric presentations, which is usually denoted by the  Cartesian coordinates $(x,y,z)$ of a Euclidean space $\mathbb{R}^{2}\times I$. A \textbf{braid string} is a curve 

\begin{equation}
	\label{eq:a braid string}
	\begin{split}
		\mathrm{\textsf{Br}}(z): I & \rightarrow \mathbb{R}^{2} \\
		z & \rightarrow x \times y
	\end{split}
\end{equation}
that increases monotonically in $z$, for example, $z$ could be time $t\in[0,\infty)$. That is, a braid string has exactly one point of intersection $\mathrm{\textsf{Br}}(z) = (x,y)$ with each plane $z\in I$. 

\begin{figure}[h]
	\centering
	\includegraphics[width=\linewidth]{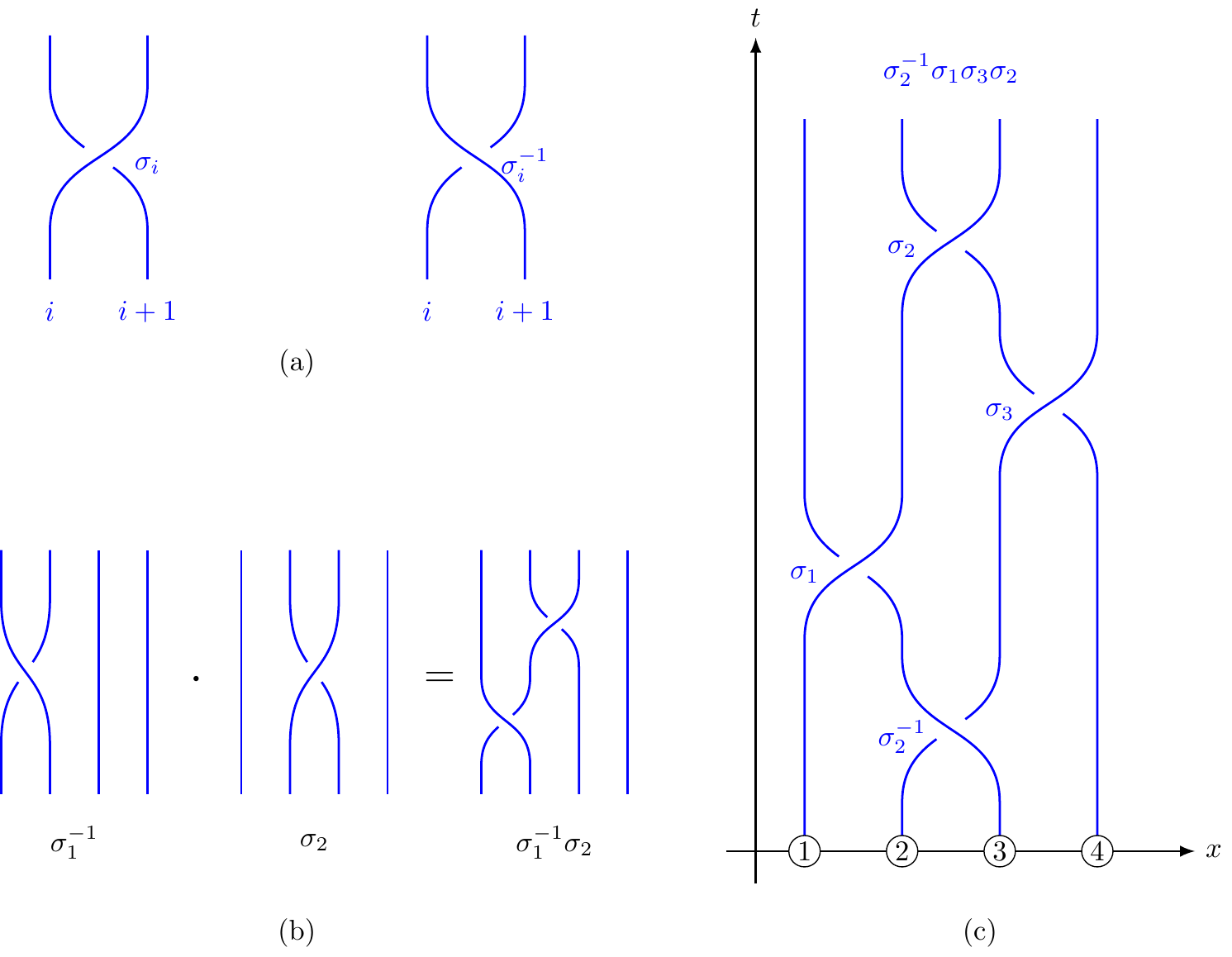}
	\caption{Braid diagram. (a) Braid generators (or primitives). (b) Algebraical operations. (c) Topological braid diagram: An example of a braid that can be written as a product of generators and generator inverses.}
	\label{fig:braids}
\end{figure}

A \textbf{braid on $n$-strings} or $n$-braids is a set of $n$ strings $\mathrm{\textsf{Br}}_{i}(z)$, $i\in \{1, 2, \dots, n\}$, i.e., $\mathcal{B}(z) = \{\mathrm{\textsf{Br}}_{i}(z)\}$, for which some properties hold:

\begin{enumerate}
	\item [(i)] $\mathrm{\textsf{Br}}_{i}(z)\neq\mathrm{\textsf{Br}}_{j}(z)$ for $i\neq j, \ \forall z\in \mathbb{R}$;
	\item [(ii)] $\mathcal{B}(0) = (i,0)$ and $\mathcal{B}(1) = (\mathrm{\textsf{permu}}(i),0)$
\end{enumerate}
where $\mathrm{\textsf{permu}}(i)$ is the image of an element $i \in N$ through a permutation $\mathrm{\textsf{permu}}: N \rightarrow N$ from the set of permutations of $N$, 

\begin{equation}
	\mathrm{\textsf{permu}} = 
	\begin{bmatrix}
		1 & 2 & \cdots & N \\
		\mathrm{\textsf{permu}}(1) & \mathrm{\textsf{permu}}(2) & \cdots & \mathrm{\textsf{permu}}(N) \\
	\end{bmatrix}
\end{equation}
This geometric representation of a braid is commonly treated as a \textbf{geometric braid}. In applications, a geometric braid is often represented with a \textbf{braid diagram} --- a projection of the braid onto the plane $\mathbb{R}\times 0 \times I$.

The \textbf{set of all braids} on $n$-strings, along with the composition operation over elementary braids (called braid primitives or generators), form a group $B_{n}$ generated from a set of $n-1$ elementary braids $\sigma_{1}, \sigma_{2}, \dots, \sigma_{n-1}$. Figure~\ref{fig:braids} illustrates the relationship between braid generators, algebraical operations, and braid diagrams. 

A \textbf{braid generator} $\sigma_{i}$ is described as the crossing pattern that emerges upon exchanging the $i$-th string (counted from left to right) with the ($i+1$)-th string, such that the initially left string passes over the initially right one. The inverse element $\sigma_{i}^{-1}$ implements the same string exchange but with the left string passing under the right (see Figure~\ref{fig:braids} (a)).

\section{Joint Strategy}
\label{app:jointstrategy}
In the multiagent cooperative navigation system, a \textbf{joint strategy} refers to a sequence of strategy profiles of all rounds, at each of which each agent $i$ decides an action $a_{i}^{k}$ from a set of available actions $\mathcal{A}_{i}^{k}$ by maximizing their utilities. Therefore, a joint strategy, $\tau$, of a cooperative game can be represented as

\begin{equation}
	\tau = \left[\tau_{1}\tau_{2}\dots\tau_{K}\right] = \left[A_{1}A_{2}\dots A_{K}\right]=
	\begin{bmatrix}
		a_{1}^{1} & \cdots &  a_{1}^{k} & \cdots & a_{1}^{K} \\
		\vdots & \vdots & \vdots & \vdots & \vdots \\
		a_{i}^{1} & \cdots &  a_{i}^{k} & \cdots & a_{i}^{K} \\
		\vdots & \vdots & \vdots & \vdots & \vdots \\
		a_{N}^{1} & \cdots &  a_{N}^{k} & \cdots & a_{N}^{K}
	\end{bmatrix}
\end{equation}

\begin{figure}[h]
	\centering
	\includegraphics[width=0.5\linewidth]{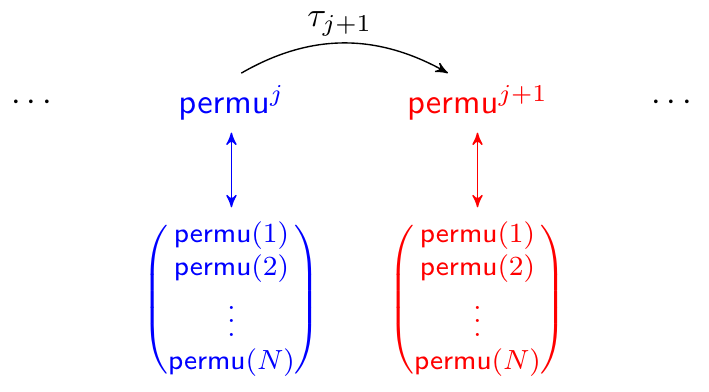}
\end{figure}

By treating the whole navigation process as a geometric braid, each agent trajectory profile represents a string of the braid. Thus, each round can be represented by operating on the elementary braid. Specifically, 
assume $N$ agents generating the whole system path corresponds to a path of permutations. A transition from the $j$-th permutation $\mathrm{\textsf{permu}}^{j}$ to the $(j+1)$-th permutation $\mathrm{\textsf{permu}}^{j+1}$ refers to the occurrence of an \textbf{event} $\tau_{j+1}$. The event $\tau_{j+1}$ may be represented as an elementary braid $\tau_{j+1}\in \{\sigma_{j+1}, \sigma_{j+1}^{-1}\}$.  

\section{Topological Invariance}
\label{app:topologicalinvariance}
Consider a closed curve $\rho: [0,T]\rightarrow \mathbb{C}\setminus \{0\}$ with $\rho (0) = \rho(T)$ and a well-defined function

\begin{equation}
	\label{eq:closed curve}
	\lambda(t) = \frac{1}{2\pi i} \oint_{\rho} \frac{dz}{z}
\end{equation}
where $z=\rho(t)$, $t\in [0,T]$. The closed curve $\rho(t)$ can be represented in the polar coordinates as 

\begin{equation*}
	\rho(t) = r(t) e^{i\theta(t)}
\end{equation*}
with $r(t) = \parallel \rho(t) \parallel$ and $\theta(t) = \angle \rho (t)$. The Cauchy integral formula makes (\ref{eq:closed curve}) equal to 

\begin{equation}
	\begin{split}
		\rho(t) & = \frac{1}{2\pi i} \int_{0}^{t} \frac{\dot{r}}{r} d \tau + \frac{1}{2\pi}\int_{0}^{t} \dot{\theta} d\tau \\
		& = \frac{1}{2\pi i} \log \left( \frac{r(t)}{r(0)} \right) + \underbrace{\frac{1}{2\pi} (\theta(t) - \theta(0))}_{\mathrm{real \ part}}
	\end{split}
\end{equation}
What we are interested in is the real part of this integral

\begin{equation}
	\label{eq:topological invariant}
	\mathrm{Re}(\rho(t)) = \frac{1}{2\pi} (\theta(t) - \theta(0))
\end{equation}
which is a \textbf{topological invariant}. Intuitively, Equation (\ref{eq:topological invariant}) represents the counting number of times that the curve $\rho(t)$ encircled the origin in the time interval $[0, T]$.

\backmatter  

\printbibliography

\end{document}